\documentclass[times,twocolumn,final,authoryear]{elsarticle}

\newcommand{\generator}{\mathcal{G}}
\newcommand{\netnoise}{\mathcal{E}}
\newcommand{\ourT}{MDT}
\newcommand{\masking}{IAM}
\newcommand{\attbias}{PAAB}
\newcommand{\ourcfg}{MC-CFG}

\newcommand{\methodname}{Collaborative Neural Painting}
\newcommand{\methodshort}{CNP}

\newcommand{\vecdiffnoisepred}{\boldsymbol{\epsilon}^p}

\newcommand{\logsnr}{\log{\text{SNR}}}

\newcommand{\vecsequence}{\mathbf{s}}
\newcommand{\stroke}{\mathbf{s}}
\newcommand{\vecsequencecond}{\mathbf{s}^{ctx}}
\newcommand{\vecsequencepred}{\mathbf{s}^p}
\newcommand{\vecsequenceemb}{\mathbf{e}}
\newcommand{\vecsequenceembcond}{\mathbf{e}^{ctx}}
\newcommand{\vecsequenceembpred}{\mathbf{e}^p}
\newcommand{\vecmask}{\mathbf{m}}

\newcommand{\numfeatures}{F}
\newcommand{\numstrokes}{L}

\newcommand{\class}{c}
\newcommand{\difftimestep}{t}
\newcommand{\seqtimestep}{i}

\newcommand{\strokex}{x}
\newcommand{\strokey}{y}
\newcommand{\strokeh}{h}
\newcommand{\strokew}{w}
\newcommand{\stroker}{\theta}
\newcommand{\strokecr}{r}
\newcommand{\strokecg}{g}
\newcommand{\strokecb}{b}

\newcommand{\suppmat}{Supp. Mat.}

\usepackage{ycviu}
\usepackage{framed,multirow}

\usepackage{amssymb}
\usepackage{amsthm}
\usepackage{amsmath}
\usepackage{latexsym}

\usepackage{url}
\usepackage{hyperref}
\usepackage{xcolor}
\definecolor{newcolor}{rgb}{.8,.349,.1}

\usepackage{xspace}
\makeatletter
\DeclareRobustCommand\onedot{\futurelet\@let@token\@onedot}
\def\@onedot{\ifx\@let@token.\else.\null\fi\xspace}

\def\eg{\emph{e.g}\onedot} 
\def\ie{\emph{i.e}\onedot}

\def\wrt{w.r.t\onedot} 
\def\etal{\emph{et al}\onedot}
\makeatother

\usepackage{booktabs}
\usepackage{multirow}

\usepackage{array}
\usepackage{ragged2e}
\newcolumntype{P}[1]{>{\RaggedRight\hspace{0pt}}p{#1}}
\newcolumntype{X}[1]{>{\RaggedRight\hspace*{0pt}}p{#1}}

\usepackage{tcolorbox}

\usepackage{tikz}
\usetikzlibrary{arrows,shapes,positioning,shadows,trees,mindmap}
\usepackage[edges]{forest}
\usetikzlibrary{arrows.meta}
\colorlet{linecol}{black!75}
\usepackage{xkcdcolors} 

\usepackage{tikz}
\usetikzlibrary{backgrounds}
\usetikzlibrary{arrows,shapes}
\usetikzlibrary{tikzmark}
\usetikzlibrary{calc}
\newcommand{\highlight}[2]{\colorbox{#1!17}{$\displaystyle #2$}}

\usepackage{pifont}
\usepackage{tabularx}
\usepackage{enumitem}
\setlist[itemize]{noitemsep, topsep=0pt}
\usepackage{color, colortbl}

\definecolor{DarkGreen}{HTML}{008B72}
\newcommand{\cmark}{\textcolor{DarkGreen}{\ding{51}}}
\newcommand{\xmark}{\textcolor{red}{\ding{55}}}

\renewcommand{\highlight}[2]{\colorbox{#1!17}{#2}}

\newcommand\st{\textsuperscript{st}\xspace}
\newcommand\nd{\textsuperscript{nd}\xspace}
\newcommand\pp[1]{\noindent\textbf{#1.}}

\journal{Computer Vision and Image Understanding}

\begin{document}

\thispagestyle{empty}

\clearpage

\ifpreprint
  \setcounter{page}{1}
\else
  \setcounter{page}{1}
\fi

\begin{frontmatter}

\title{Collaborative Neural Painting}

\author[1,2]{Nicola \snm{Dall'Asen}\corref{cor1}} 
\cortext[cor1]{Corresponding author:}\ead{nicola.dallasen@unitn.it}
\author[1]{Willi \snm{Menapace}}
\author[1]{Elia \snm{Peruzzo}}
\author[3]{Enver \snm{Sangineto}}
\author[4]{Yiming \snm{Wang}}
\author[1,4]{Elisa \snm{Ricci}}

\address[1]{University of Trento, Italy}
\address[2]{University of Pisa, Italy}
\address[3]{University of Modena and Reggio Emilia, Italy}
\address[4]{Fondazione Bruno Kessler, Trento, Italy}

\received{1 May 2013}
\finalform{10 May 2013}
\accepted{13 May 2013}
\availableonline{15 May 2013}
\communicated{S. Sarkar}

\begin{abstract}
The process of painting fosters creativity and rational planning. However, existing generative AI mostly focuses on producing visually pleasant artworks, without emphasizing  the painting process. 
We introduce a novel task, \textit{\methodname~(\methodshort)}, to facilitate collaborative art painting generation between users and agents. Given any number of user-input  \textit{brushstrokes} as the context or just the desired \textit{object class}, \methodshort~should produce a sequence of strokes supporting the completion of a coherent painting. 
Importantly,  the process can be gradual and iterative, so  allowing users' modifications at any phase until the completion. 
Moreover, we propose to solve this task using a painting representation based on  a sequence of parametrized strokes, which makes it easy both editing and composition operations. 
These parametrized strokes are processed by a Transformer-based architecture with a novel attention mechanism to model the relationship between the input strokes and the strokes to complete. We also propose a new masking scheme to reflect the interactive nature of \methodshort~and adopt diffusion models as the basic learning process for its effectiveness and diversity in the generative field. 
Finally, to develop and validate methods on the novel task, we introduce a new dataset of painted objects and an evaluation protocol to benchmark \methodshort~both quantitatively and qualitatively. We demonstrate the effectiveness of our approach and the potential of the \methodshort~task as a promising avenue for future research. 
Project page and code: \href{https://fodark.github.io/collaborative-neural-painting/}{this https URL}. 
\end{abstract}


\end{frontmatter}



\section{Introduction}

\begin{figure*}[!ht]
  \centering
  \includegraphics[width=.9\textwidth]{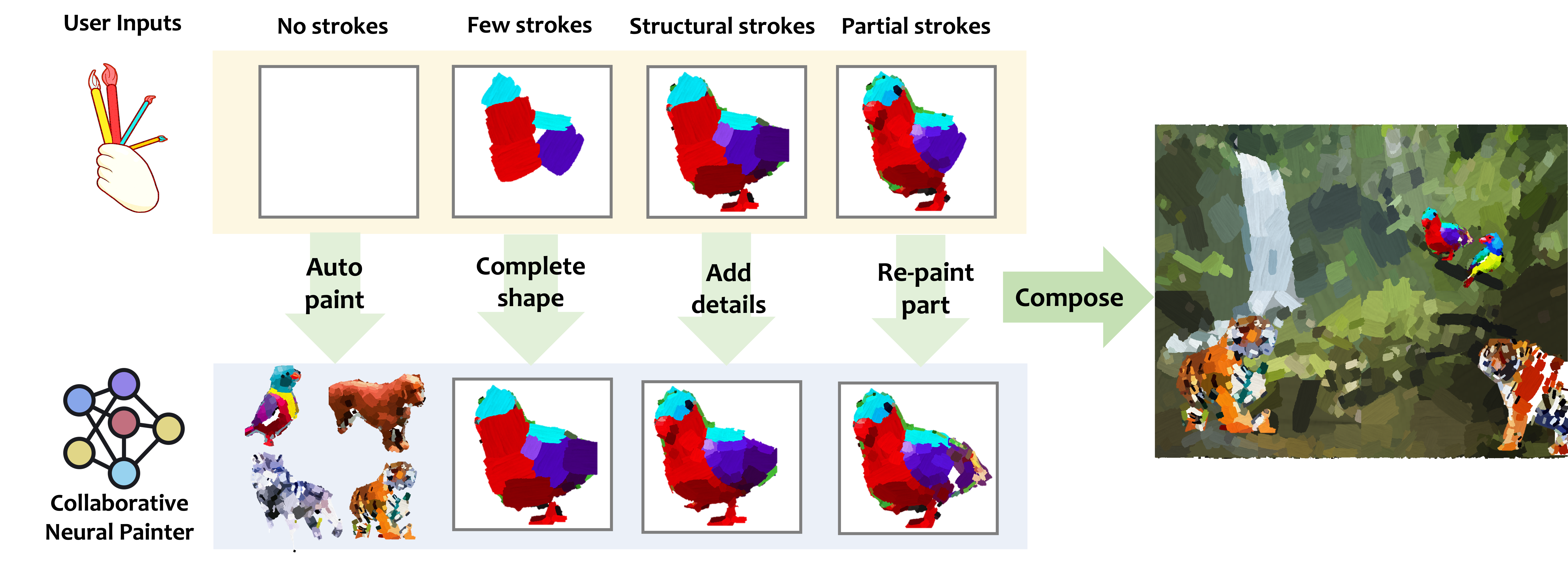}
  \caption{The proposed \methodname~task envisions a collaborative procedure in which users produce and compose artworks iteratively interacting with a Neural Painter. This interaction  includes auto-painting without user input, and assistive painting/editing at any granularity level. We solve this generation task using a vectorized stroke parametrization whose joint distribution is learned using diffusion models.} 
\label{fig:teaser}
\end{figure*}

\label{sec:intro}
In recent years we have experienced an explosive growth of \textit{AI Art}, empowering users with the possibility to generate images and other media content given various forms of conditioning, such as text~\citep{saharia2022photorealistic,rombach2022high} or semantic maps~\citep{zeng2022scenecomposer,avrahami2023spatext,zhang2023adding}. 
Generative AI art in the visual domain has led to astonishing results for image synthesis tasks~\citep{nichol2021glide,ramesh2022hierarchical,saharia2022photorealistic,rombach2022high}. It typically operates on the pixel space, both for content generation and editing, where users can modify the image by indicating the editing areas or describing the desired modifications~\citep{nichol2021glide,meng2021sdedit}.  

However, this process of art generation and editing is fundamentally different from how humans create art, of which painting is a primary example~\citep{nakano2019neural}. When painting, humans reason on individual \textit{brushstrokes} rather than \emph{pixels}, and creativity is fostered throughout the stroke design and their compositional planning. 
While several works have been proposed which employ a brushstroke formulation, they mainly focus on producing the entire painting given a reference image or artwork~\citep{liu2021paint, zou2021stylized,Kotovenko_2021_CVPR}, without involving humans in the generation process.
This limits the degree to which such methods promote creativity and empathy~\citep{gerry2017paint,pelowski2017creativity} that are particularly important for enhancing educational and pedagogical development in children~\citep{pajooheffectiveness} or rehabilitation~\citep{zhang2021gaze}. More recently, \citep{peruzzo2023interactive} proposes an interactive formulation of neural painting in which the model suggests the next strokes to paint according to a given \textit{reference} image. The reference-guided interaction might be useful for the learning purpose, however, we argue that such setup may limit the fostering of creativity during painting, as the generation is bounded to the given reference picture.

In this work, we propose the novel task of \textbf{\methodname} (\methodshort) which aims to enable a collaborative creation process of high-quality paintings that promotes active engagement of human users for creative generation. 
We design the interactive process based on brushstrokes that the human users can provide as context at any phase of their painting. This is different from previous works~\citep{peruzzo2023interactive,zou2021stylized,liu2021paint}, as we do not require any reference image and the user can specify any arbitrary number of strokes.
The agent, \textit{i.e.}, the collaborative neural painter, subsequently reasons on the context strokes and produces new strokes in the painting to match the user's painting intentions \eg completing the rough shape of the painting, adding details or regenerating a missing part, as shown in Figure~\ref{fig:teaser}.

\methodshort~is a non-trivial task. It requires understanding the geometric properties of the strokes and how they translate to the final painting to generate rich and diverse outputs. It also requires to deal with an interactive generation process, where the user can intervene at any time during the painting creation.
In this paper, we also propose to solve \methodshort~using a novel Transformer architecture which models the relationships between the user's  and the generated strokes. Specifically, we introduce a novel Position-aware Attention Bias (PAAB) mechanism in order to encourage  neighboring strokes to share similar semantics.
Furthermore, we propose to model user-agent interactions using an Interaction-aware Masking (IAM) procedure which simulates interactions with the user at training time. To capture the complex conditional distribution of the strokes given the user's input, we adopt the diffusion framework~\citep{ho2020denoising} which has demonstrated remarkable results in conditional generation tasks and naturally supports the generation of diverse outputs.

Moreover, to facilitate this study and to promote future research in the proposed task, we present a \methodshort~benchmark, composed of a novel dataset of painted objects and a new evaluation protocol which takes into account both the quantitative and qualitative aspects of the \methodshort~task. Creating such a dataset is challenging, since objects in natural images are frequently occluded, incomplete, or have a low resolution, also depending on their relative position within the scene. To overcome these limitations, we employ a state-of-the-art model to create~\citep{rombach2022high} and segment~\citep{liu2023grounding,zhao2023fast} objects from generated images, and convert the resulting segmented objects into a stroke representation~\citep{zou2021stylized}.
Our method achieves the best performance among all metrics  when compared with other state-of-the-art models that work with sequential data.

The contributions of our paper are summarized as follows:
\begin{itemize}[noitemsep,nolistsep,leftmargin=*]
    \item We propose a novel task,~\textit{\methodname}, to encourage art generation process with active human engagement.
    \item We create and release a novel curated dataset to facilitate research on the task.
    \item We design a benchmark for the task with an evaluation protocol with both objective measures and subjective evaluation.
    \item We propose a diffusion model framework to address \methodname, with a novel attention mechanism and masking scheme to model the user-agent interaction. Our proposed method outperforms recent baselines on our benchmark. We also showcased its effectiveness in real demonstration for collaborative painting with human users.
\end{itemize}
\section{Related Works}
In this section, we thoroughly discuss research works that are related to Neural Painting, Controllable Image Generation and Editing and Masked Data Modelling in generative tasks.  
\subsection{Neural Painting}
Neural Painting (NP) refers to the task of decomposing a natural image into a set of parameterized strokes. The resulting strokes can be rendered on the canvas, obtaining an artistic version of the original image which mimics the result of an actual painting. The seminal works of \citep{Haeberli1990, Litwinowicz97, Hertzmann98} target this task for the first time, proposing heuristic-based methods to decompose a given image into a set of strokes. A parallel line of works focuses on developing rendering techniques that could faithfully represent, on digital media, the effect of different physical brushes \citep{wang2014towards, Sochorov2021, Strassmann86, Curtis97}.  
With the progress of deep learning, these methods have been replaced by learning-based methods. In particular, the task is formulated as a reinforcement learning (RL) problem, where the agent is the painter, the action space is represented by the different strokes the agent can displace in the canvas, and the reward is given by the similarity between the painted canvas and the reference image \citep{Huang2019LearningTP, Singh2021IntelliPaintTD, schaldenbrand2021content}. Differently, Zou \emph{et.~al.~}\citep{zou2021stylized} formalizes the task as an optimization problem, where the stroke parameters are directly optimized to approximate the reference image leveraging a differentiable renderer. To overcome the time burden of the optimization process, \citep{liu2021paint} proposes to adopt a Transformer-based architecture for regressing the stroke parameters and design a synthetic data generation pipeline to train the model.

Despite achieving good qualitative results, previous methods do not consider the dimension of user interaction in the generation process. Intelli-Paint \citep{Singh2021IntelliPaintTD} introduces this idea in a reinforcement learning framework, rewarding the agent for painting in a layered and localized way, more like a human would do. Another line of work is represented by Interactive Neural Painting~\citep{peruzzo2023interactive} in which the model \textit{interacts} with the user by predicting the \textit{next} strokes based on the user input. The final goal of the model is to paint a reference image provided by the user.

Our work shares the same motivation with \citep{Singh2021IntelliPaintTD,peruzzo2023interactive}, but we further push the boundaries of interactivity by explicitly targeting a collaborative scenario with the user without a reference image and without limiting the number of strokes predicted by the model. The goal of our model is to coherently complete an \textit{object} based on the interactions with the user. Moreover, we address the more realistic problem of generating a painting without the reference image, which is a task a human can easily accomplish but none of the previous NP methods can do.

\subsection{Controllable Image Generation and Editing}
There exists a substantial corpus of literature for the purposes of controllable image generation and editing. Numerous GAN-based approaches have been developed, which involve the inversion and manipulation of images within the latent space. Other methods condition the generation on additional signals, which are intuitive for the user to control and modify, like segmentation maps, sketches, and text~\citep{park2019gaugan, zhu2020sean, ghosh2019interactive, liu2021deflocnet, ling2021editgan}. 
Notably, \citep{singh2022paint2pix} proposes to use parameterized strokes, similar to the ones adopted by NP methods, as an intuitive manner of sketching an image, and train a StyleGAN-like model to generate realistic images from incomplete drawings. 
Recently, diffusion-based approaches emerged as state-of-the-art for controllable image generation. Most of these methods are powered by large pre-trained diffusion models ~\citep{rombach2022high, nichol2021glide, balaji2023ediffi, saharia2022photorealistic}, which are then finetuned or adapted for the specific task at hand. One line of works explores generating image variations given a set of representative images, either by finetuning the whole model~\citep{ruiz2022dreambooth} or by optimizing a text embedding~\citep{gal2022textual, mokady2022null}. Exploiting the recent advancement in language and vision understanding, other works propose to use text as an intuitive way of modifying the generated image~\citep{brooks2022instructpix2pix, hertz2022prompt, parmar2023zero, pnpDiffusion2022}.
To improve control over the spatial layout of the final output, segmentation masks are introduced as conditioning in~\citep{zeng2022scenecomposer, avrahami2023spatext}, while sketches are used in~\citep{voynov2022sketch}. ControlNet~\citep{zhang2023adding} proposes an effective condition mechanism by training a hypernetwork on a dataset of paired examples consisting of images and condition signals, and conditioning the main pre-trained network using skip-connections.

Our work differs from these methods because we do not operate directly in the pixel space, but adopt a vectorial representation of the image. By representing the image as a set of parameterized strokes, we can achieve editing capabilities by design. Our method offers the flexibility to make adjustments at any level of detail, without relying on external models. Full control is given to the user, who can change the layout of the image by modifying the position of the strokes, resizing them, and rendering at arbitrary output resolution.

As summarized in Tab.~\ref{tab:positioning}, we position our work with respect to the existing literature in terms of \textcolor{red}{(i)} the model's capability to synthesize images from scratch, \textcolor{blue}{(ii)} the ability to support interactive generation, \textcolor{DarkGreen}{(iii)} the dependency on large-scale datasets or external models (\eg pretrained segmentation networks), and \textcolor{orange}{(iv)} the vectorial representation of images, which provides built-in editing capabilities.

\subsection{Masked Data Modeling in Generative tasks}
Masked Data Modeling refers to the task of reconstructing partially corrupted data, and recently gained popularity both for representation learning and generative purposes.  Notably, BERT~\citep{devlin2018bert} proposes a masked reconstruction strategy as a strong pretraining objective for NLP tasks. Different from autoregressive formulation, masking enables bidirectional context, increasing the model's expressiveness and performance. In the visual domain, MaskGIT~\citep{chang2022maskgit} successfully applies a similar strategy to image generation, introducing an iterative sampling policy at inference.  
Recently, denoising diffusion models~\citep{sohl2015deep,ho2020denoising, song2019generative} have emerged as a class of generative models showing state-of-the-art performances in many tasks. Masked Data Modeling can be integrated into the training of Diffusion Models, treating the unmasked regions as conditioning information. MAGVIT \citep{yu2023magvit} extends MaskGIT to video domains, using a similar masking strategy. Wei \etal \citep{wei2023diffusion} propose to condition diffusion models on masked input to formulate them as masked autoencoders, which enables image inpainting at different levels of detail. Following this line of work, Tashiro \etal~\citep{tashiro2021csdi} propose a score-based diffusion model for the imputation of missing values in time series, simulating the missing data at training time through a carefully designed masking procedure. 
We build on these emerging trends and introduce a diffusion-based method for controllable neural painting exploiting masked data modeling as a methodology to simulate users' interactions with the model. Differently from previous works, our masking strategies are not \textit{random} but \textit{interaction-aware}, making it suitable for interactive tasks.

\begin{table}[!t]
\caption{Positioning of our work \wrt Neural Painting and Image Editing models.}
\centering
\resizebox{1.\linewidth}{!}{%
\begin{tabular}{@{}lccc@{}}
\toprule
\textbf{Task} & \textbf{Neural Painting} & \textbf{Image Editing} & \textbf{Ours} \\
\midrule
\textcolor{red}{(i) }Synthesis from scratch & \xmark & \cmark & \cmark \\
\textcolor{blue}{(ii) }Interactive generation & \xmark & \cmark & \cmark \\
\textcolor{DarkGreen}{(iii) }No dependency on external models & \cmark & \xmark & \cmark \\
\textcolor{orange}{(iv) }Vectorial representation & \cmark & \xmark & \cmark \\
\bottomrule
\end{tabular}
\vspace{-2.em}
}
\label{tab:positioning}
\end{table}

\section{Preliminaries}
We first introduce some knowledge that serves as background for our method, in terms of diffusion models and the conditional generation with classifier-free guidance.

\pp{Diffusion Models}
Gaussian diffusion models rely on a forward noising process that gradually applies noise to real data $x_0$: $q(x_t|x_0) = \mathcal{N}(x_t; \sqrt{\bar{\alpha}_t}x_0, (1 - \bar{\alpha}_t)\mathbf{I})$, where constants $\bar{\alpha}_t$ are hyperparameters. The model then learns the reverse process to invert forward corruption process: $p_\theta(x_{t-1}|x_t) = \mathcal{N}(\mu_\theta(x_t), \Sigma_\theta(x_t))$, using neural networks to approximate the intractable distribution $q(x_{t-1}|x_t)$. Using the $\epsilon$-parametrization \citep{ho2020denoising}, the model is trained with the mean-squared error between the predicted noise $\epsilon_\theta(x_t)$ and the ground truth sampled Gaussian noise $\epsilon_t$:

\begin{equation}
    \mathcal{L}_{simple}(\theta) = ||\epsilon_\theta(x_t) - \epsilon_t||_2^2
    \label{eq:loss_simple}
\end{equation}

At inference time, new data can be sampled by initializing $x_{T} \sim \mathcal{N}(0,\mathbf{I})$, and sampling $x_{t-1} \sim p_\theta(x_{t-1}|x_t)$.
The reverse process can be expressed as:

\begin{equation}
    p_\theta(x_{t-1}|x_t) = \frac{1}{\sqrt{\alpha_t}} \left( x_t - 
    \frac{1 - \alpha_t}{\sqrt{1 - \bar{\alpha_t} }} \epsilon_\theta(x_t) \right) + \sigma_t\mathbf{z}
    \label{eq:denoising}
\end{equation}

We adapt the DDPM formulation~\citep{ho2020denoising} and modify it to work in the masked setting of interactive neural painting, as we aim at denoising only a part of the stroke sequence.

\pp{Classifier-free guidance} Conditional diffusion models take additional information, such as a class label $c$, as input. In this scenario, the reverse process becomes $p_\theta(x_{t-1}|x_t,c)$. To guide the probability mass towards data where the implicit classifier $p_\theta(c|x_t)$ has a high likelihood, \textit{classifier-free guidance} can be employed~\citep{ho2022classifier} and can produce considerably improved samples over generic sampling methods~\citep{nichol2021glide,ramesh2022hierarchical,peebles2022scalable}. This requires training the model in both conditional and unconditional cases and merging the predicted scores. During training, the evaluation of the diffusion model with $c=\emptyset$ is accomplished by randomly dropping out $c$ and replacing it with a learned "null" embedding $\emptyset$. At inference time, given a guidance scale $s > 1$, the modified score becomes:

\begin{equation}
    \hat{\epsilon}_\theta(x_t, c) = \epsilon_\theta(x_t, \emptyset) + s \cdot (\epsilon_\theta(x_t, c) - \epsilon_\theta(x_t, \emptyset))
\end{equation}

\section{Task formulation}
\label{sec:formulation}

\begin{figure*}[!ht]
  \centering
  \includegraphics[width=.9\textwidth]{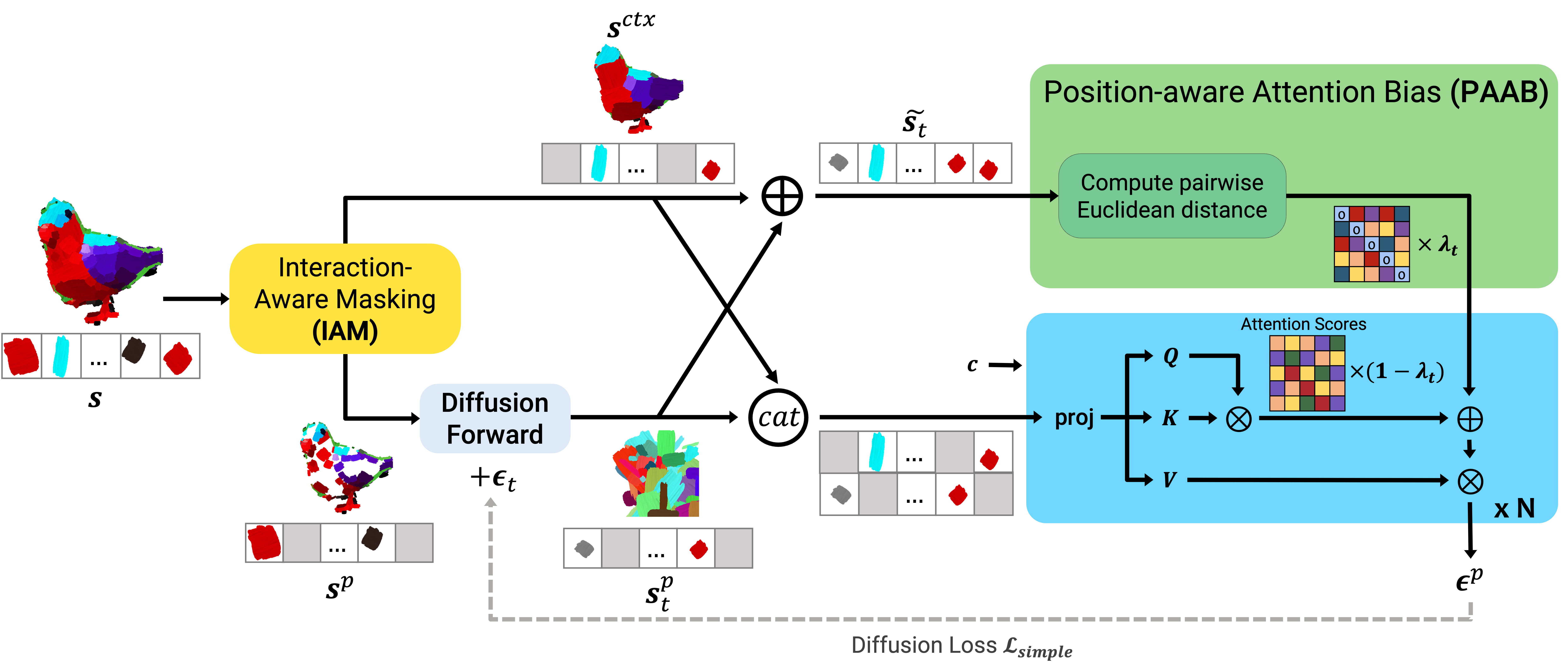}
  \caption{Given an input sequence $\vecsequence$, the Interaction-Aware Masking block divides the sequence in two, a conditioning sequence $\vecsequencecond$ which acts as context to denoise the missing strokes $\vecsequencepred$ that, using a diffusion framework, are noised to $\vecsequencepred_t$. Our Position-aware Attention Bias modifies the attention scores of our Transformer based on the Euclidean distance between the conditioning and noised strokes.
  }
  \label{fig:architecture}
\end{figure*}

In this section, we define the  \methodname~(\methodshort) task, where a painting is iteratively generated according to different conditioning signals provided by the user. We represent a painting using a sequence of parametrized strokes $\vecsequence=(\stroke_1, ..., \stroke_\numstrokes) \in \mathbb{R}^{\numstrokes \times 8}$, where $\numstrokes$ is the number of strokes. Following previous literature \citep{zou2021stylized}, each stroke $\stroke_\seqtimestep \in \mathbb{R}^8$ is defined as a set of 8 parameters 
which describe: the position ($\strokex$, $\strokey$), the size ($\strokew$, $\strokeh$), the rotation ($\stroker$) and the color ($\strokecr$, $\strokecg$, $\strokecb$). 
The stroke sequence can be used to render the painting in the pixel space using a \emph{parameter-free renderer} \citep{liu2021paint}. Given a primitive brushstroke, we apply a set of affine transformations dictated by the stroke parameter, obtaining the foreground and the alpha matte of the given stroke. The final painting is obtained by sequentially composing the individual strokes on the canvas. We refer to \citep{liu2021paint} for additional details.

\begin{figure}[t!]
\resizebox{1.\linewidth}{!}{%
\setlength\tabcolsep{0.4pt}
\begin{tabular}{c|c|c|c}
$m=1$ & $m=2$ & $m=3$ & $m=4$ \\
\includegraphics[width=0.1\textwidth]{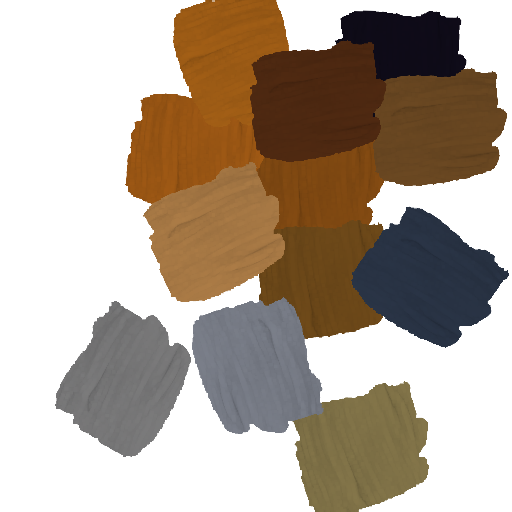} &
\includegraphics[width=0.1\textwidth]{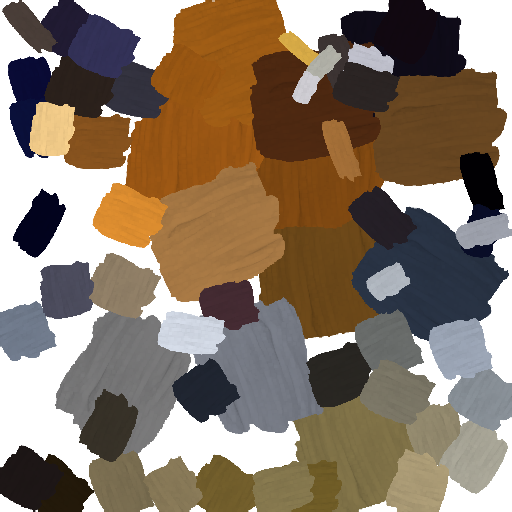} &
\includegraphics[width=0.1\textwidth]{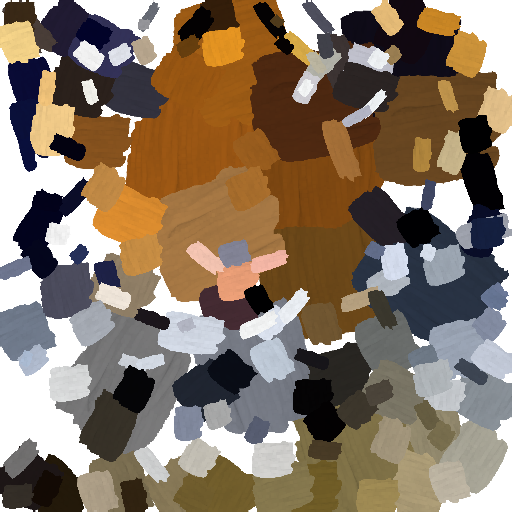} &
\includegraphics[width=0.1\textwidth]{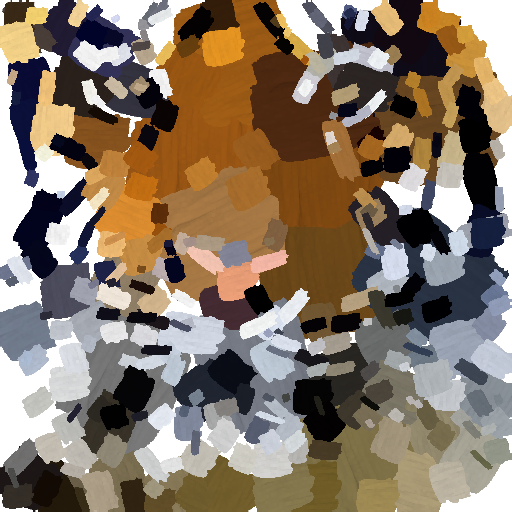} \\
\includegraphics[width=0.1\textwidth]{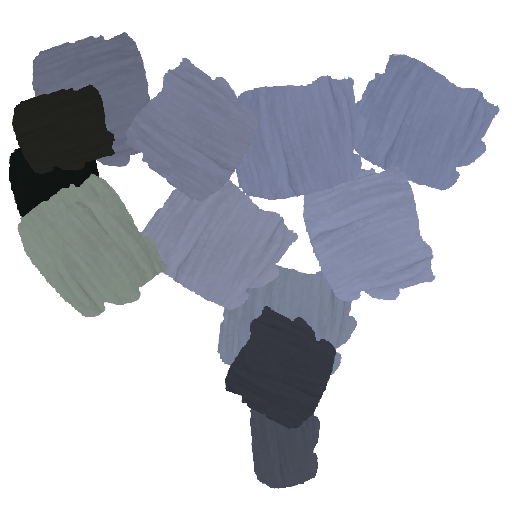} &
\includegraphics[width=0.1\textwidth]{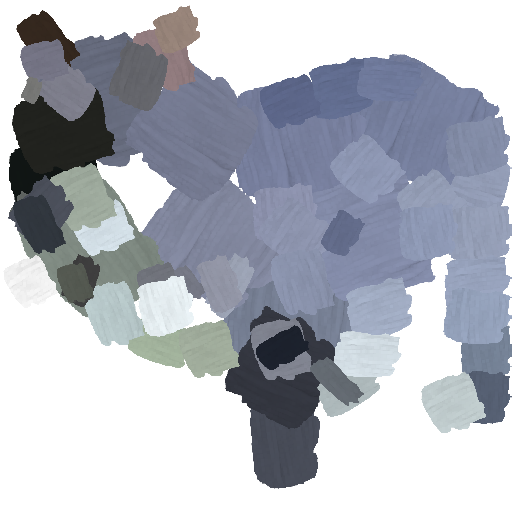} &
\includegraphics[width=0.1\textwidth]{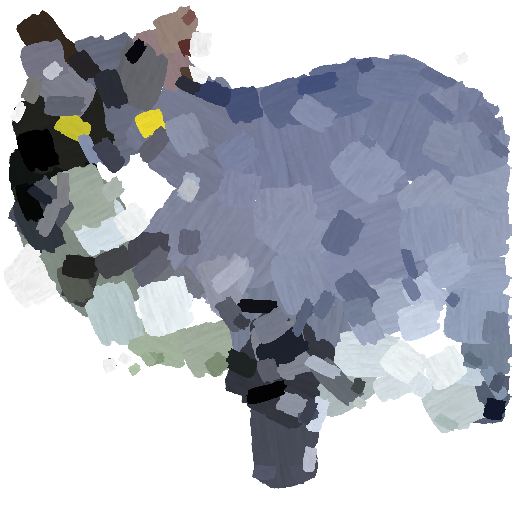} &
\includegraphics[width=0.1\textwidth]{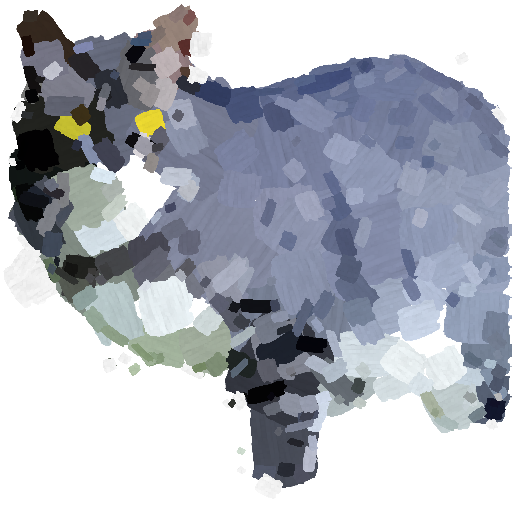} \\
\includegraphics[width=0.1\textwidth]{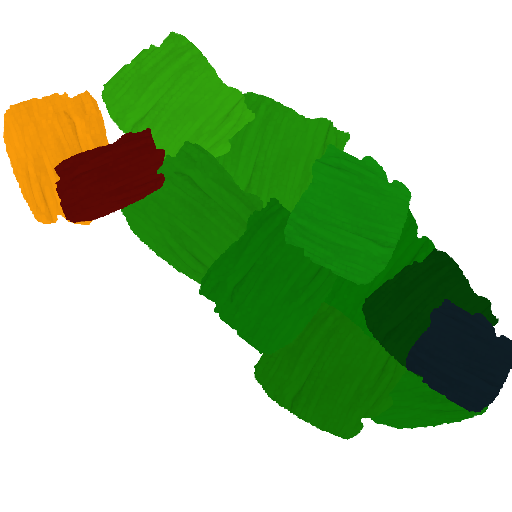} &
\includegraphics[width=0.1\textwidth]{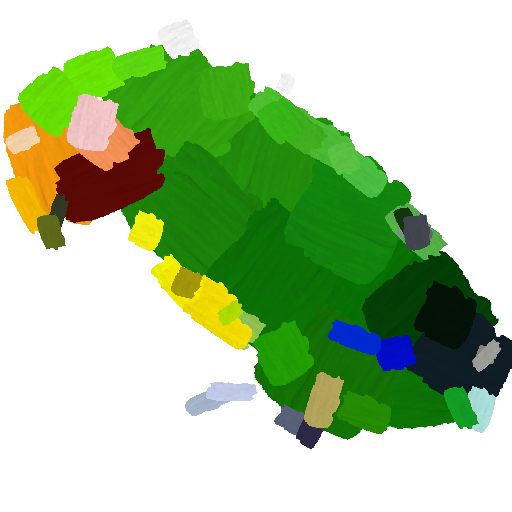} &
\includegraphics[width=0.1\textwidth]{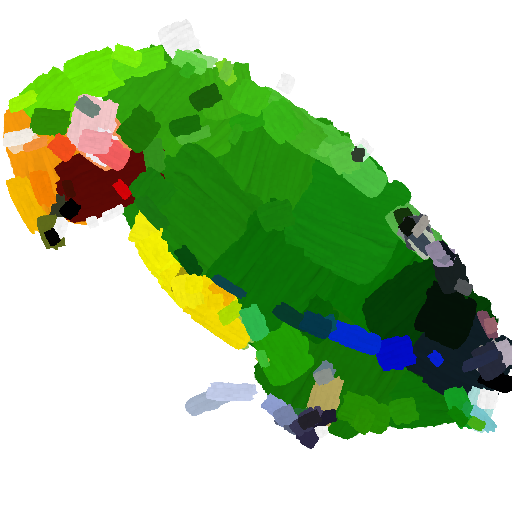} &
\includegraphics[width=0.1\textwidth]{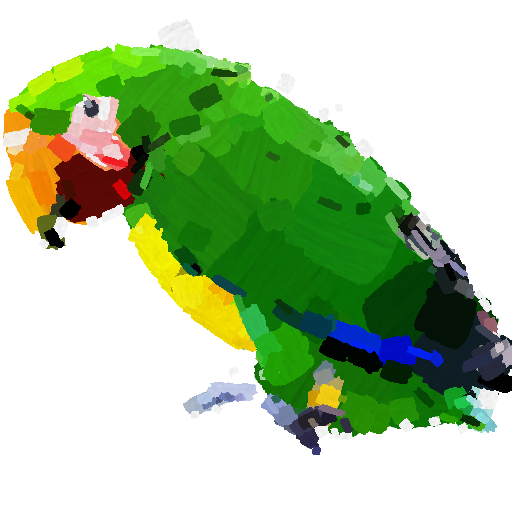} \\
\includegraphics[width=0.1\textwidth]{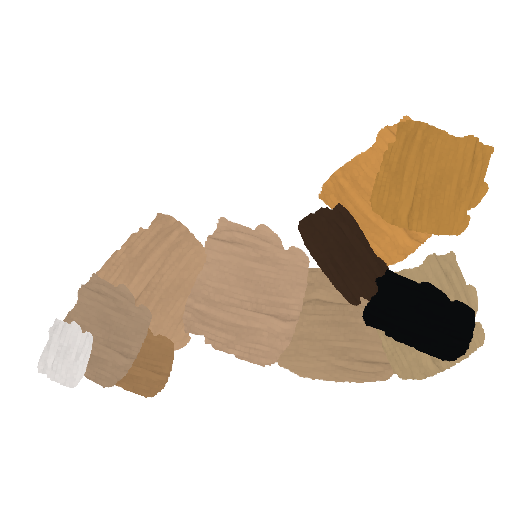} &
\includegraphics[width=0.1\textwidth]{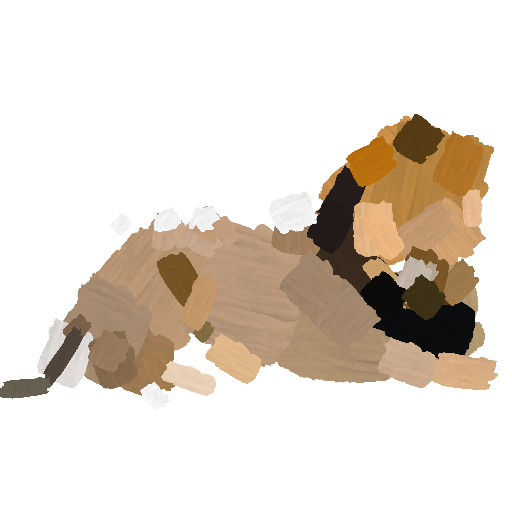} &
\includegraphics[width=0.1\textwidth]{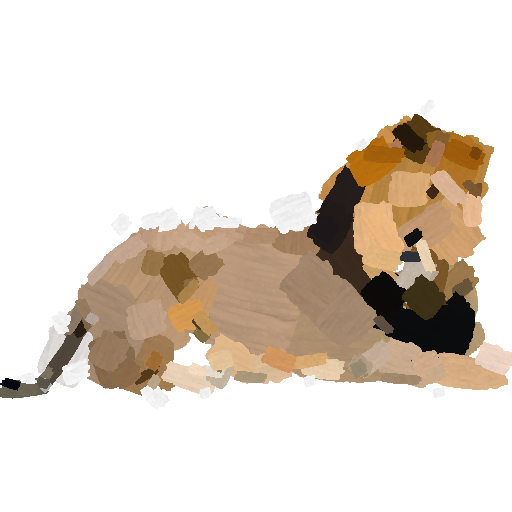} &
\includegraphics[width=0.1\textwidth]{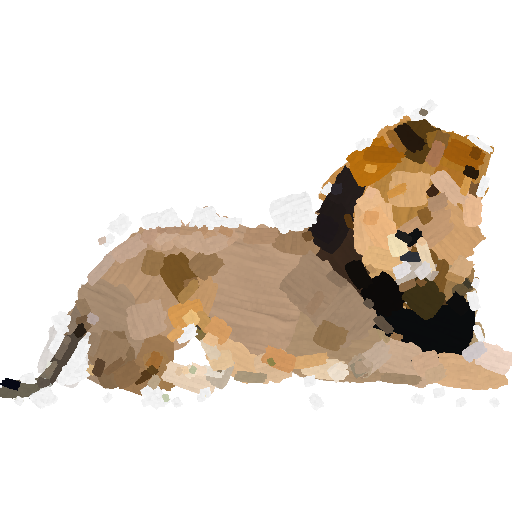} \\

\end{tabular}
}
\caption{Examples of granularity levels. From left to right shows the granularity from the lowest ($m\!=\!1$) to the highest ($m\!=\!4$).}
\label{fig:granularity}
\end{figure}

To mimic how human paints, we define and organize the stroke sequence into different \textbf{granularity levels}~\citep{zou2021stylized}: from coarse strokes that define the object's shape in the first level towards smaller and finer details in the following levels. 
Specifically, we progressively divide the input image into overlapping blocks of size $m \times m$, with $m$ representing the granularity level. For each block in the current level, we initialize $N = 12$ strokes and optimize the parameters following \citep{zou2021stylized} (See Fig.~\ref{fig:granularity}). In this work, we set the maximum number of levels $m$ to 4, resulting in 400 strokes per image.


To enable collaborative painting, we also introduce two complementary conditioning signals which can be used by the user to interact with the painting process: a class label $\class$ and a sequence of strokes $\vecsequencecond \in \mathbb{R}^{\numstrokes' \times 8}$ representing a partially incomplete painting. 
More formally, the \methodname~process consists in a $\generator$, a.k.a. the paining agent, that produces the sequence of strokes $\vecsequence$ given the user-provided conditioning signals: 

\begin{equation}
\label{eq:generator_equation}
\vecsequence=\generator(\class,\vecsequencecond).
\end{equation}

This formulation enables the \emph{collaborative} generation process with great freedom. In its simplest form, the \methodshort~framework can be used for the completion of paintings without any context, \ie by setting $\vecsequencecond$ to an empty sequence $\emptyset$, and specifying only the desired object class $c$. Other functionalities are also possible by varying the conditioning signals. For instance: i) a coarse representation of the painting can be predicted with few strokes from the user, ii) a detailed painting can be produced from a coarse drawing by representing the coarse painting as the conditioning sequence $\vecsequencecond$ with or without class supervision, iii) an entire part of a generated subject can be altered by erasing it and letting the generator complete the painting with or without class supervision. 
The generation can be iterative according to user's demand, by simply updating the sequence $\vecsequencecond$ in Eq.~\ref{eq:generator_equation},until the desired painting is obtained (see Fig.~\ref{fig:teaser} for visual examples of these interactions).

\section{Proposed Method}
\label{sec:method}

Our method is designed to mimic a collaborative scenario between the user and the painting agent $\generator$. Given a sequence of strokes $\vecsequence$ describing the painting of an object of class $c$, we divide it into two subsequences: the conditioning sequence $\vecsequencecond$ and the target sequence $\vecsequencepred$. We propose an \emph{Interaction-aware masking (IAM)} procedure to capture the different types of interactions between the user and $\generator$ and use it to split the sequence $\vecsequence$ into context and target respectively. 

Moreover, to effectively model the relationship between the target and conditioning stroke sequences, we introduce a \emph{Masked Diffusion Transformer (MDT)} based on~\citep{peebles2022scalable}. However, we find that the standard attention mechanism does not capture the relationship between strokes satisfactorily. To address this problem, we devise a \emph{Position-aware Attention Bias (PAAB)} that encourages higher attention scores between spatially neighboring strokes. 

We train our model using the $\epsilon$-parametrization of the diffusion framework (see Eq.~\ref{eq:loss_simple}). Random noise is added to the target sequence $\vecsequencepred$, and the model is trained to reconstruct the noise conditioned on the class information $c$ and the clean context sequence $\vecsequencecond$. At inference time, the user provides the context strokes (if any) and the class information. 
We then initialize the predicted strokes $\vecsequencepred_T$ with Gaussian noise and denoise them for $T$ steps conditioned on the user's inputs. 
In this stage, we leverage classifier-free guidance to increase the faithfulness of the predictions to the conditioning signals and extend it to a \emph{Multiple Conditions Classifier-free guidance} to treat each of them separately. 

In the next sections, we describe each part of our method in detail; 
Sec.~\ref{sec:masking} describes the Interaction-aware masking procedure, 
Sec.~\ref{sec:transformer} describes our Transformer-based architecture, 
Sec.~\ref{sec:attention_bias} describes our Position-aware Attention Bias, 
and finally, Sec.~\ref{sec:classifier_free} describes the employed Multiple Conditions Classifier-free guidance procedure.

\subsection{Interaction-aware Masking}
\label{sec:masking} 

The collaborative scenario described in Sec.~\ref{sec:formulation} covers diverse ways the user and the painting agent can interact. They include modifying, adding, or deleting individual strokes, re-generating a localized area of the painting, completing a coarse sketch provided by the user, and even generating an entire painting from scratch.

To mimic these types of interactions at training time,  
we introduce Interaction-aware masking (\masking) which, differently from the random masking strategies employed in previous works~\citep{devlin2018bert}, simulates at training time the interaction patterns that are expected at inference. In practice, we design different masks $\vecmask$ to cover various use cases: 

\begin{itemize}[leftmargin=*]   
    \item\textbf{Granularity level}: simulates the task of generating fine details from coarse sketches of an object. We sample a random level of detail in the input sequence (see Sec.~\ref{sec:formulation}), and mask all the strokes belonging to successive levels with finer details. 
    \item\textbf{Random}: simulates retouching operations on a given painting, involving the generation of strokes in different spatial locations and conveying different levels of details. We use random masking to mimic this use case. \item\textbf{Square}: simulates the generation of a local area of the painting. We 
    remove multiple strokes in the sequence, based on the distance from a pivot stroke chosen at random, effectively creating a square mask in the rendered image. Note that spatially neighbor strokes are not necessarily consecutive in the stroke sequence.
    \item\textbf{Block}: simulates the undoing of the last $N$ consecutive strokes suggested by the model. We achieve it by masking adjacent strokes in the sequence which can span across granularity levels and be not spatially related to each other. 
    \item\textbf{No context}: simulates the case where no context stroke $\vecsequencecond$ is provided by the user. In this case, we mask out the whole conditioning sequence and condition the model only on the desired object class $c$.
\end{itemize}

\noindent During training, for each sample, we randomly select one of these strategies and produce $\vecsequencecond = \vecsequence \odot \vecmask$, where $\vecsequence$ is the stroke sequence sampled from the dataset.

\subsection{Masked Diffusion Transformer}
\label{sec:transformer}

In this section, we describe the Masked Diffusion Transformer on which the collaborative painting generation process is based. We adopt the DDPM diffusion framework~\citep{ho2020denoising} and introduce a model $\netnoise$ parametrized to predict the noise in the stroke sequence (see Fig.~\ref{fig:architecture}). The model is conditioned on the class $\class$, the context strokes $\vecsequencecond$, and the diffusion timestep $\difftimestep$:
\begin{equation}
    \vecdiffnoisepred = \netnoise(\vecsequencepred_\difftimestep | \class,\vecsequencecond,\difftimestep),
\end{equation}

with $\vecsequencepred_\difftimestep$ being the noisy sequence of target stroke parameters, and $\vecdiffnoisepred$ the predicted noise. Note that no noise is applied to the conditioning information $\vecsequencecond$~\citep{tashiro2021csdi}. We denote with $\tilde{\boldsymbol{s}}_\difftimestep = \vecsequencecond \oplus \vecsequencepred_\difftimestep$ the combination of conditioning and noisy strokes. At inference time, we initialize the predicted strokes $\vecsequencepred_T \sim \mathcal{N}(0,1)$, and denoise them for $T$ steps, with $\tilde{\boldsymbol{s}}_0$ corresponding to the combination of the final predicted sequence and the context information.

Modeling an effective architecture for $\netnoise$ is a challenging task, due to the sequential structure of strokes, and the necessity to model their relationship with the \emph{variable-length} conditioning sequence provided by the user.
We propose an approach based on a Transformer encoder architecture~\citep{vaswani2017attention}, which can handle long input sequences and model pairwise stroke relationships thanks to self-attention. 
To accommodate our problem formulation, the network accepts two sequences of length $\numstrokes$ as inputs. The first sequence is the noisy sequence of strokes $\vecsequencepred_\difftimestep$, while the second is the conditioning strokes sequence $\vecsequencecond$, which provides the context for the denoising process. Each sequence is embedded using separate linear layers to form embedding sequences $\vecsequenceembpred,\vecsequenceembcond \in \mathbb{R}^{\numstrokes \times \numfeatures}$, with $\numfeatures$ being the feature dimension of our Transformer.
We consider strokes in corresponding positions to be mutually exclusive, \emph{i.e.} a stroke is either a conditioning stroke, or a stroke to be predicted, and use a binary mask $\vecmask \in \{0,1\}^\numstrokes$ which indicates, for each position in the sequence, if the stroke represents a conditioning signal (1) or is to be predicted (0). Finally, the input to the Transformer model is $\vecsequenceemb=\vecsequenceembpred \odot (1-\vecmask) + \vecsequenceembcond \odot \vecmask$. 

Additionally, we condition the model on the class $\class$, which is embedded into a learnable vector of size $\numfeatures$.
We combine this vector with the Fourier mapping of the logSNR corresponding to the diffusion time-step $\difftimestep$ \citep{hoogeboom2023simple}, and condition the model with the adaLN-Zero block~\citep{peebles2022scalable}.

The output of the Transformer blocks is projected by a linear layer to the predicted noise $\vecdiffnoisepred \in \mathbb{R}^{L\times8}$. We exclude the context from the loss computation by masking the predicted noise, and train the model with the loss in Eq.~\eqref{eq:loss_simple}.

\subsection{Position-aware Attention Bias}
\label{sec:attention_bias}

The default attention mechanism makes each token attend to any other independently from its semantics. We argue that this behavior is suboptimal for the painting generation task, where each stroke in the sequence bears semantic-rich information. In particular, we consider local proximity as a robust prior information source. It is likely that spatially neighboring strokes will have an impact on one another and present consistent features, such as color or orientation.
These cues can come from either the conditioning sequence, which provides real grounding for missing parameters, or the spatial neighboring strokes in the noisy sequence.

Recalling that traditional attention scores in self-attention are defined as $\sigma_{\text{attn}}(Q, K) = \text{softmax}\left(QK^T / \sqrt{d_k}\right)$, we address this limitation by introducing a novel Position-aware Attention Bias, which biases attention based on the distance between strokes. We compute the distance on the sequence obtained by combining the conditioning and the noisy strokes:
\begin{equation}
    \sigma_{\text{PAAB}}(\tilde{\boldsymbol{s}}_\difftimestep) = \text{softmax}\left(
       - (( \tilde{\boldsymbol{s}}_\difftimestep^i|_x -  \tilde{\boldsymbol{s}}_\difftimestep^j|_x)^2 + ( \tilde{\boldsymbol{s}}_\difftimestep^i|_y -  \tilde{\boldsymbol{s}}_\difftimestep^j|_y)^2) 
    \right),
\end{equation}
where $i, j \in [1, \numstrokes]$ denote the position of the stroke in the sequence. We then combine the two scores with a weighted sum:

\begin{equation}
    \sigma(Q, K, \tilde{\boldsymbol{s}}_\difftimestep) = \lambda_t \cdot \sigma_{\text{PAAB}}(\tilde{\boldsymbol{s}}_\difftimestep) + (1 - \lambda_t) \cdot \sigma_{\text{attn}}(Q, K)
\end{equation}

\noindent
where $\lambda_\difftimestep$ represents the weighting factor.
In early experiments, we found that using constant weighting value leads to suboptimal performances (see Tab.\ref{tab:ablation}). In fact, the larger the corrupting noise, the less confident we can be that two strokes will be neighbors in the final generated sequence. To account for this phenomenon, we propose to control the strength of the attention bias as a function of the logSNR of the diffusion process:

\begin{equation}
    \lambda_\difftimestep = \frac{\logsnr_\difftimestep - \logsnr_{\text{min}}}{\logsnr_{\text{max}} - \logsnr_{\text{min}}} \cdot (\lambda_{\text{max}} - \lambda_{\text{min}}) + \lambda_{\text{min}}
\end{equation}

\subsection{Multiple Conditions Classifier-free guidance}
\label{sec:classifier_free}

Our setting considers two different types of conditioning: the class information $\class$ and the conditioning strokes $\vecsequencecond$.
Liu \emph{et al.}~\citep{liu2022compositional} show that a conditional diffusion model can produce improved results by combining score estimates from various conditioning signals. We apply the same strategy to our model with two separate conditioning input types.
Following Classifier-free guidance formulation~\citep{ho2022classifier}, we train a single model dropping the conditioning signals during training. We use a special learnable token $\emptyset$ to learn the case where the class is not provided, while our \emph{IAM} strategy already captures the scenario where no context stroke is given. 
We introduce two guidance scales, $s_1$ and $s_2$, which control, respectively, the correspondence to the conditioning strokes $\vecsequencecond$ and the class $\class$. The modified score estimate becomes:

    \begin{equation}
    \label{eq:mc_cfg}
    \begin{split}
        \hat{\epsilon}(\vecsequencepred_\difftimestep, \class, \vecsequencecond) = 
        &\tikzmarknode{u}{\highlight{red}{$\hat{\epsilon}(\vecsequencepred_\difftimestep, \emptyset, \emptyset)$}} \\
        &+ s_1 \cdot (\hat{\epsilon}(\vecsequencepred_\difftimestep, \emptyset, \vecsequencecond) - \hat{\epsilon}(\vecsequencepred_\difftimestep, \emptyset, \emptyset)) \\
        &+ s_2 \cdot (\tikzmarknode{b}{\highlight{blue}{$\hat{\epsilon}(\vecsequencepred_\difftimestep, \class, \vecsequencecond)$}} - \tikzmarknode{g}{\highlight{DarkGreen}{$\hat{\epsilon}(\vecsequencepred_\difftimestep, \emptyset, \vecsequencecond)$}})
    \end{split}
    \end{equation}
    \begin{tikzpicture}[overlay,remember picture,>=stealth,nodes={align=left,inner ysep=1pt},<-]
        \path (u.east) ++ (0.2,0.em) node[anchor=west,color=red!67] (scalep){\textbf{unconditional}};
        \draw [color=red!57](u.east) -- ([xshift=.5ex,color=red]scalep.west);
        \path (b.south) ++ (0,-0.5em) node[anchor=north east,color=blue!67] (mean){\textbf{full conditional}};
        \draw [color=blue!57](b.south) |- ([xshift=-0.3ex,color=blue]mean.east);

        \path (g.south) ++ (0.124em,-0.5em) node[anchor=north,color=DarkGreen!67] (mean){\textbf{no class}};
        \draw [color=DarkGreen!57](g.south) -- ([xshift=-0.3ex,color=DarkGreen]mean.north);
    \end{tikzpicture}
    \vspace{1.5\baselineskip}

\noindent Setting $s_1$ and $s_2$ to $1$ would leave only the \textcolor{blue}{\textbf{full conditional}} part. Note that alternative formulations to combine the scores are possible, \eg by switching the position of $\class$ and $\vecsequencecond$, but this formulation naturally accommodates our \methodshort~task.

\section{Evaluation Benchmark}

\begin{table*}[!th]
\caption{Quantitative results of our models compared to the baselines. Stroke $\mathcal{L}_1$ is reported in $\times 10^{-3}$, Image $\mathcal{L}_2$ is reported in $\times 10^{-4}$. All the metrics the lower the better.}
\resizebox{1.\linewidth}{!}{%
\begin{tabular}{@{}cccccccccccccc@{}}
\toprule
\multirow{2}{*}{Method} & \multicolumn{3}{c|}{Block} & \multicolumn{3}{c|}{Level} & \multicolumn{3}{c|}{Random} & \multicolumn{3}{c|}{Square} & \multicolumn{1}{c}{No ctx} \\ \cmidrule(l){2-14} 
\multicolumn{1}{c}{} & FID & Stroke~$\mathcal{L}_1$ & \multicolumn{1}{c|}{Image~$\mathcal{L}_2$} & FID & Stroke~$\mathcal{L}_1$ & \multicolumn{1}{c|}{Image~$\mathcal{L}_2$} & FID & Stroke~$\mathcal{L}_1$ & \multicolumn{1}{c|}{Image~$\mathcal{L}_2$} & FID & Stroke~$\mathcal{L}_1$ & \multicolumn{1}{c|}{Image~$\mathcal{L}_2$} & FID \\ \midrule
Continuous Transformer  & 35.87         & 144                    & 188                              & 81.12         & 145                    & 269                              & 101.54         & 127                    & 665                              & 25.54         & \textbf{135}           & 288                              & 443.86         \\
BERT                    & 142.89        & 198                    & 328                              & 250.54        & 196                    & 406                              & 247.07         & 196                    & 117                              & 33.89         & 197                    & 195                              & 320.43         \\
MaskGIT                 & 149.54        & 205                    & 265                              & 261.27        & 206                    & 398                              & 250.75         & 200                    & 111                              & 35.16         & 211                    & 193                              & 336.18         \\ \midrule
Ours                    & \textbf{6.20} & \textbf{127}           & \textbf{94}                      & \textbf{7.29} & \textbf{125}           & \textbf{100}                     & \textbf{12.69} & \textbf{126}           & \textbf{103}                      & \textbf{5.53} & 142                    & \textbf{154}                     & \textbf{30.12} \\ \bottomrule
\end{tabular}
}
\label{tab:quantitative1}
\vspace{-0.4cm}
\end{table*}

In this section, we introduce a novel benchmark for the \methodname~task, which is designed to evaluate the performance of different methods in a collaborative scenario. We curate a novel dataset~(Sec. \ref{sec:dataset}) that covers a wide range of possible interactions between humans and the agent in the context of painting~(Sec. \ref{sec:setting}). Additionally, we devise a set of metrics to quantitatively evaluate the performance of these models with this dataset~(Sec. \ref{sec:metrics}). 

\subsection{Dataset}
\label{sec:dataset}
Given the novel nature of the proposed task, we could not find any ready-to-use public dataset for it. At the same time, existing large-scale datasets of natural images are not suitable as they do not specifically feature objects, with occlusion possibly compromising the stroke representation.
Therefore, we design a data engine to produce a large curated dataset to enable the study on the \methodshort~task. The data generation pipeline is divided into three steps and we showcase it in Fig.~\ref{fig:pipeline}: 

\begin{itemize}[leftmargin=*]
    \item[1.] We feed the prompts "\textit{a photo of a single full-size, full-body \texttt{[obj]}, whole figure}", "\textit{a photo of a full-body \texttt{[obj]}}", "\textit{a high-resolution photo of a full-body \texttt{[obj]}}", "\textit{a DSLR photo of a full-body \texttt{[obj]}}" to Stable Diffusion~\citep{rombach2022high}, with \texttt{[obj]} indicating the desired class. We use Stable Diffusion v2.1 and generate images at resolution $512\times512$ using the DPMSolver~\citep{lu2022dpm} scheduler.
    \item[2.] We remove the background from the generated image, as we are interested in the object itself. To do so, we first use GroundingDINO~\citep{liu2023grounding} conditioned on the known class \texttt{[obj]} to obtain a bounding box of the object and FastSAM~\citep{zhao2023fast} to obtain the corresponding segmentation mask. Then, with the bounding box and segmentation mask, we isolate the object from the background and we center crop the image to have the object in the middle of it. Since we know the desired class beforehand, we exploit the rich semantics from text to segment the animals. This also allows us to build an automatic pipeline from \texttt{[obj]} to the final stroke representation without manual intervention.
    \item[3.]  We represent the foreground image as a sequence of strokes, using Stylized Neural Painting~\citep{zou2021stylized} for its flexibility. We set the parameters of SNP to have four layers of details and limit the number of strokes describing each image to 400.
\end{itemize}

Note that our pipeline follows a modular design, where each module can be flexibly updated. Our current implementation is composed of off-the-shelf components that represents the state of the art in order to achieve the best quality dataset. 
Moreover, our data-generation pipeline is fully automatic and class agnostic. Thus, it can readily include more object categories or increase the dataset size and diversity. 
In this work, we choose to generate data of 10 animal classes: \textit{cat, dog, eagle, elephant, lion, parrot, rabbit, squirrel, tiger} and \textit{wolf}. They are easily recognizable with great inter and intra-class diversity, thus allowing for objective and subjective evaluation with less ambiguity to kick off the research. 
We generate at least 10K images for each class, with 500 left out for testing. There are a total of 101.052 decomposed sequences for the whole dataset. We show some examples in Fig.~\ref{fig:dataset} while we show examples of generated and segmented images in Fig.~\ref{fig:pipeline}.

\subsection{Evaluation Protocol}
\label{sec:setting}
\begingroup

\renewcommand{\arraystretch}{0.} 
\setlength\tabcolsep{0.pt}

\begin{figure}[!t]
\resizebox{1.\linewidth}{!}{%
\begin{tabular}{ccccc}
\includegraphics[width=0.2\textwidth]{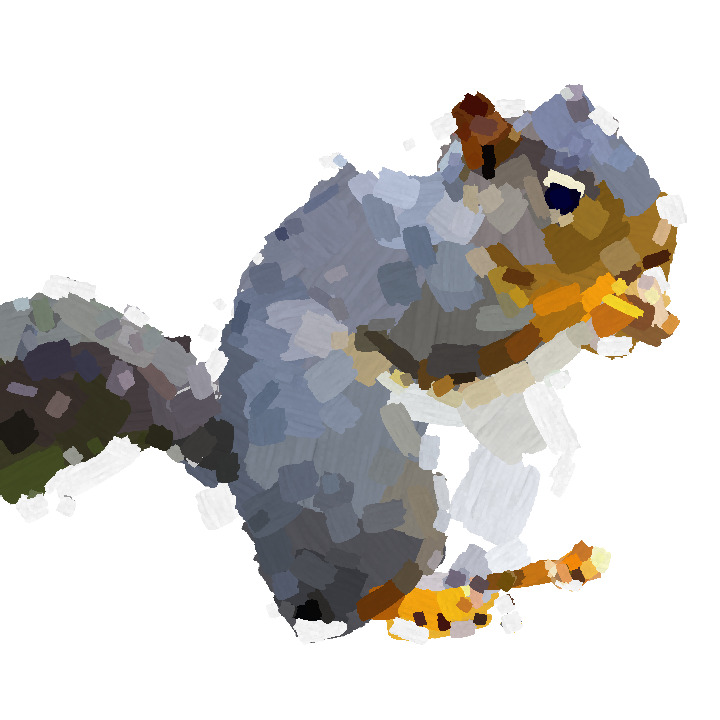} &
\includegraphics[width=0.2\textwidth]{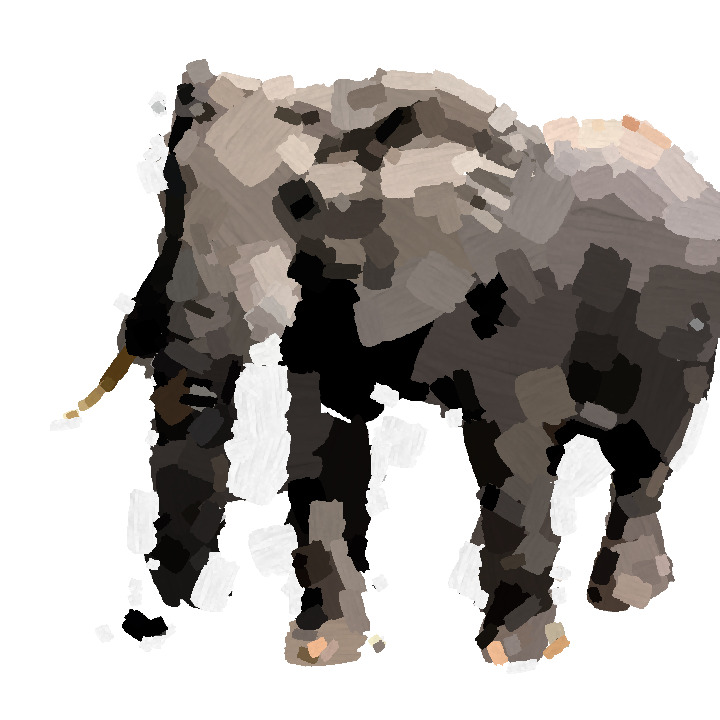} &
\includegraphics[width=0.2\textwidth]{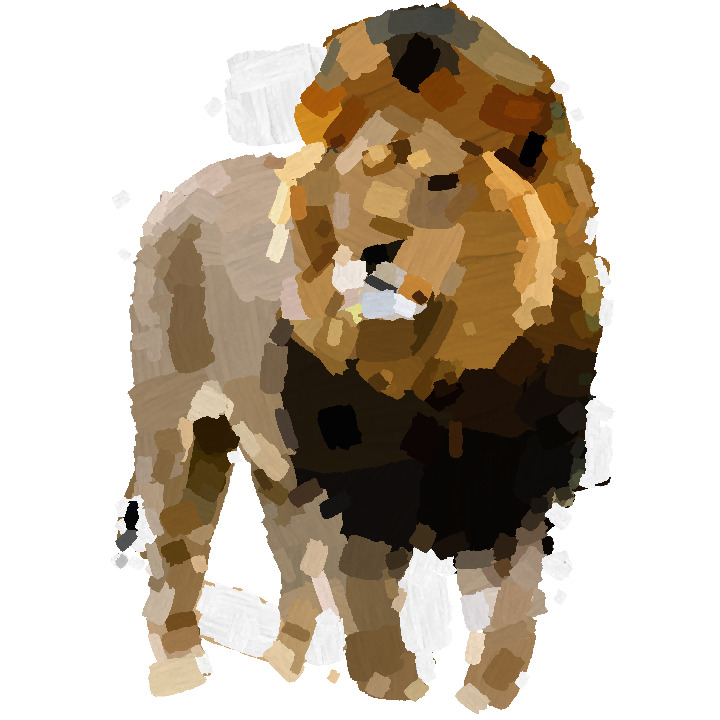} &
\includegraphics[width=0.2\textwidth]{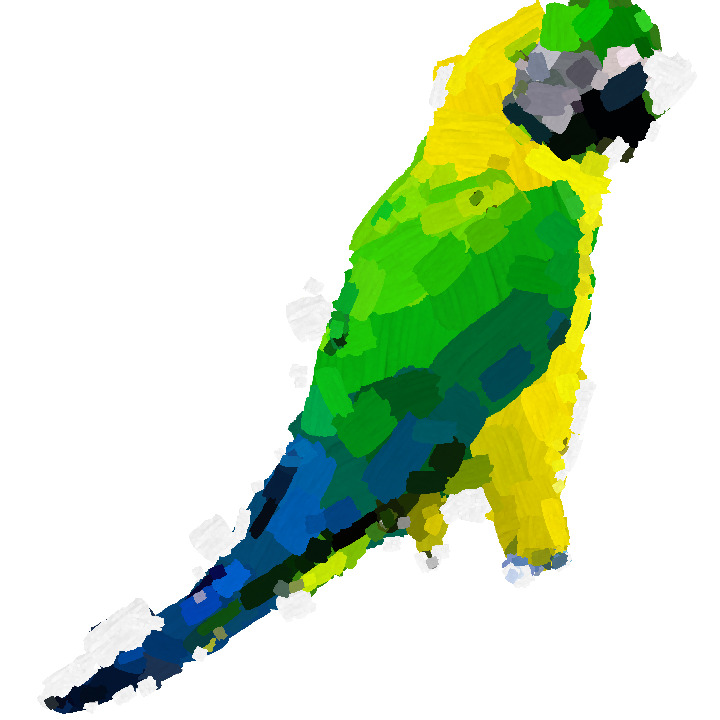} &
\includegraphics[width=0.2\textwidth]{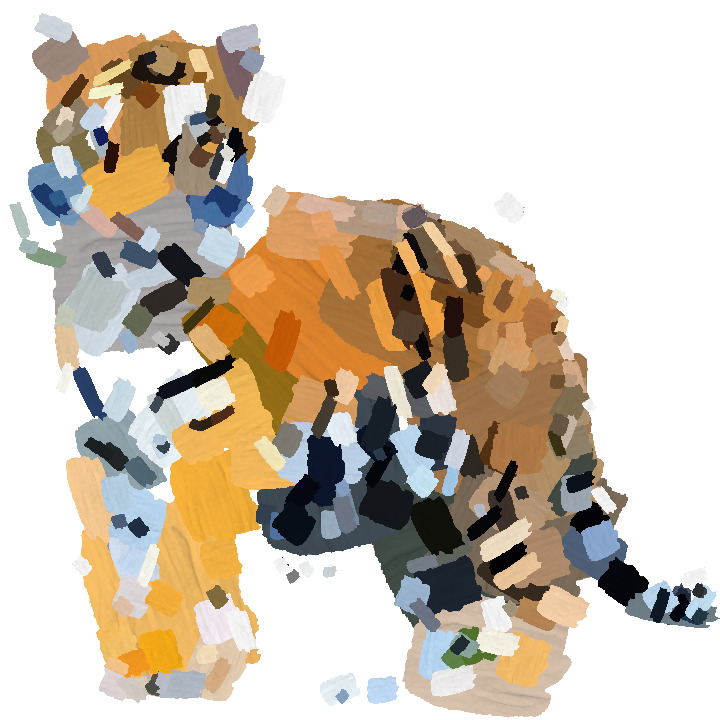} \\

\includegraphics[width=0.2\textwidth]{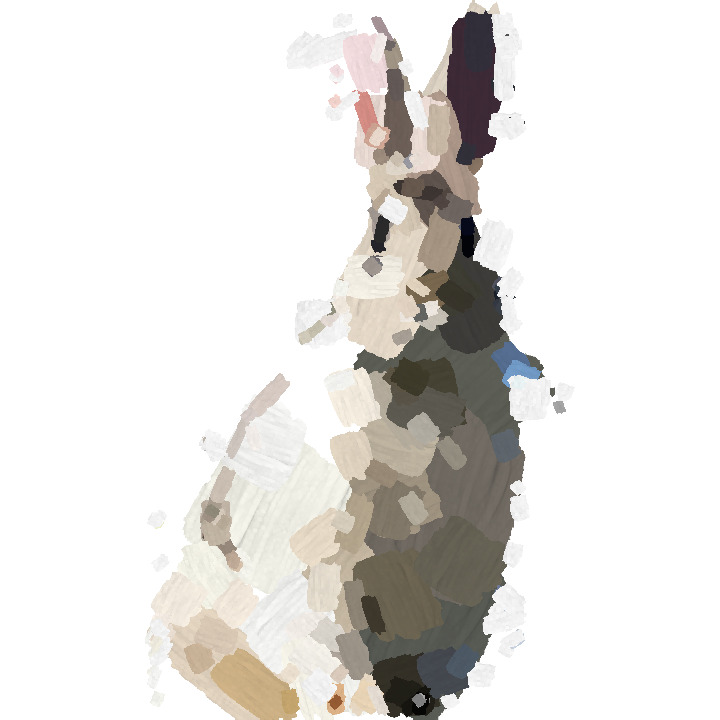} &
\includegraphics[width=0.2\textwidth]{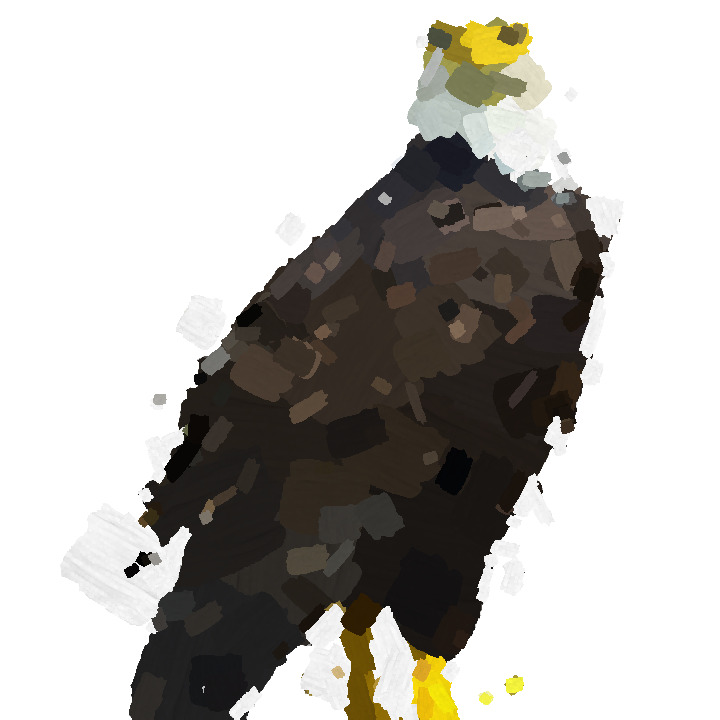} &
\includegraphics[width=0.2\textwidth]{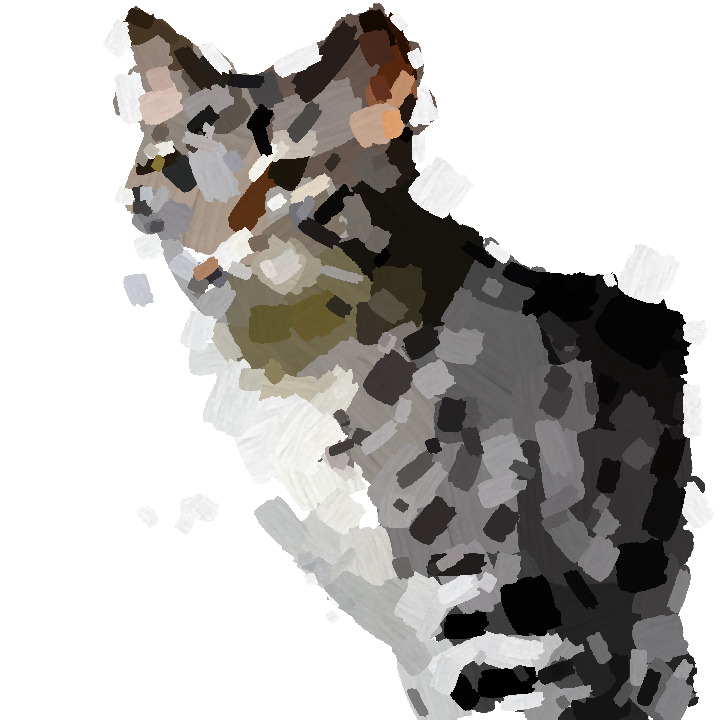} &
\includegraphics[width=0.2\textwidth]{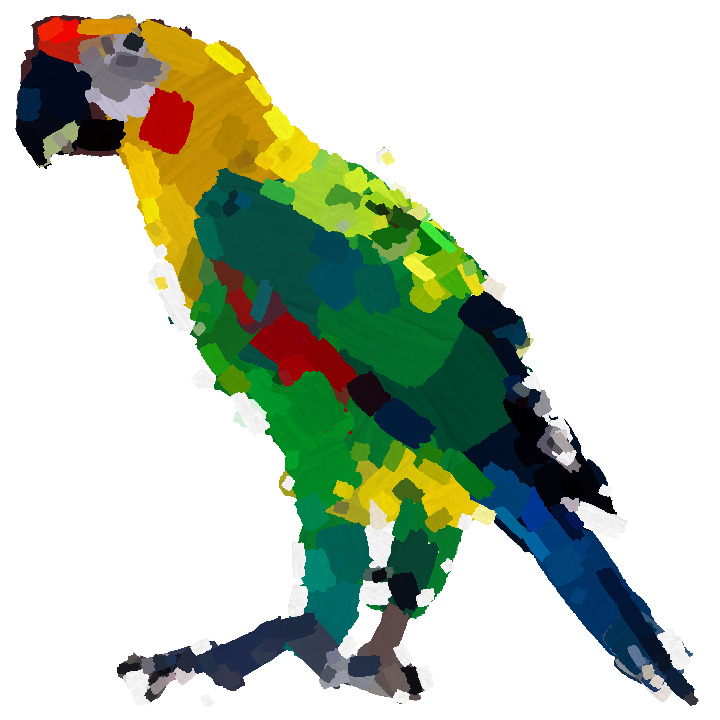} &
\includegraphics[width=0.2\textwidth]{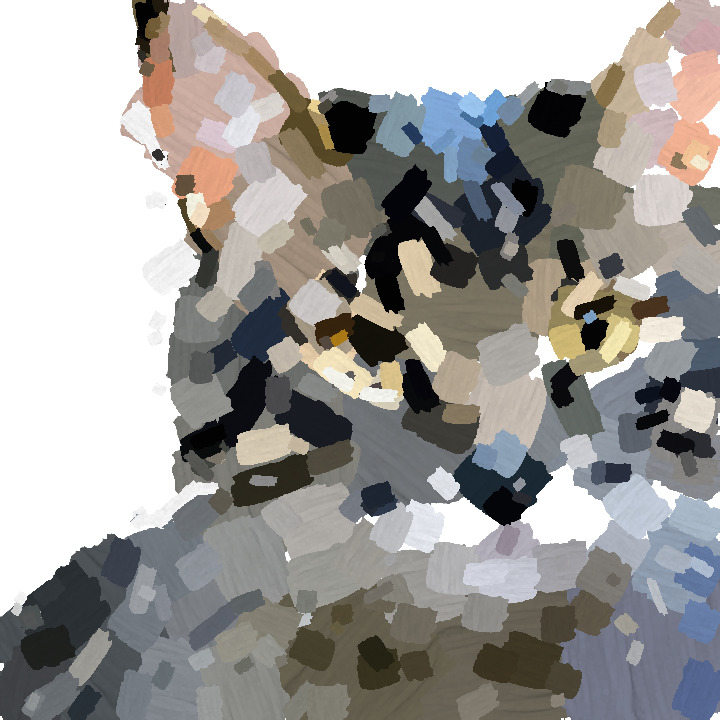} \\

\includegraphics[width=0.2\textwidth]{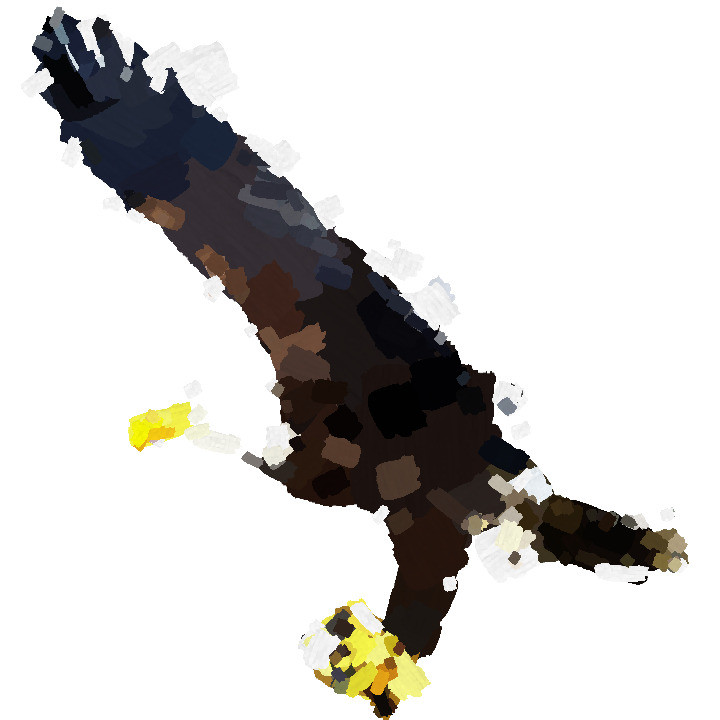} &
\includegraphics[width=0.2\textwidth]{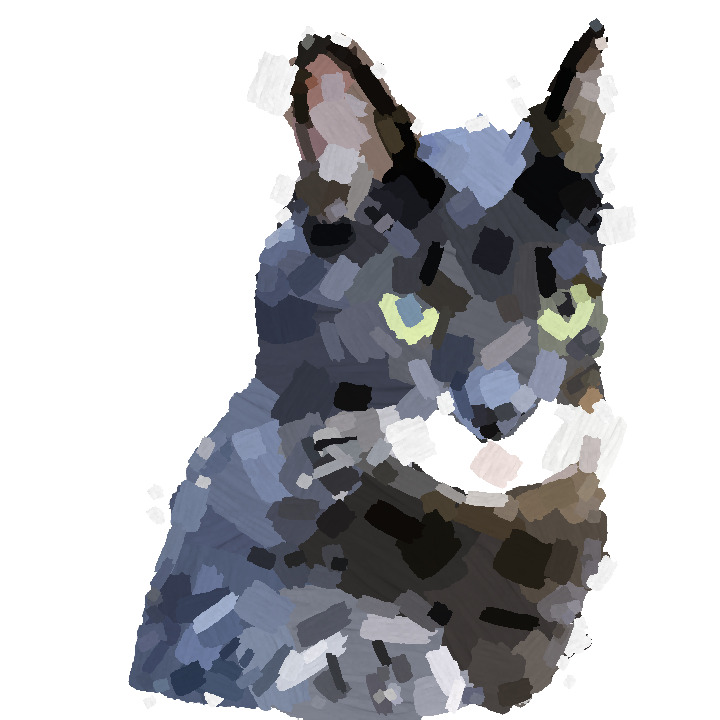} &
\includegraphics[width=0.2\textwidth]{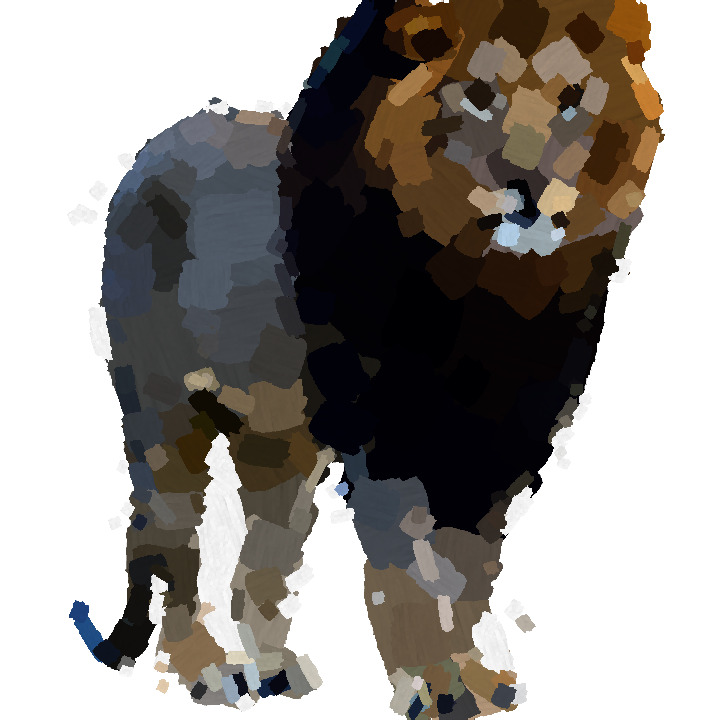} &
\includegraphics[width=0.2\textwidth]{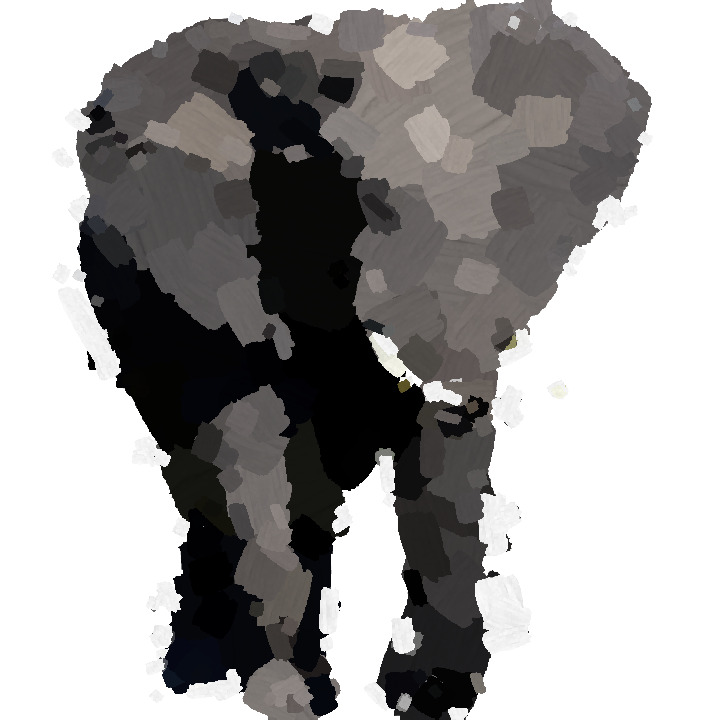} &
\includegraphics[width=0.2\textwidth]{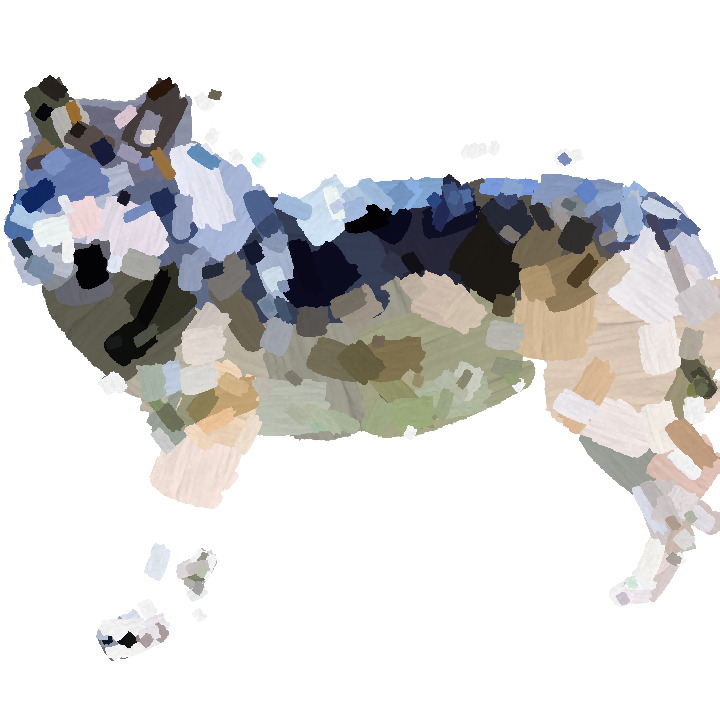} \\

\includegraphics[width=0.2\textwidth]{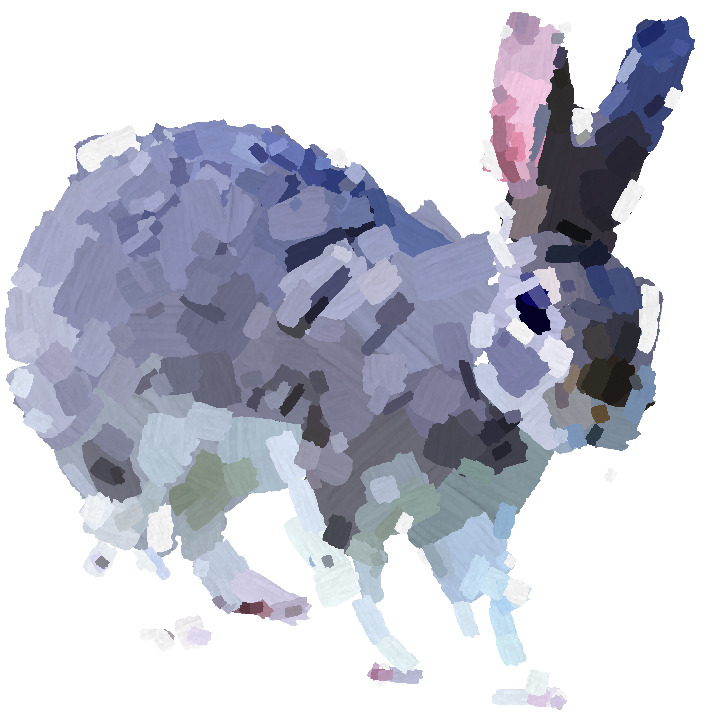} &
\includegraphics[width=0.2\textwidth]{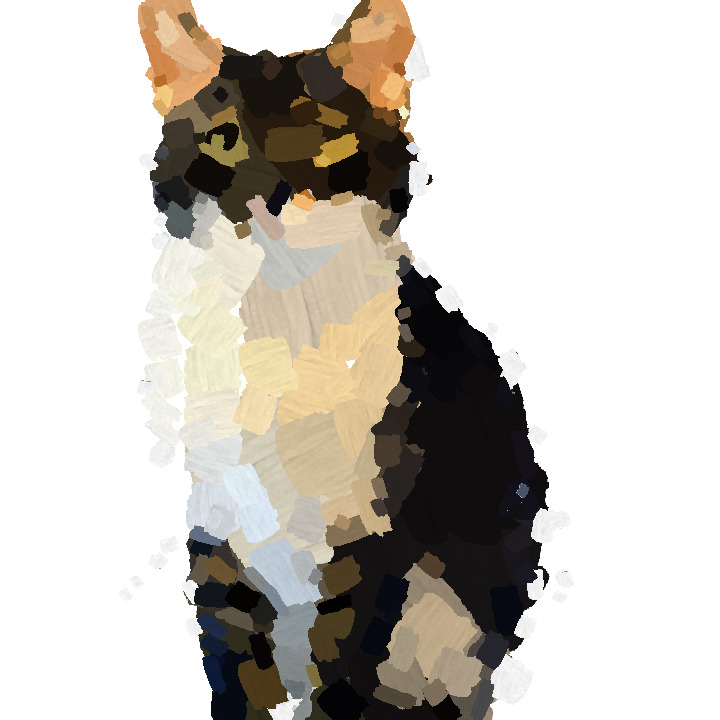} &
\includegraphics[width=0.2\textwidth]{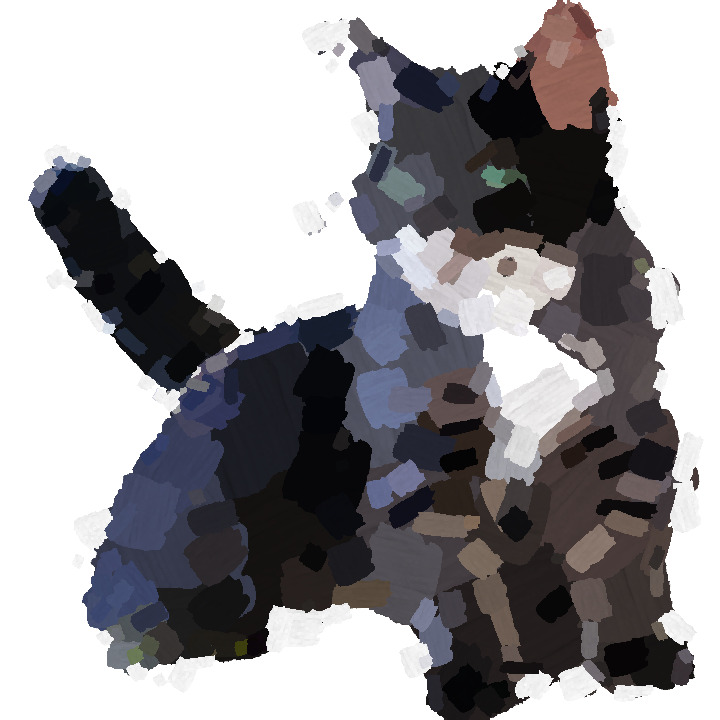} &
\includegraphics[width=0.2\textwidth]{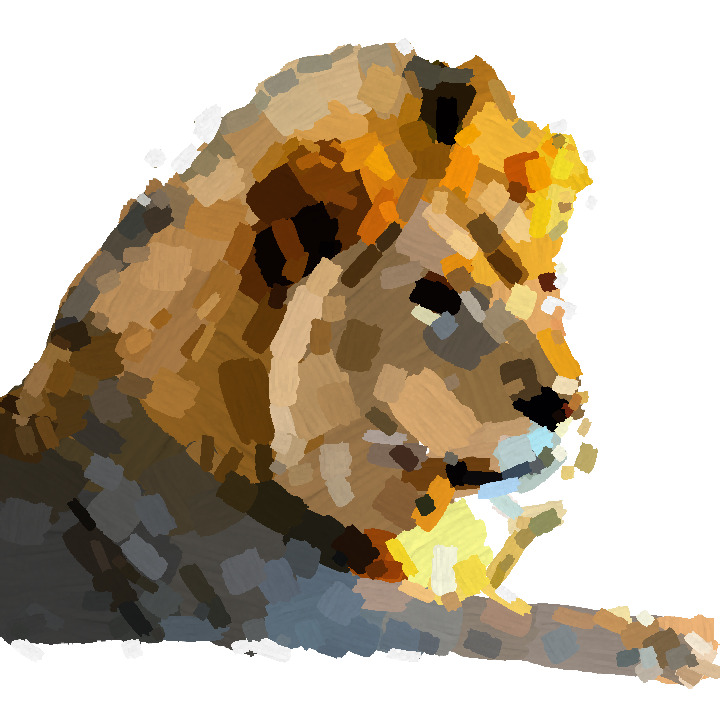} &
\includegraphics[width=0.2\textwidth]{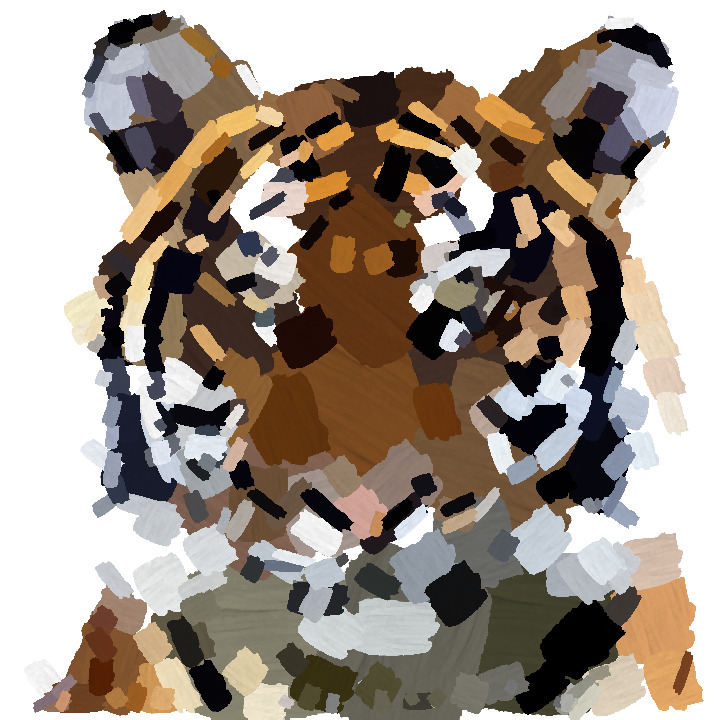} \\

\includegraphics[width=0.2\textwidth]{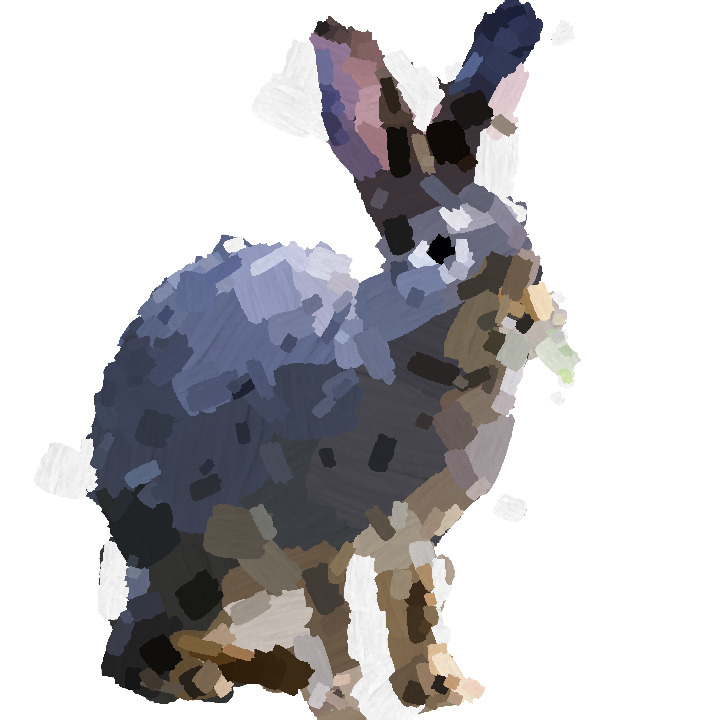} &
\includegraphics[width=0.2\textwidth]{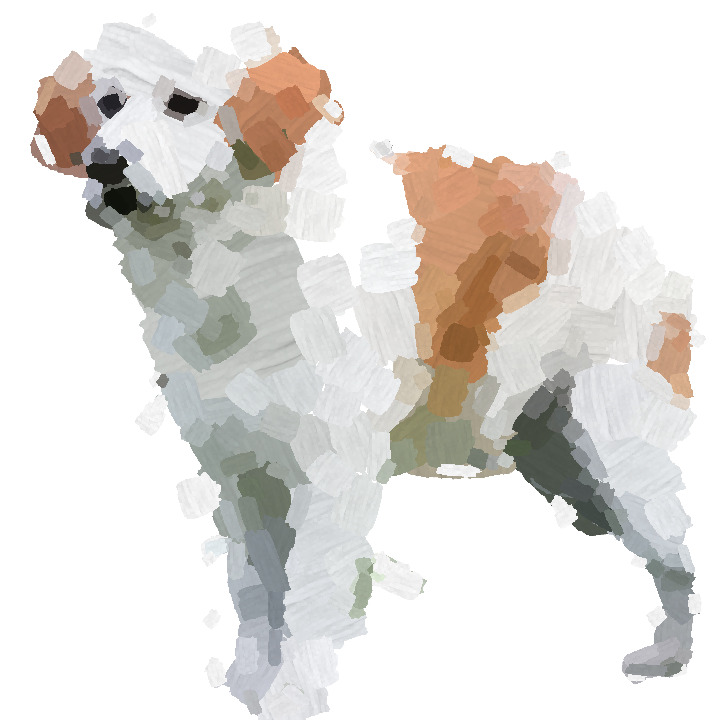} &
\includegraphics[width=0.2\textwidth]{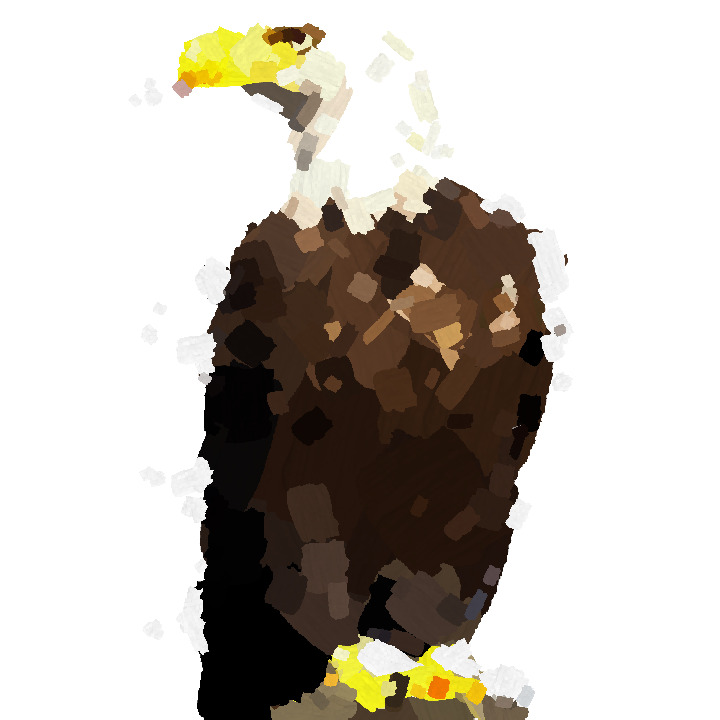} &
\includegraphics[width=0.2\textwidth]{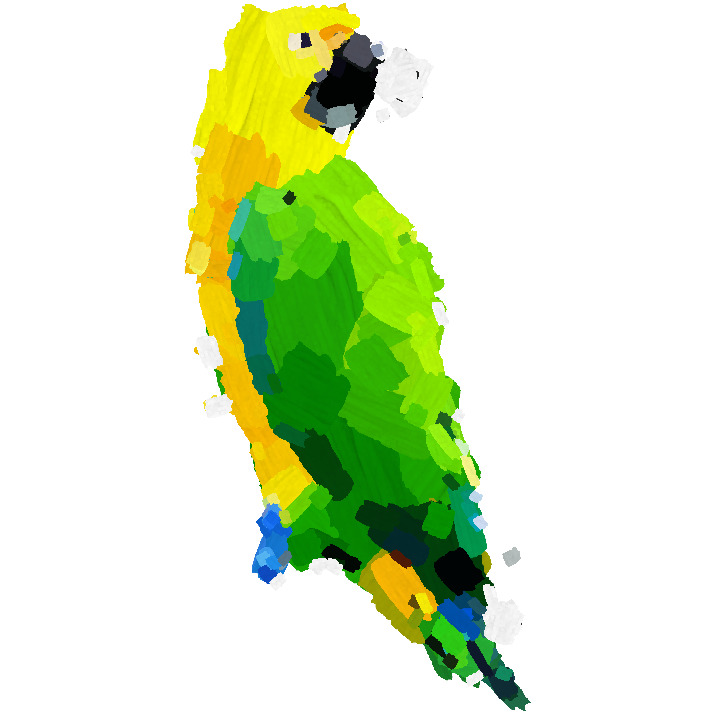} &
\includegraphics[width=0.2\textwidth]{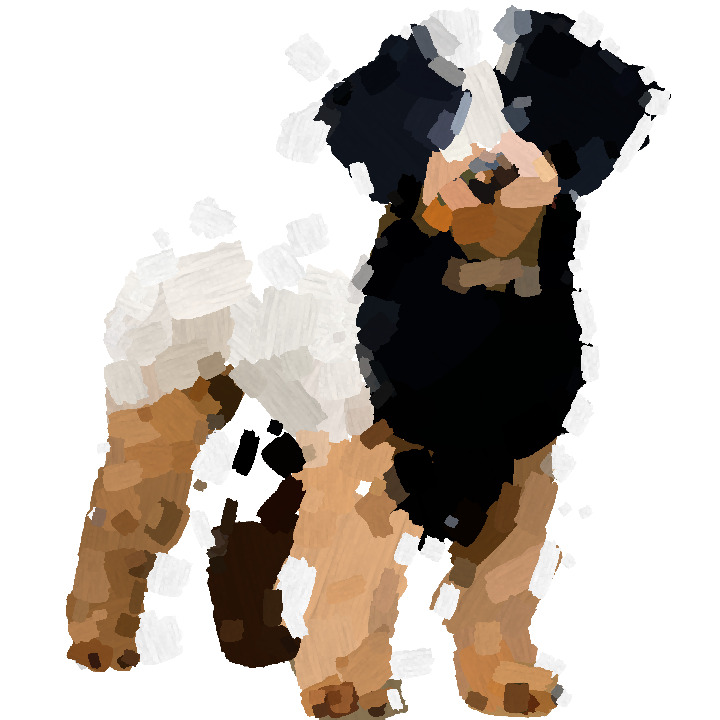} \\
\end{tabular}
}
\caption{Examples of rendered sequences from the dataset.}
\label{fig:dataset}
\vspace{-0.4cm}
\end{figure}

\endgroup

Existing benchmarks for image synthesis are inadequate to capture the  interactive nature of the \methodshort~task.
In Sec.~\ref{sec:masking} we discussed several masking strategies, namely \textit{block, level, random, no context}, and \textit{square}, designed to capture multiple interactive scenarios. We use those masking strategies as tasks to mimic, in an automatic way, the interaction between the user and the painting agent, and use them for evaluation. 

\noindent\textbf{Metrics.}
\label{sec:metrics}
We introduce a set of metrics to quantitatively evaluate the proposed methods. First, we render the predicted and context strokes $\tilde{\boldsymbol{s}}_0$, and compare it with the rendered test set. Following the image inpainting literature~\citep{zheng2022image,liu2019coherent}, we compute the Fréchet Inception Distance (\textbf{FID})~\citep{heusel2017gans}, and a sample-wise $\boldsymbol{\mathcal{L}_2}$ distance. 
Second, we design a metric specific for the stroke-based formulation and evaluate the similarity between predicted parameters and ground truth strokes computing a $\boldsymbol{\mathcal{L}_1}$ distance. 
Although ordering is crucial for determining the rendering priority of overlapping strokes and subsequent levels, non-overlapping strokes in the same granularity level can be swapped without affecting the result. For this reason, before computing the distance, we perform a Hungarian Matching~\citep{kuhn1955hungarian}.

We compute the metrics for each task separately to gauge their respective difficulty. In practice, we sample $\sim$5K sequences from the test set and create the conditioning sequence $\vecsequencecond$ by applying the masking corresponding to the given task.
Note that for the \textit{no context} case, we cannot compute sample-wise metrics, thus we rely only on the FID.
\section{Experiments}
\label{sec:exp}

\begingroup
\renewcommand{\arraystretch}{0.} 

\begin{figure*}[!ht]
\resizebox{1.\linewidth}{!}{%
\setlength\tabcolsep{0.pt}
\begin{tabular}{cc|cc|cc|cc|c}
\toprule

\multicolumn{2}{c|}{\textbf{Level}} & \multicolumn{2}{c|}{\textbf{Square}} & \multicolumn{2}{c|}{\textbf{Random}} & \multicolumn{2}{c|}{\textbf{Block}} & \multicolumn{1}{c}{\textbf{No context}} \\ 
\midrule
\multicolumn{1}{c|}{$\vecsequencecond$} & \multicolumn{1}{c|}{$\tilde{\boldsymbol{s}}_0$} & \multicolumn{1}{c|}{$\vecsequencecond$} & \multicolumn{1}{c|}{$\tilde{\boldsymbol{s}}_0$} & \multicolumn{1}{c|}{$\vecsequencecond$} & \multicolumn{1}{c|}{$\tilde{\boldsymbol{s}}_0$} & \multicolumn{1}{c|}{$\vecsequencecond$} & \multicolumn{1}{c|}{$\tilde{\boldsymbol{s}}_0$} & \multicolumn{1}{c}{$\tilde{\boldsymbol{s}}_0$} \\
\midrule

\includegraphics[width=0.1\textwidth]{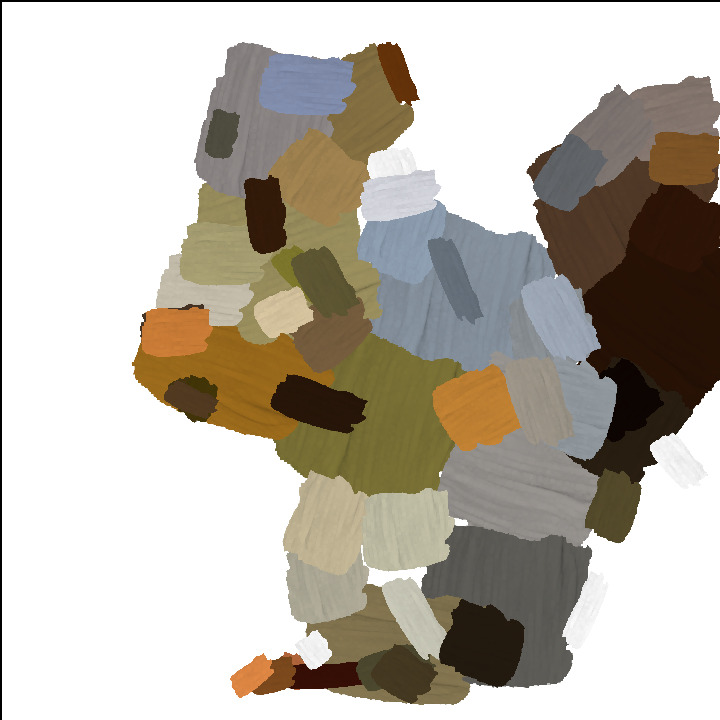} &
\includegraphics[width=0.1\textwidth]{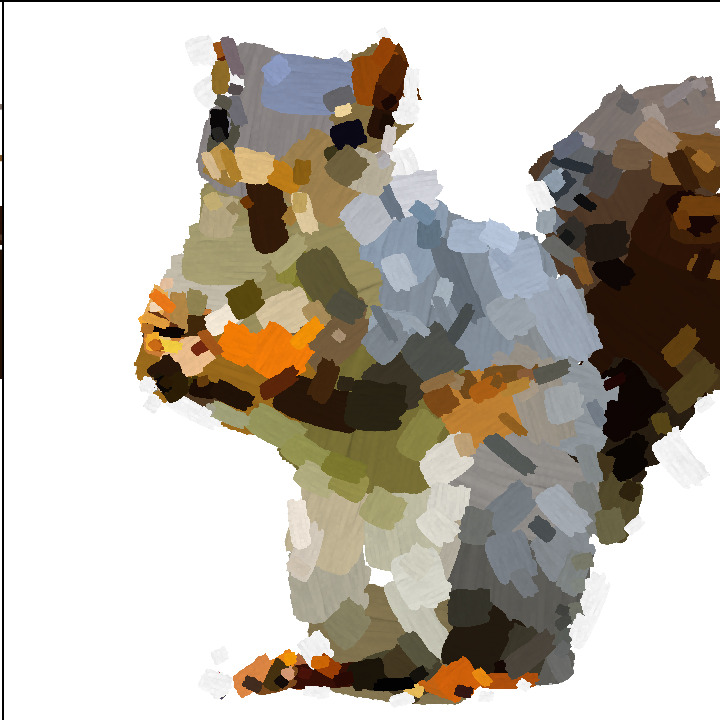} &
\includegraphics[width=0.1\textwidth]{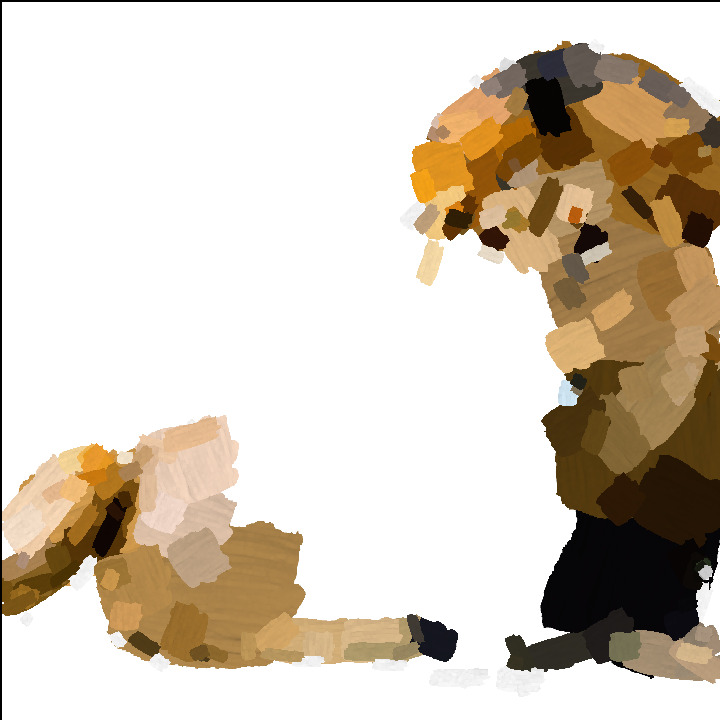} &
\includegraphics[width=0.1\textwidth]{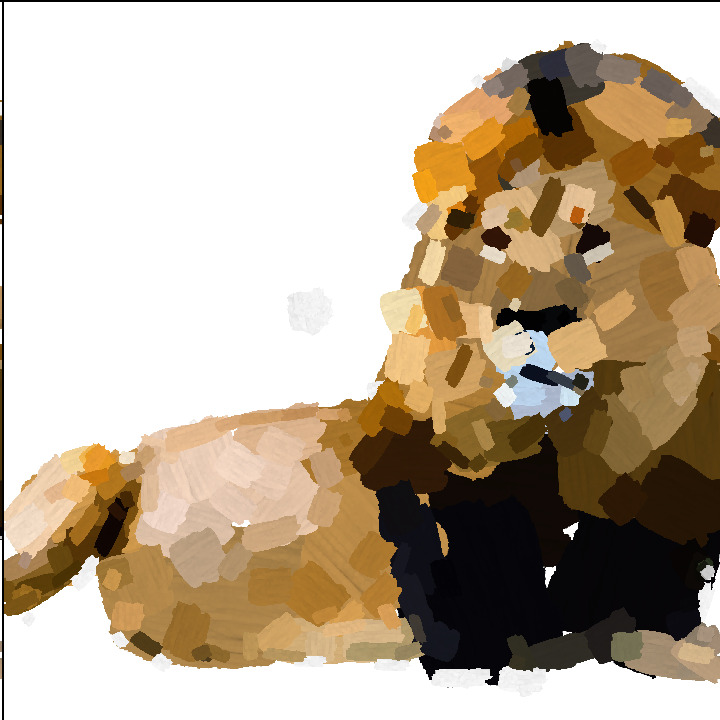} & 
\includegraphics[width=0.1\textwidth]{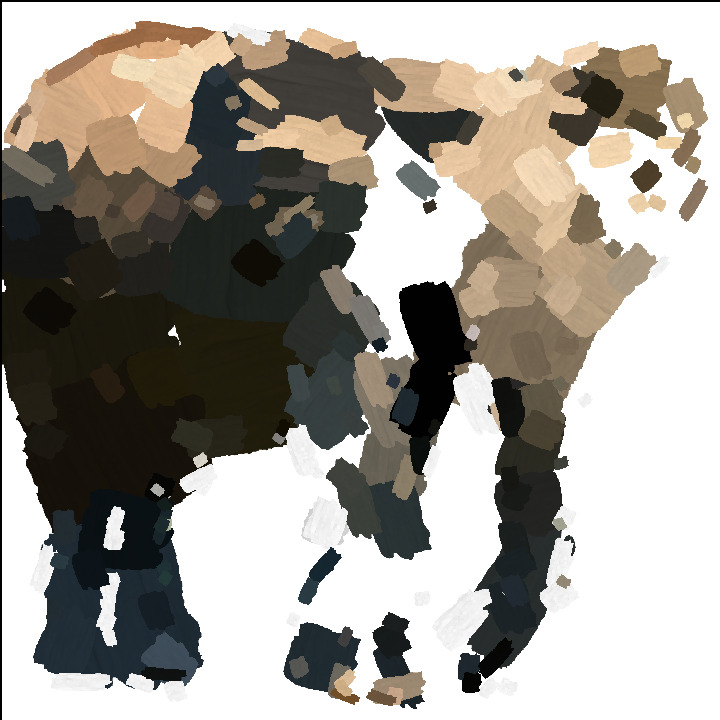} &
\includegraphics[width=0.1\textwidth]{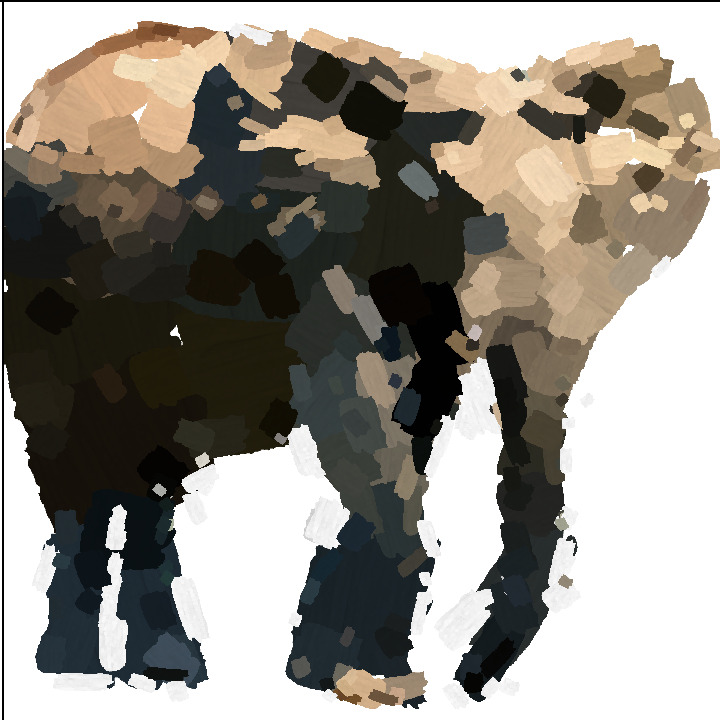} &
\includegraphics[width=0.1\textwidth]{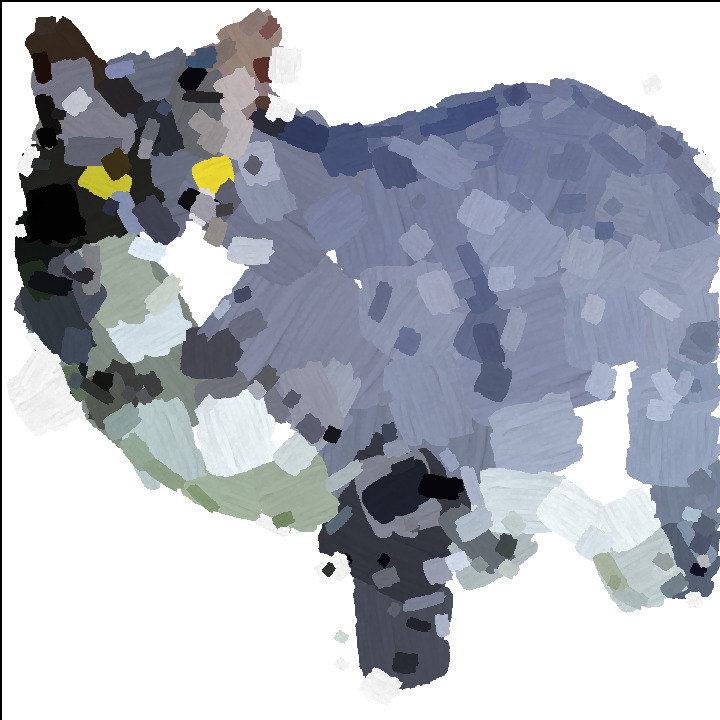} &
\includegraphics[width=0.1\textwidth]{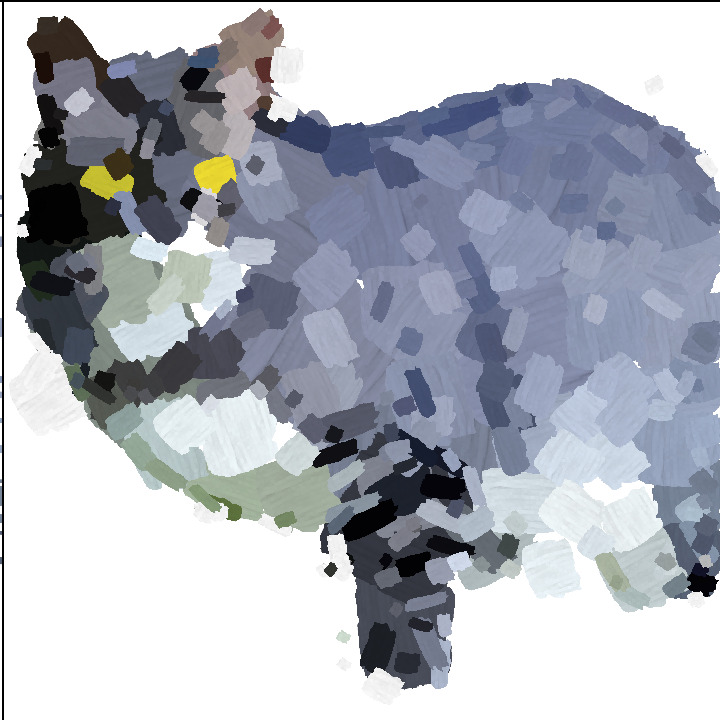} &

\includegraphics[width=0.1\textwidth]{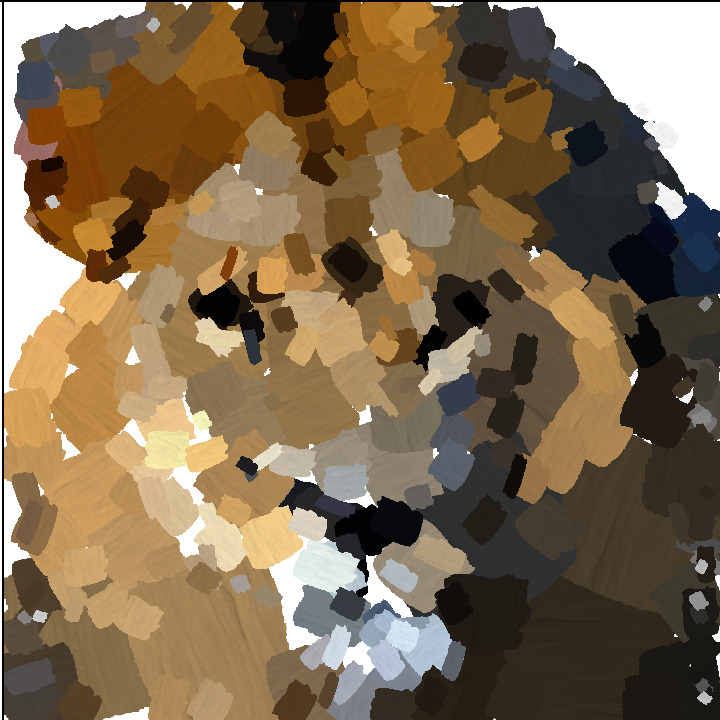} \\

\includegraphics[width=0.1\textwidth]{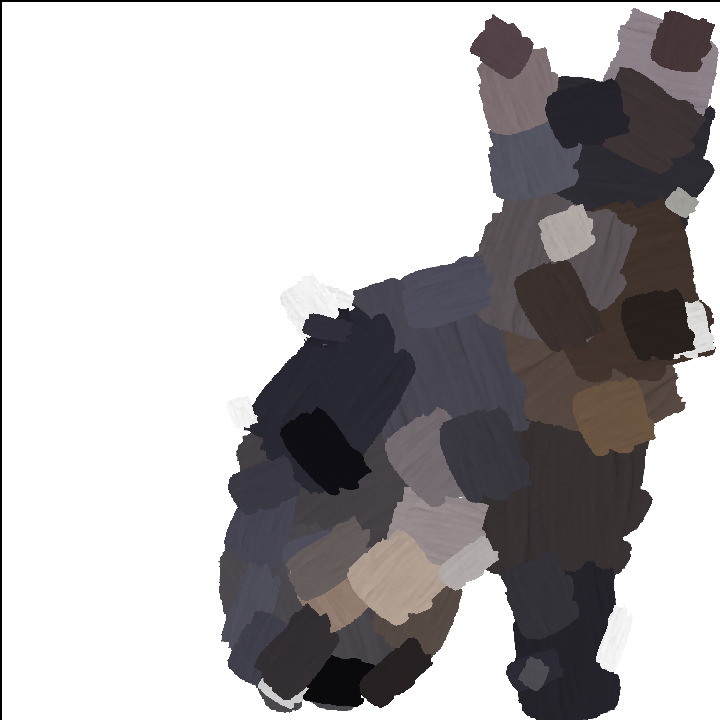} &
\includegraphics[width=0.1\textwidth]{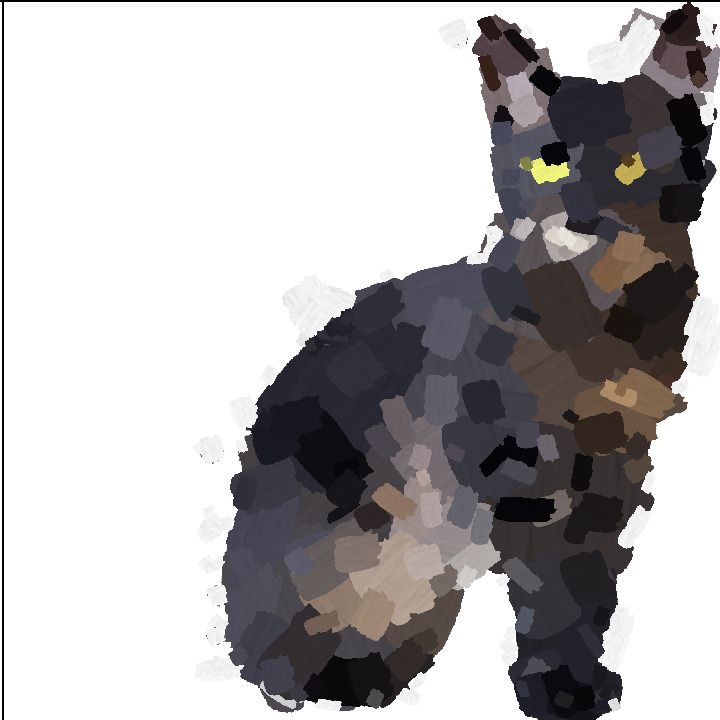} &
\includegraphics[width=0.1\textwidth]{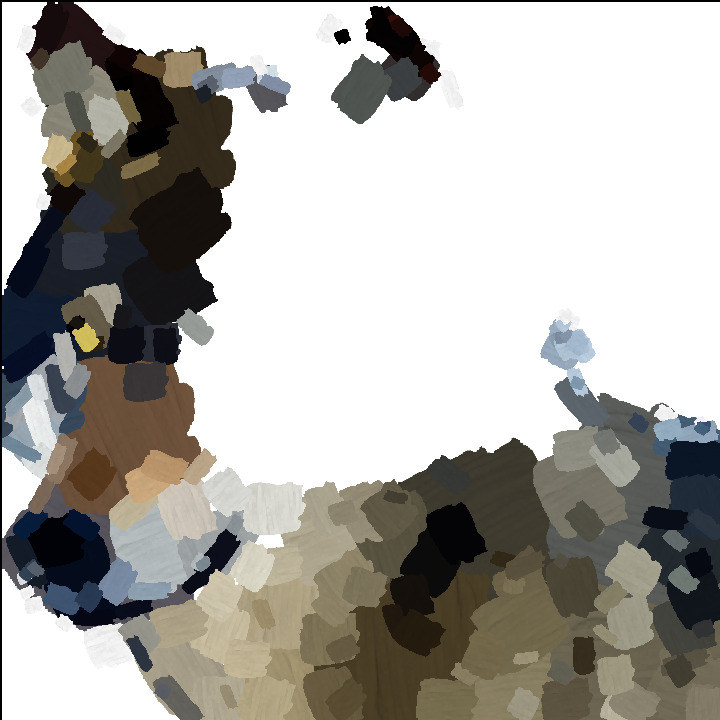} &
\includegraphics[width=0.1\textwidth]{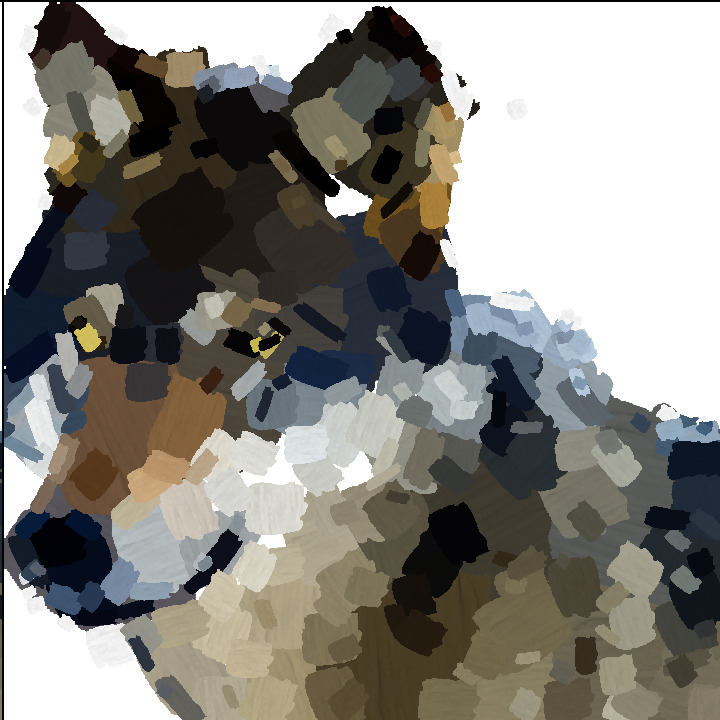} & 
\includegraphics[width=0.1\textwidth]{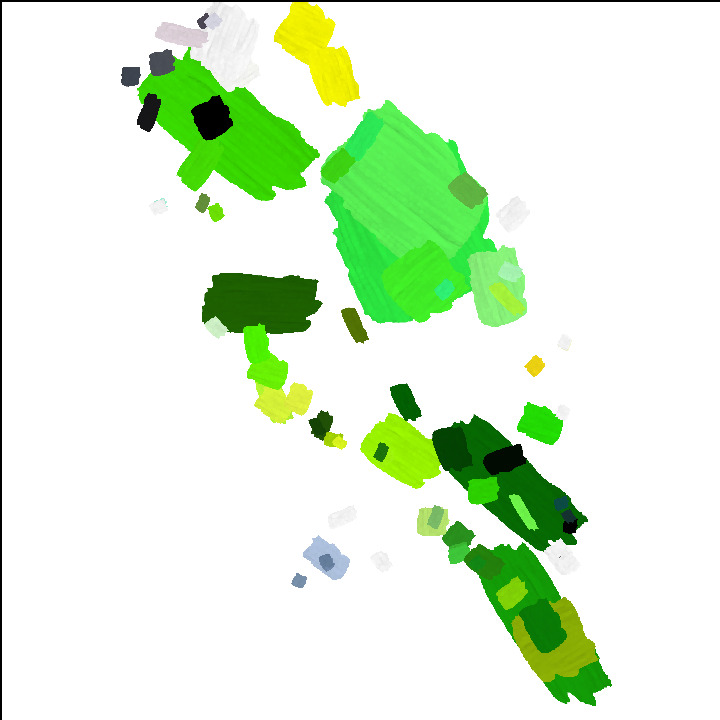} &
\includegraphics[width=0.1\textwidth]{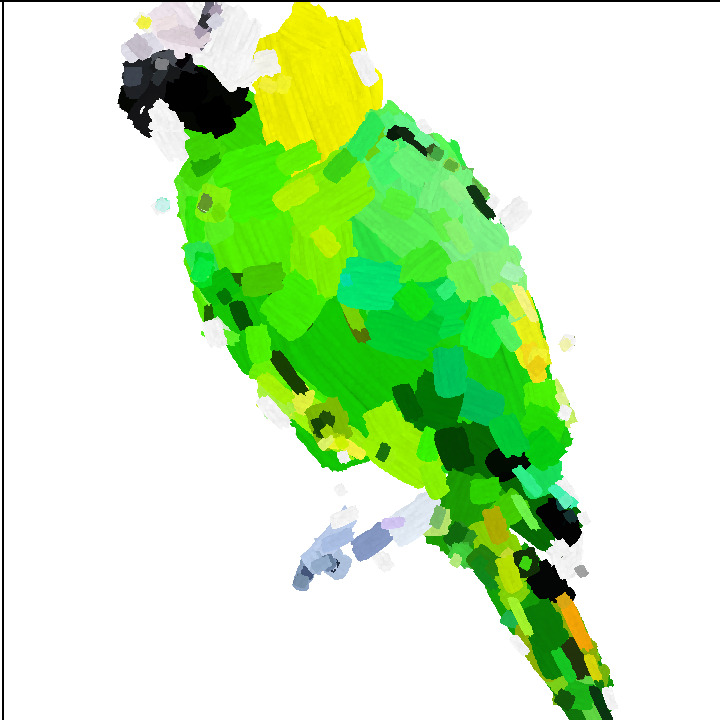} &
\includegraphics[width=0.1\textwidth]{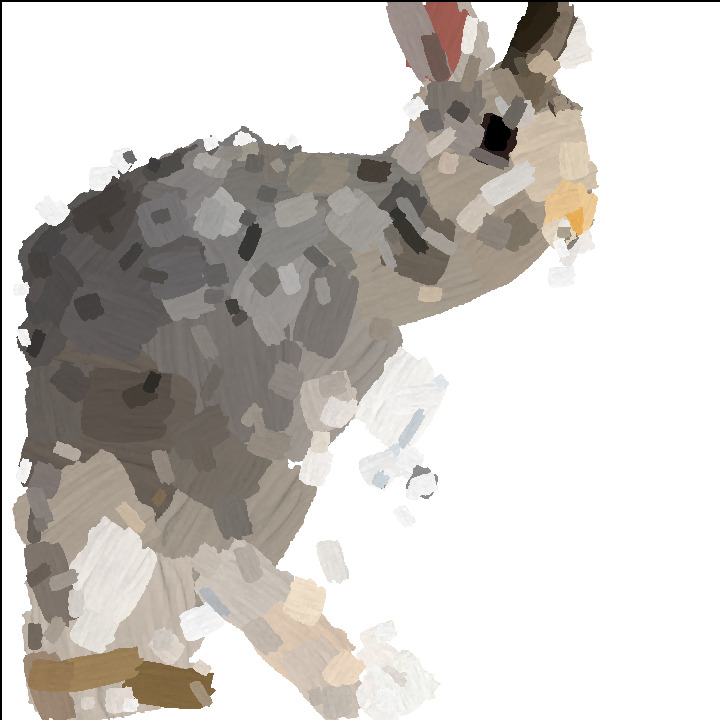} &
\includegraphics[width=0.1\textwidth]{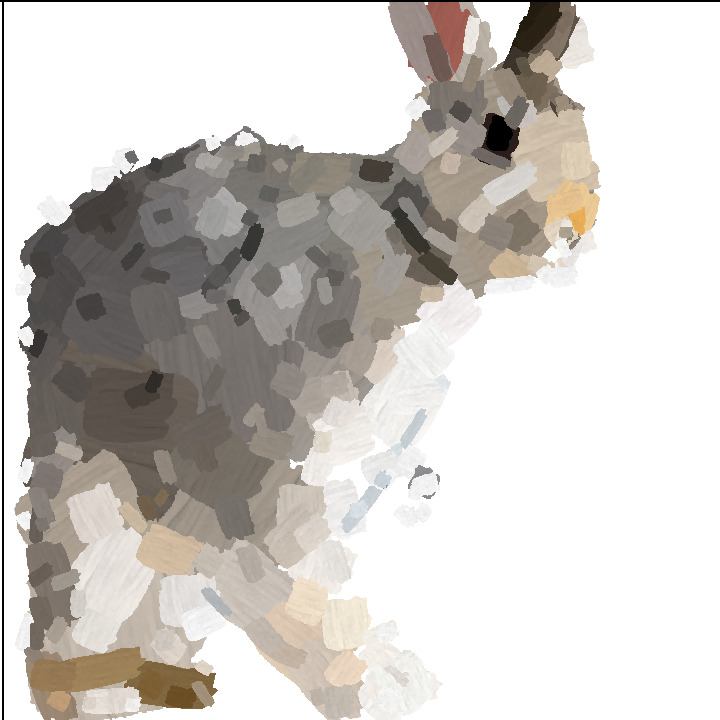} &

\includegraphics[width=0.1\textwidth]{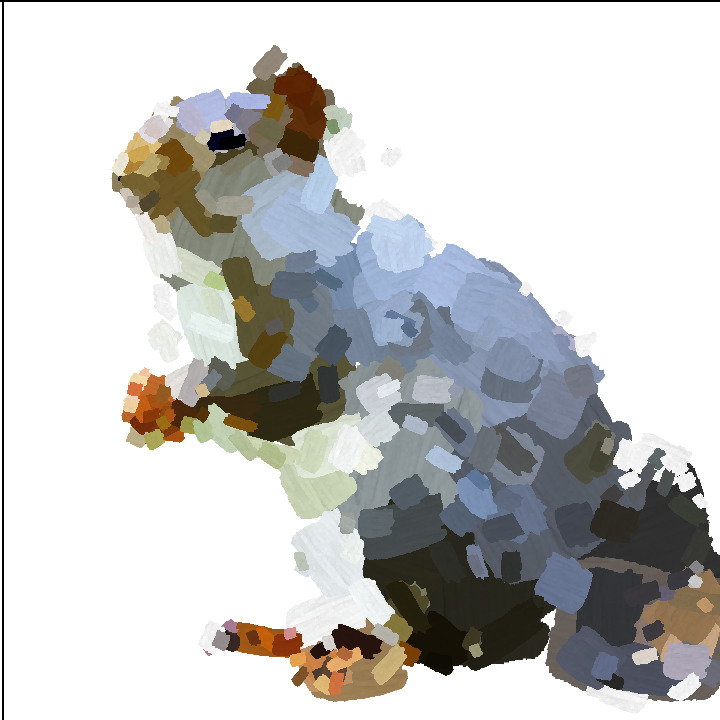} \\

\includegraphics[width=0.1\textwidth]{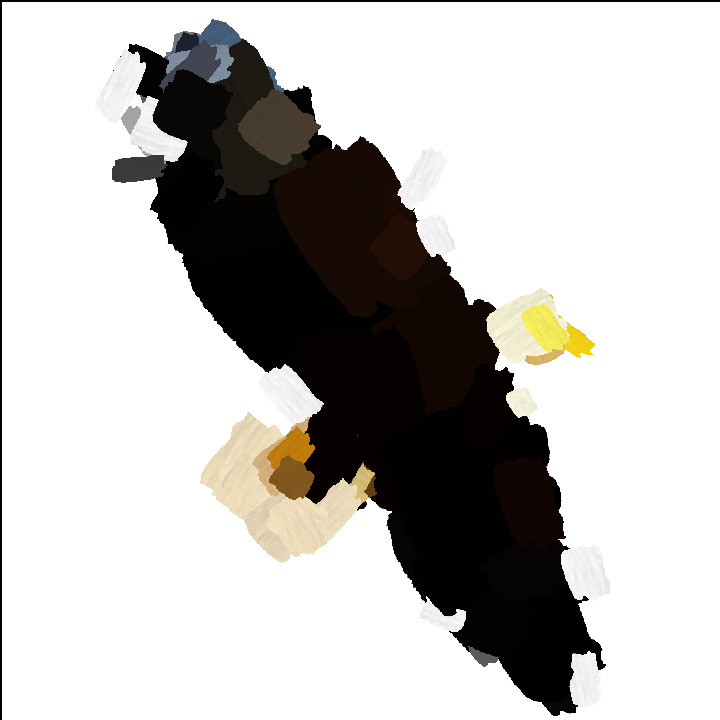} &
\includegraphics[width=0.1\textwidth]{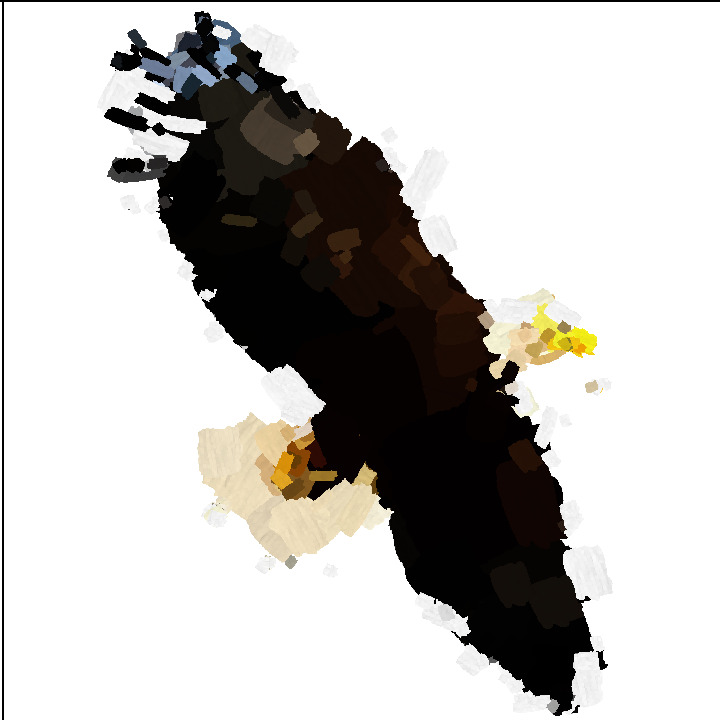} &
\includegraphics[width=0.1\textwidth]{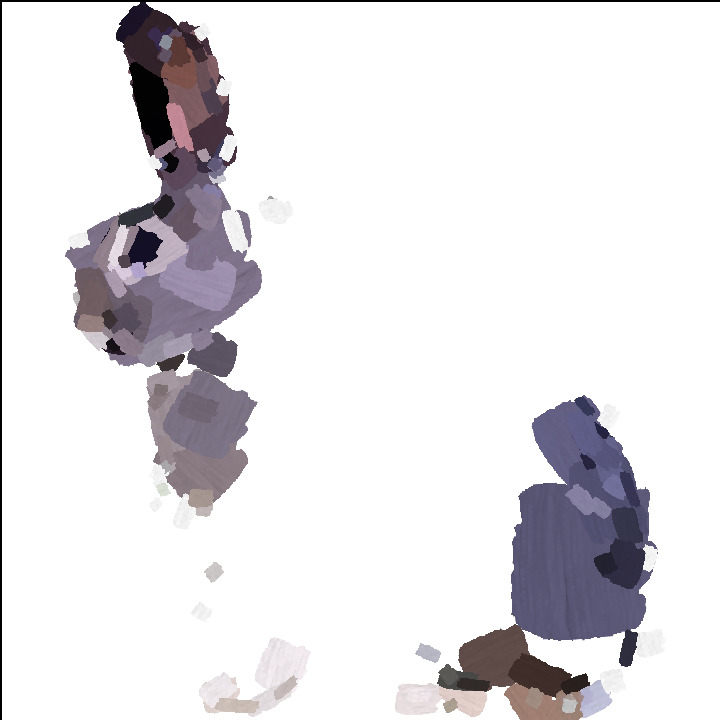} &
\includegraphics[width=0.1\textwidth]{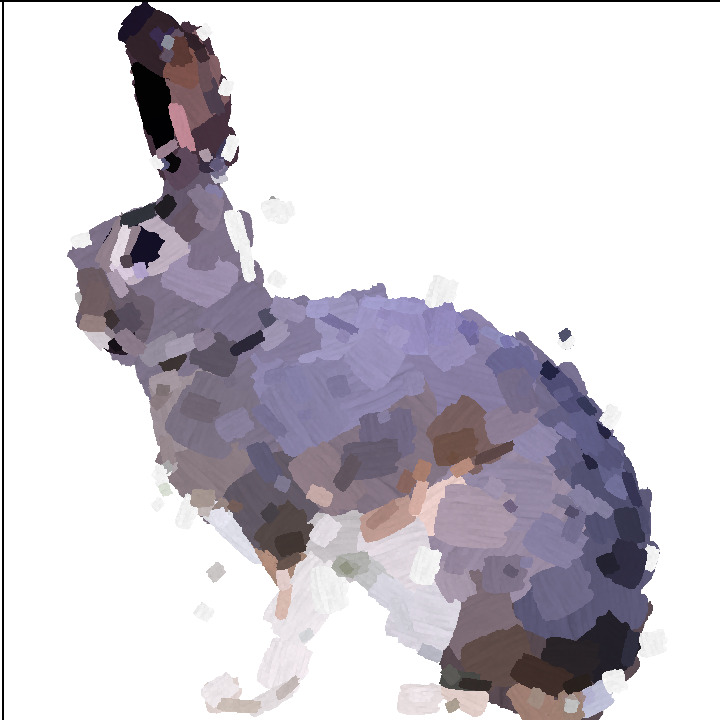} & 
\includegraphics[width=0.1\textwidth]{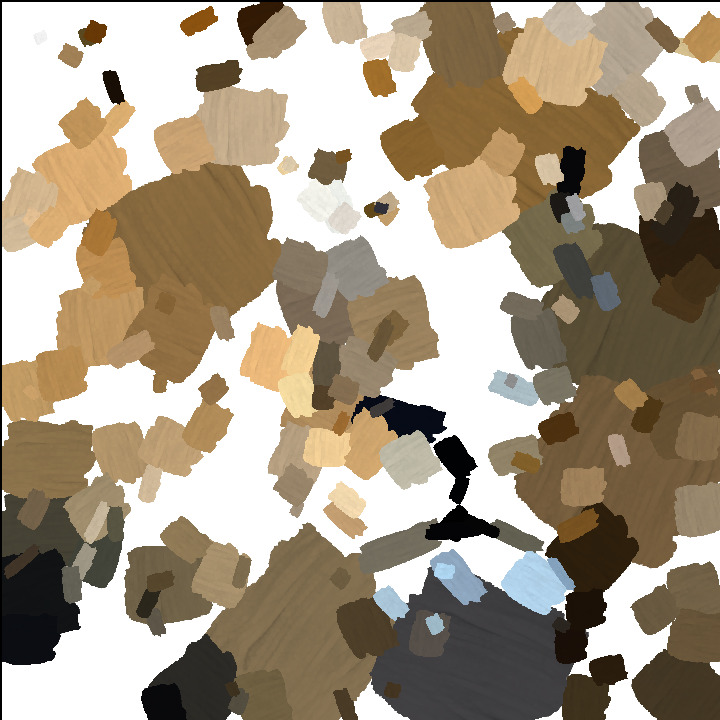} &
\includegraphics[width=0.1\textwidth]{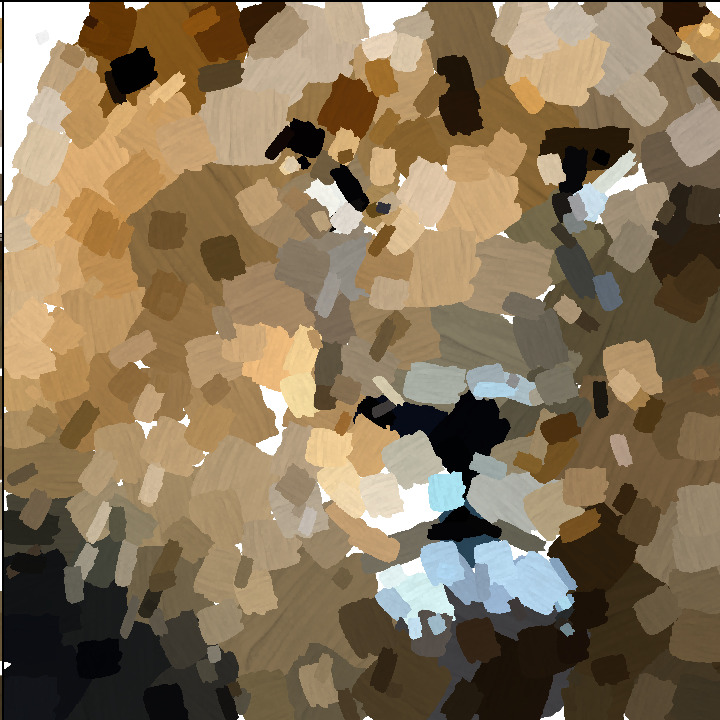} &
\includegraphics[width=0.1\textwidth]{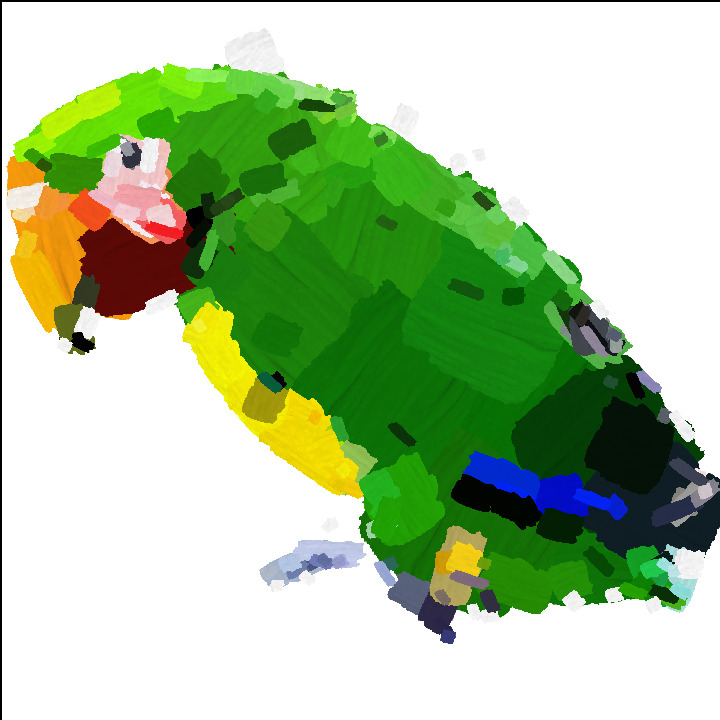} &
\includegraphics[width=0.1\textwidth]{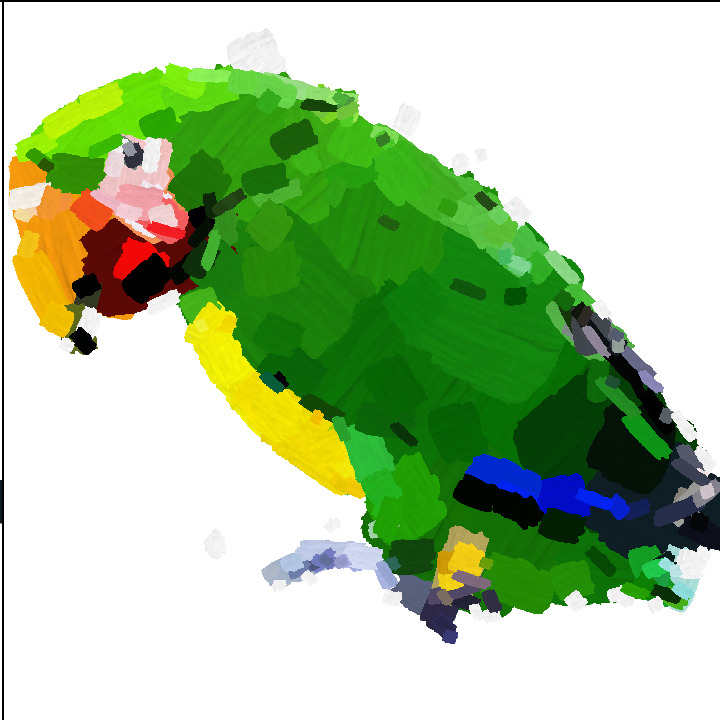} &

\includegraphics[width=0.1\textwidth]{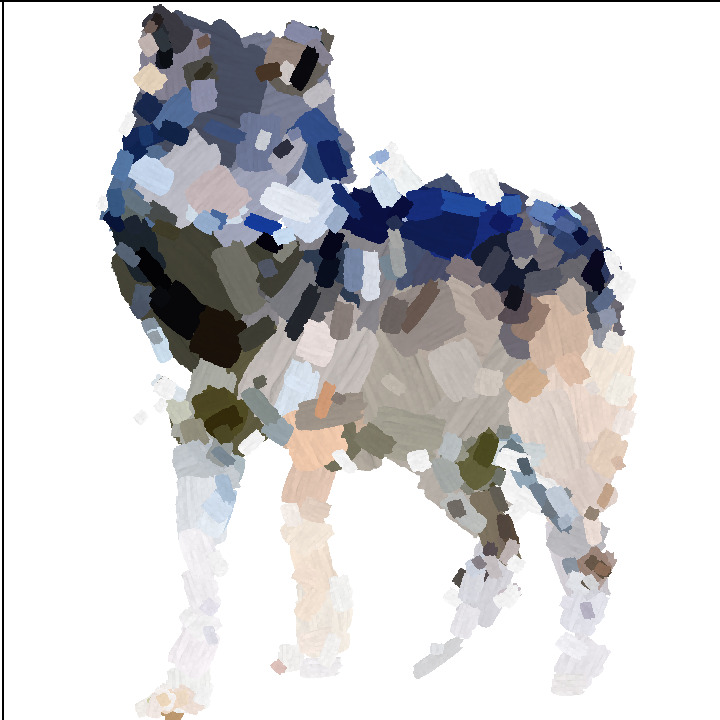} \\

\includegraphics[width=0.1\textwidth]{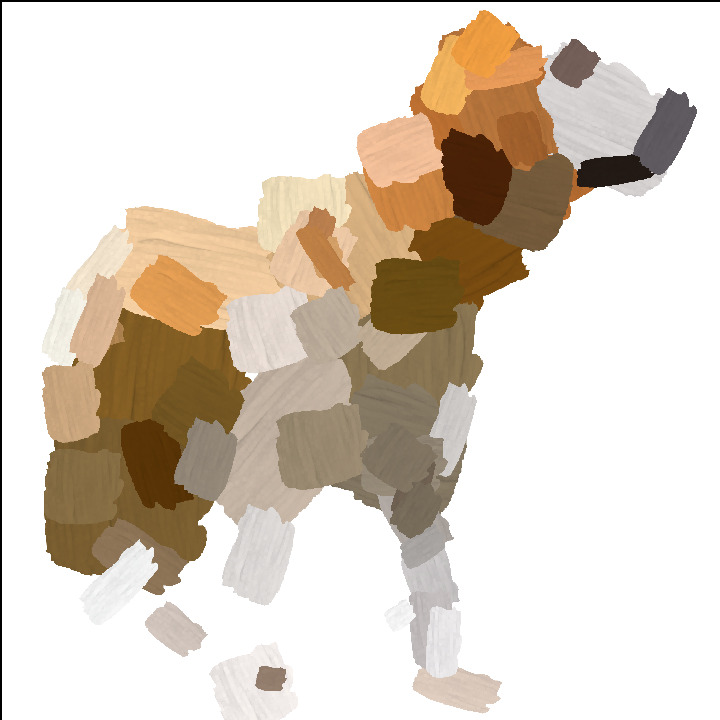} &
\includegraphics[width=0.1\textwidth]{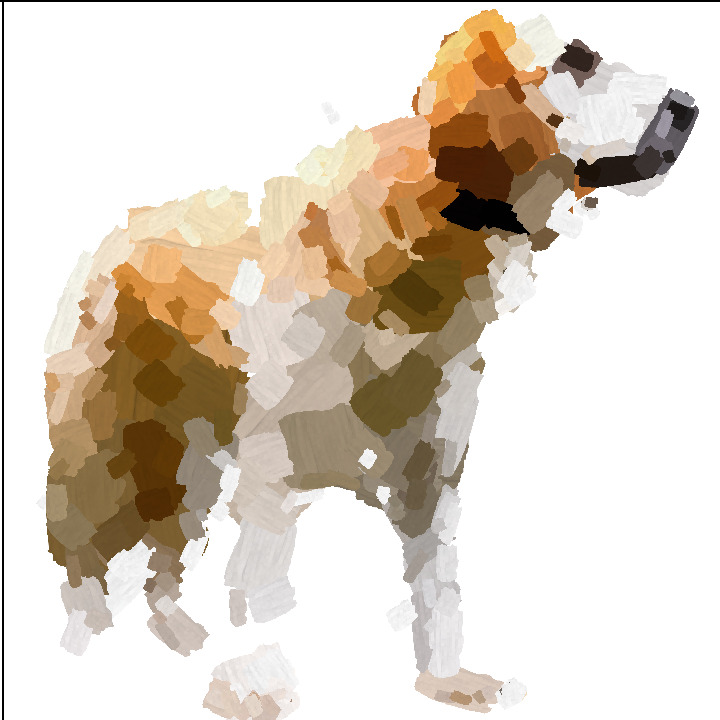} &
\includegraphics[width=0.1\textwidth]{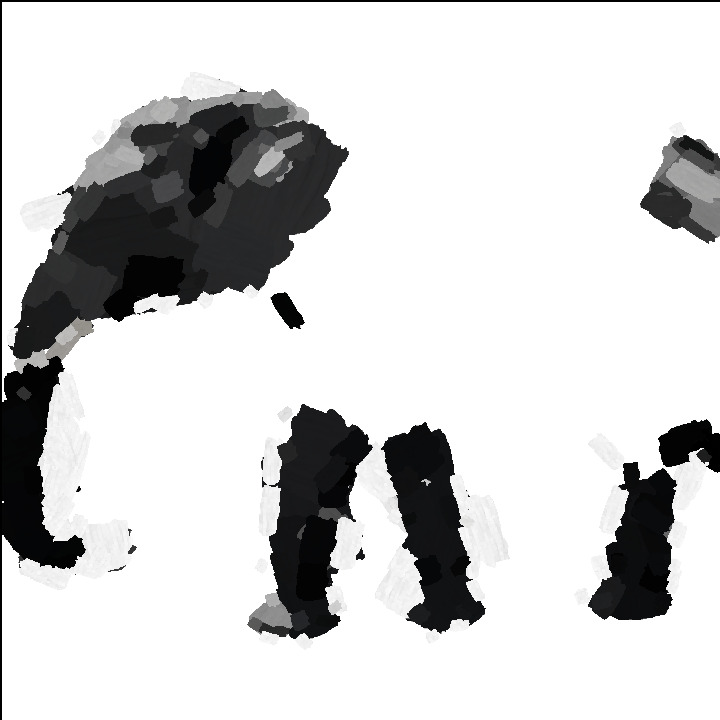} &
\includegraphics[width=0.1\textwidth]{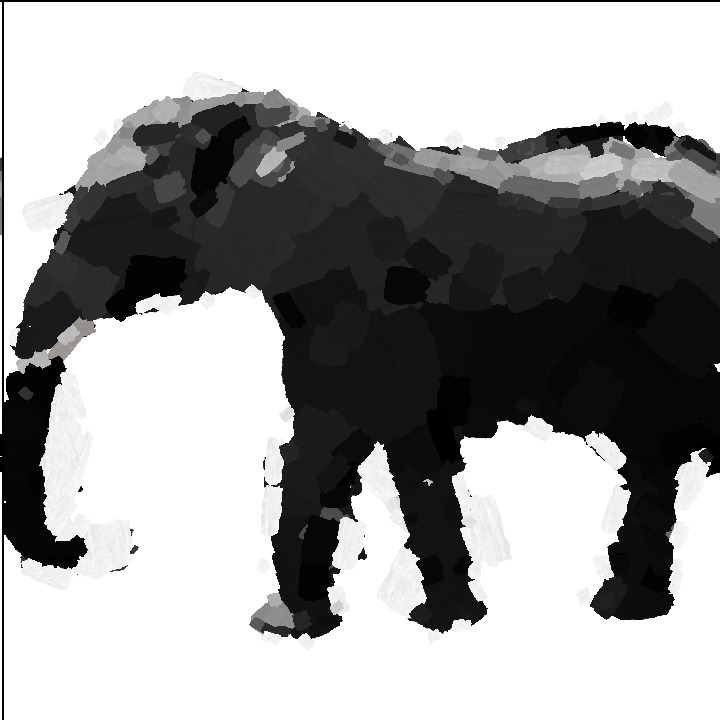} & 
\includegraphics[width=0.1\textwidth]{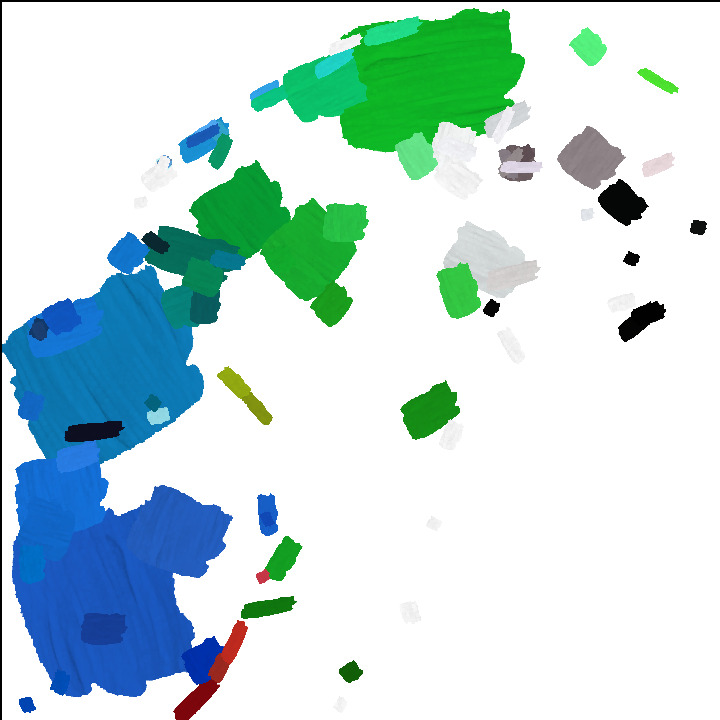} &
\includegraphics[width=0.1\textwidth]{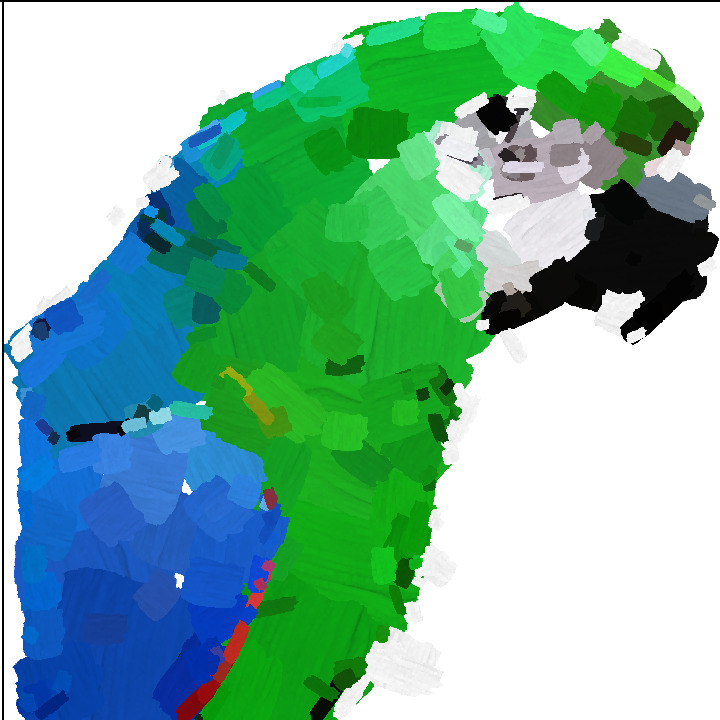} &
\includegraphics[width=0.1\textwidth]{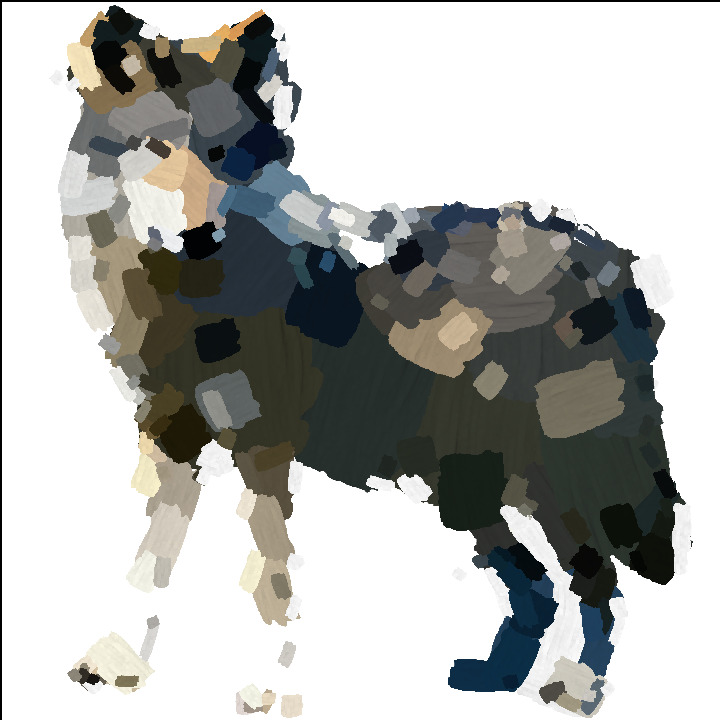} &
\includegraphics[width=0.1\textwidth]{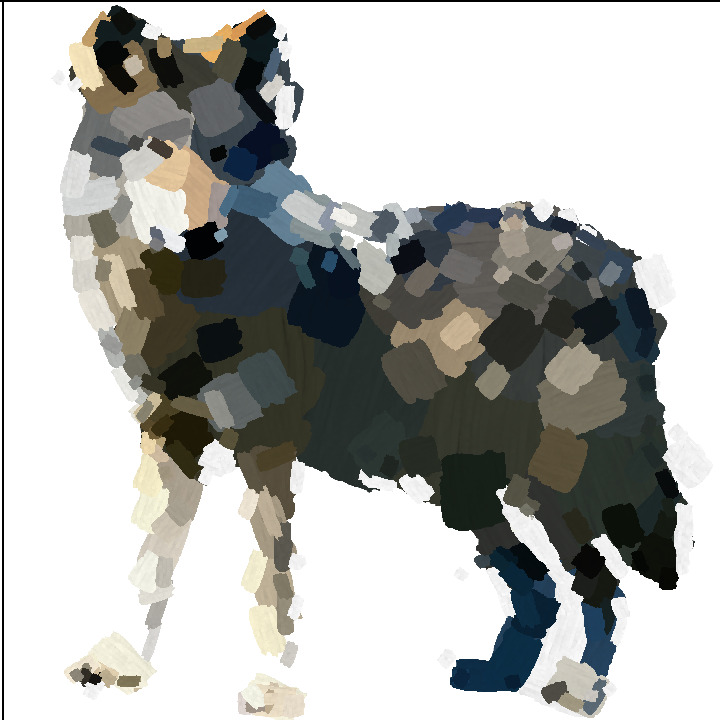} &

\includegraphics[width=0.1\textwidth]{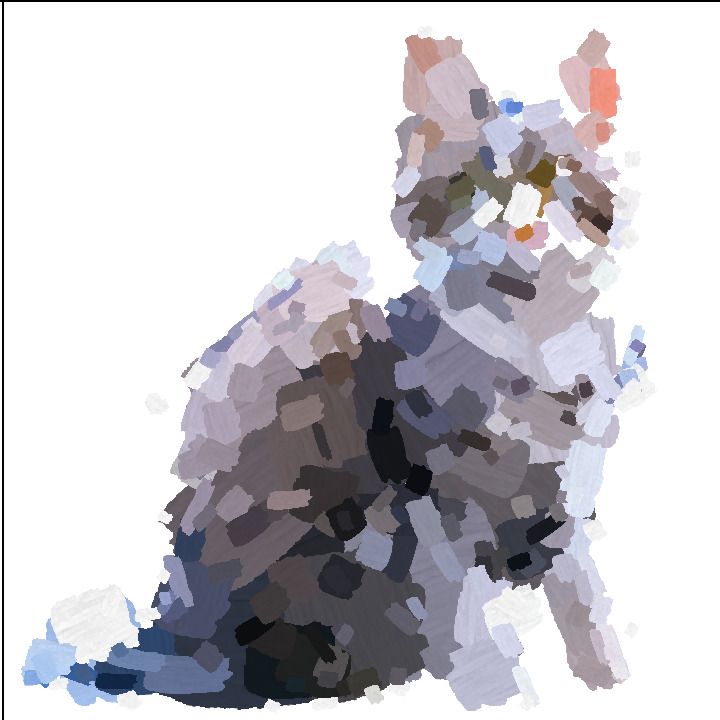} \\

\includegraphics[width=0.1\textwidth]{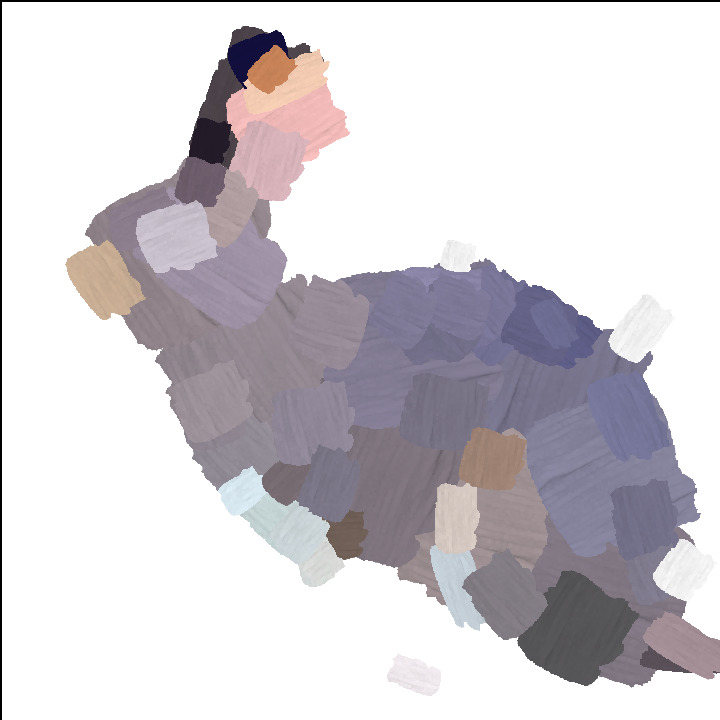} &
\includegraphics[width=0.1\textwidth]{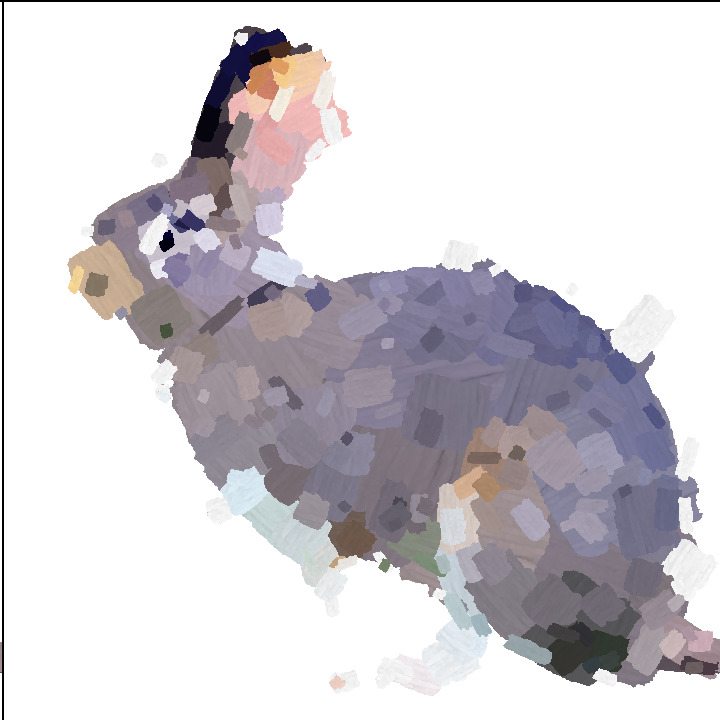} &
\includegraphics[width=0.1\textwidth]{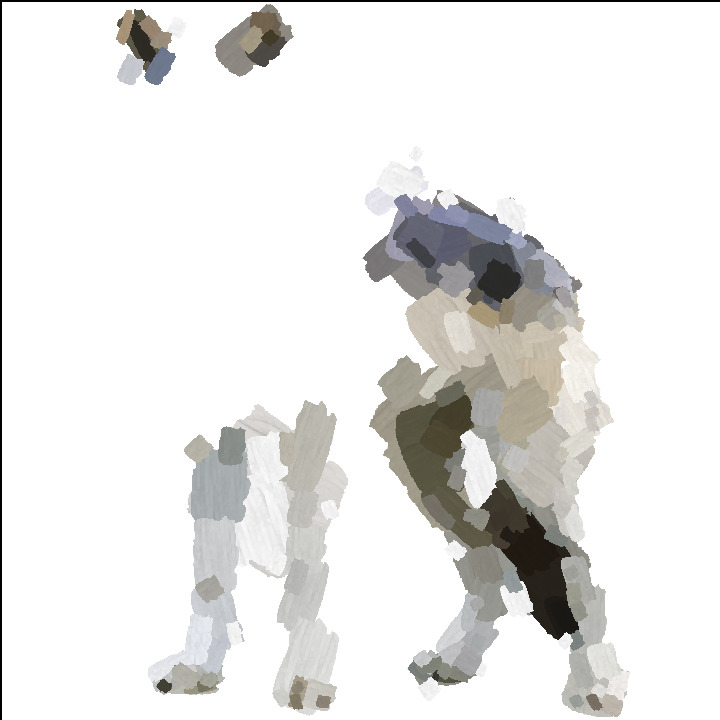} &
\includegraphics[width=0.1\textwidth]{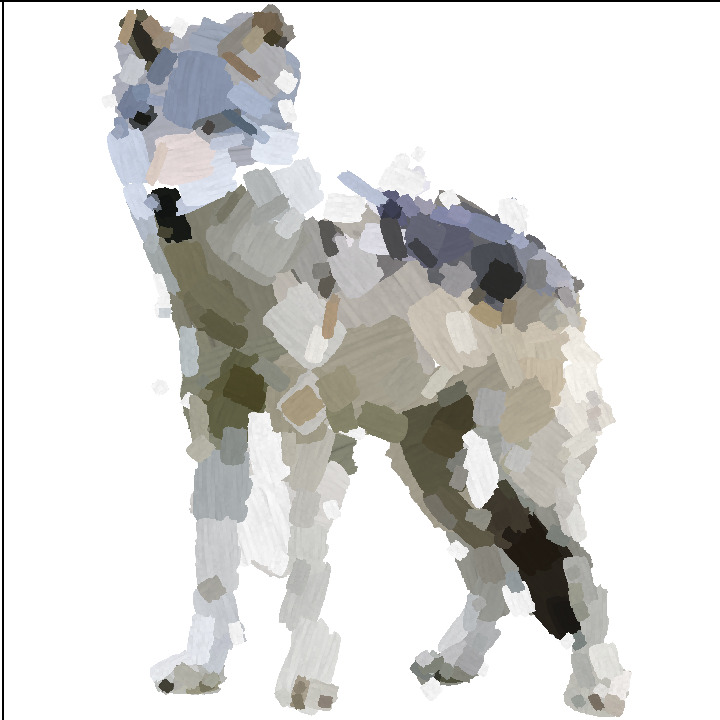} & 
\includegraphics[width=0.1\textwidth]{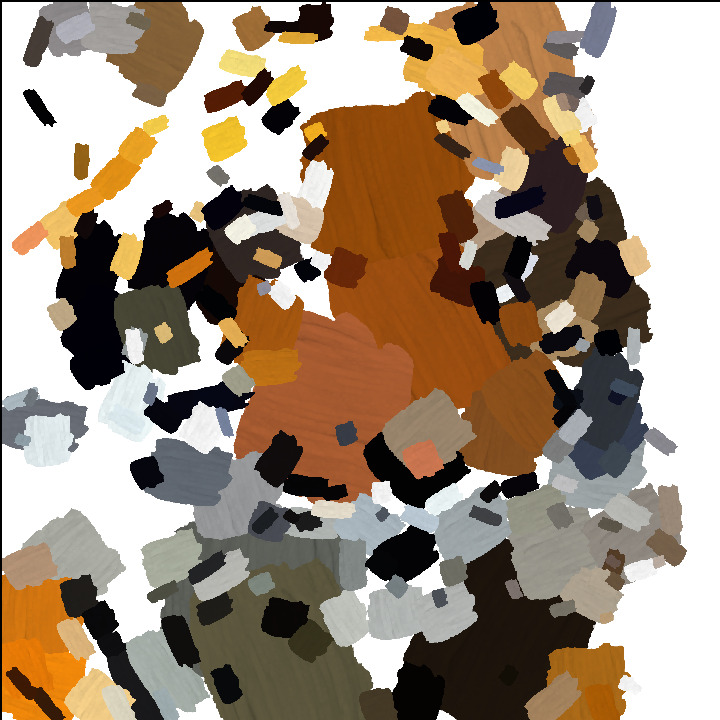} &
\includegraphics[width=0.1\textwidth]{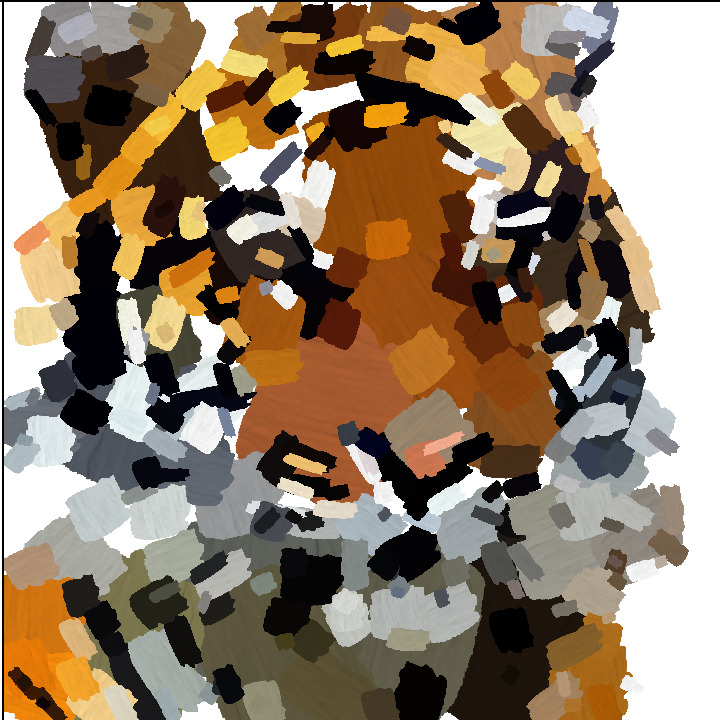} &
\includegraphics[width=0.1\textwidth]{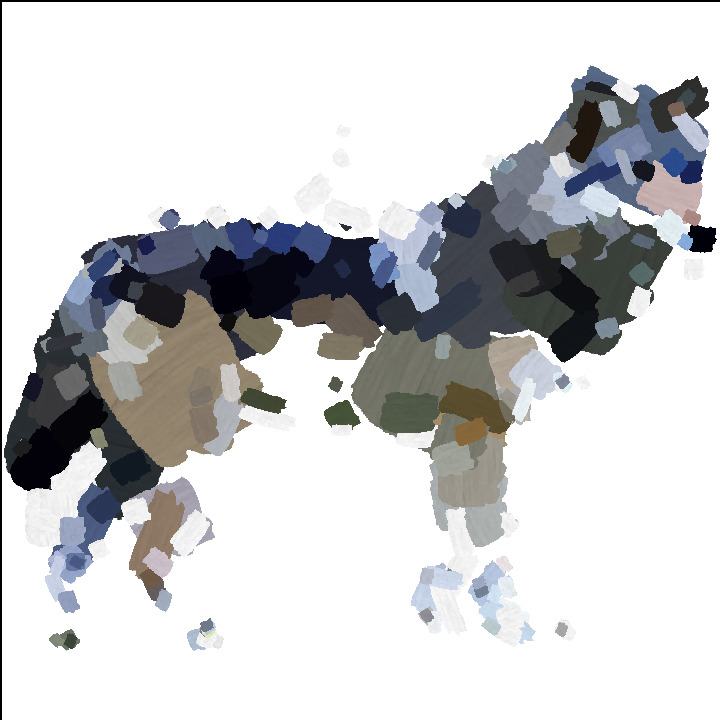} &
\includegraphics[width=0.1\textwidth]{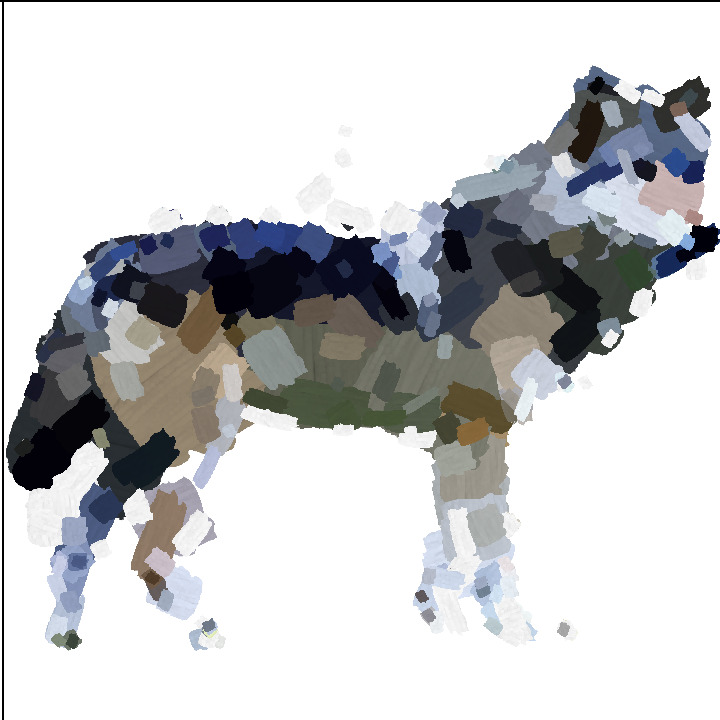} &

\includegraphics[width=0.1\textwidth]{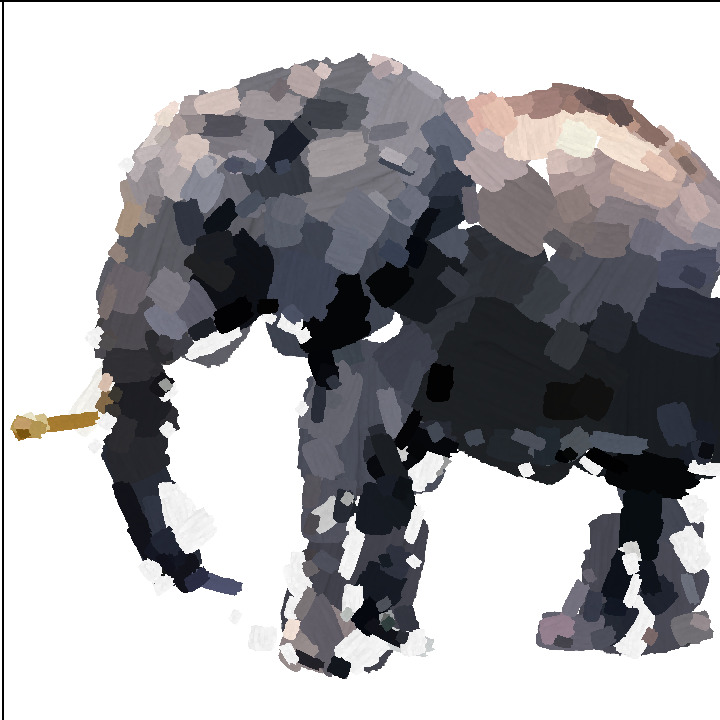} \\
\bottomrule
\end{tabular}
}
\caption{Qualitative completion results of our method. Images are organized in blocks of 2. On the left the conditioning sequence $\vecsequencecond$, on the right the conditioning with the completion added, \ie $\tilde{\boldsymbol{s}}_0$.}
\label{fig:qualitatives}
\vspace{-1.em}
\end{figure*}

\endgroup
To the best of our knowledge, no existing model is able to operate interactively on painting strokes for image generation. Hence, for comparative analysis, we used state-of-the-art methods for sequence modeling and adapted them to fit our task. We first provide the implementation details of our method in Sec~\ref{sec:implementation}. We then present comparison against the baselines and discuss the results in Sec.~\ref{sec:quantitative}. Besides automatic metrics, we have conducted a user study to assess human preference among different methods and report the results in Sec.~\ref{sec:userstudy}. Moreover, we evaluate the effect of our proposed components and training strategies with an ablation study in Sec.~\ref{sec:ablation}. Lastly, we report the real-time analysis of our method in Sec.~\ref{sec:realtime}.

\subsection{Implementation Details}
\label{sec:implementation}
\noindent \textbf{Architecture.} Our model is based on the Transformer~\citep{vaswani2017attention} architecture. Given an input sequence of strokes of dimension $L \times 8$, we split it into context and target using our IAM masking strategy (see Fig.~2 of main paper). In practice, the sequence is multiplied by two binary masks, with zero to mask out strokes. We then concatenate the context and target sequences on the feature dimension, resulting in a sequence of size $L \times 16$. Next, we use a linear layer to project the input sequence of strokes to the feature size of the Transformer model $L \times F$. Since the two sequences are mutually exclusive with value $0$ in the \textit{empty} positions, the linear projection does not act on these values, effectively acting as two separate projections on the two sequences. Lastly, we add sinusoidal positional embeddings~\citep{vaswani2017attention} to the sequence and feed it to the Transformer model. GELU non-linearities~\citep{hendrycks2016gaussian} (approximated with tanh) are used in the core Transformer.

To inject the diffusion timestep information $\difftimestep$ and control the diffusion process with the class information $\class$, we employ the adaLN-Zero block~\citep{peebles2022scalable}, due to its demonstrated performance. To embed the input time steps we use a 256-dimensional frequency embedding~\citep{dhariwal2021diffusion}, followed by a two-layer MLP with SiLU activations with a output dimension equal to the Transformer's hidden size $F$. Next, we sum the time-step and the class embedding and feed it to the adaLN-Zero layer.
In addition to regressing the values of the scale $\gamma$ and the shift $\beta$ as in AdaIN~\citep{huang2017arbitrary}, adaLN-Zero also regresses dimension-wise scaling parameters $\delta$, which are applied before any residual connection within the blocks. We regress the scaling parameters using a linear projection, which is initialized to output a zero-vector for all the $\delta$ values. For each operation $op$ (attention or feed-forward) the equation becomes: $\text{out} = x + \delta \cdot \text{op}(\gamma \cdot x + \beta)$.

\noindent \textbf{Masking.} The \textit{random} masking technique is implemented by randomly selecting a masking ratio in the range $[0.1, 0.9]$ and dropping strokes accordingly in the sequence. For the \textit{level} masking, we randomly select a granularity level between 1 and 3, and mask all the strokes that belong to subsequent levels, effectively training the model to add details to an existing rough representation of the given object. The \textit{block} masking technique involves randomly selecting a starting point in the sequence and masking a block of contiguous strokes with a length in the range $[10, 0.75 \cdot L]$, with $L$ the sequence length. Lastly, in \textit{square} case we select a stroke in the sequence at random as the center and build a square of side length $0.5$, masking all the strokes spatially contiguous to the center. 
During training, we select one of the above strategies with equal probability and apply it to split the sequence into context and target strokes.

\noindent \textbf{Diffusion Model.} We followed the formulation of ``simple diffusion''~\citep{hoogeboom2023simple}, defining the noising schedule in terms of $\text{SNR} = \alpha^2_t/\sigma^2_t$ and keeping the default values $\min \log\text{SNR}=-15$ and $\max \log\text{SNR}=15$.
We do not employ any weighting on the losses and train the model with the $\mathcal{L}_{simple}$ formulation~\citep{ho2020denoising}.

\noindent \textbf{Training hyper-parameters.}  We train all models with Adam~\citep{kingma2014adam} and a cosine annealing schedule for the learning rate, where the maximum value is $1 \times 10^{-4}$. We train with a global batch size of 256 divided on 4 GPUs. Following the common practice in the generative modeling literature, we maintain an exponential moving average (EMA) of the weights over training with a decay of 0.9999. All reported results use the EMA model.

\noindent \textbf{Sampling.} For the diffusion reverse process we use the standard DDPM~\citep{ho2020denoising} sampler with 1000 steps.

\noindent \textbf{\ourcfg.} We choose the hyper-parameters of our Classifier-free guidance, $s_1$ and $s_2$ to be $1.5$ in order to weigh the importance of the conditioning sequence and the class information.

\noindent \textbf{Inference-time stroke position.} At inference time, \eg the demo, we start without strokes and we do not use the \textit{IAM} module. Since Stylized Neural Painting fixes the maximum number of strokes in each cell, when the user draws strokes on the canvas at inference time, the size and position of the strokes allow finding the corresponding cell. Given a level $l$ and the grid described above, a stroke in that level cannot have a size greater than $1 / l$, therefore we can pick the correct level by choosing the first level whose maximum stroke size is the closest to the input. After the level, the correct cell can be identified by the position of the strokes. Then, the stroke gets added to the part of the sequence corresponding to the given cell in the first available slot, and we update the occupant mask \textbf{m} accordingly.

\subsection{Comparison}
\label{sec:quantitative}

\pp{Baselines}
Inspired by the success of Transformer-based models in NLP tasks, where the input belongs to a fixed-sized vocabulary, we compare our method with a baseline operating with discrete input. In practice, we discretize the stroke parameters independently into a codebook of size 256. We train the model in a BERT-like fashion \citep{devlin2018bert}, masking the strokes and predicting the original tokens with cross-entropy loss. At inference time, we operate this baseline in two ways: (i) predicting the missing strokes in one step (referred to as \emph{BERT}), or (ii) following \citep{chang2022maskgit}, iteratively sampling based on the network confidence on the generated tokens, leading to a multi-step sampling process similar to diffusion-based models (referred to as \emph{Mask-GIT}).
Additionally, we compare with a baseline working with continuous stroke parameters, named the \emph{Continuous Transformer} baseline.  In this case, the model is trained to regress the masked stroke parameters with an MSE loss. We employ a Transformer architecture with similar configurations, FLOPS, and parameter counts, and apply our \masking~to all the models.

As shown in Tab.~\ref{tab:quantitative1}, the \emph{Continuous Transformer} baseline fails to capture the distribution of target strokes and to use the contextual information provided by the conditioning sequence. Due to the MSE loss used during training, it converges to a mean representation for each class, with failure cases more pronounced in the unconditional case (see Fig.~\ref{fig:baselines}). On the other hand, the baselines working with discrete strokes fail completely at modeling the relationship between the conditioning strokes and the masked ones. We impute this to the independent conversion of each stroke parameter to a different token, leading to an input sequence of length $\numstrokes \times 8$, and making the computation of relationships between strokes challenging.
Our method consistently outperforms other approaches and achieves strong performances across all the different tasks.

\pp{Qualitative Results}
Qualitative results are presented in Fig.~\ref{fig:qualitatives} and~\ref{fig:more_qualitatives}, where the images are displayed in the format of a conditioning sequence $\vecsequencecond$ and predicted completion $\tilde{\boldsymbol{s}}_0$. We notice how our method predicts completion consistent with the context information in various tasks. For example, in the first column (\emph{Level}), given a rough sketch of the desired image as context, the model adds fine-grained details in a consistent manner. At the same time, when a large portion of a detailed painting is missing, the model generates both coarse and detailed strokes that harmonize with the context and complete it in a plausible manner (e.g. second column, \emph{Square}). Finally, our method can generate realistic paintings without any conditioning but the class information (last column, \emph{No context}), showcasing another useful use case. \\
\noindent \textit{Robustness to random masking.} We investigate the robustness of our method to varying levels of random masking and complement the quantitative results of Tab.~3 of the main paper with the qualitative results in Fig.~\ref{fig:random_masking}. This experiment showcases the effectiveness of the conditioning sequence as a strong prior for generation, even in high masking regimes. The results suggest that our model is capable of producing accurate predictions with limited input strokes. \\
\noindent \textit{Diverse suggestions from the same context.}
In Fig.~\ref{fig:multiple} we exhibit the strength of diffusion models to produce diverse results given the same conditioning sequence. This feature is crucial for the \methodname~task, providing the user with diverse but coherent completion given the same context. In practice, we input the same context strokes $s^{ctx}$ with the class $c$, and sample different initial random noise $s_T \sim \mathcal{N}(0, 1)$. We can observe how all the competitions suggested by the model are coherent with the context while providing some variety and diversity among which the users can make their final decision. \\
\noindent \textit{Automatic inference without providing class.} We probe the role of the class conditioning $c$ by dropping it and providing only the context strokes to the model.
We show the results in Fig.~\ref{fig:no_class}. We can notice that the competition suggested by the model reflects the expected class, suggesting that during training the model learns to rely on the context information and implicitly learns to associate it with a specific class. This phenomenon is further encouraged by the CFG training procedure employed, in which the class is not always used to train the model. \\
We also refer the reader to the \suppmat~for a demonstration video with human interaction.

\subsection{User Study}
\label{sec:userstudy} 
We complement our quantitative analysis with a user study to compare the users' preferences among the different methods. 
We ask the users to rank the completion from different methods from worst to best, given the rendered image of the conditioning sequence $\vecsequencecond$. We compare the predictions of our method with the ones obtained with \textit{Continuous Transformer}, \textit{MaskGIT}, and the \textit{ground truth}. The study has been conducted on 20 different users, collecting a total of 894 votes. We report the preferences in Tab.~\ref{tab:userstudy}, where we exhibit competitive results with the ground truth as the best completion, and outperform the other methods as the second-best choice. 

\begin{table}[!th]
\caption{User study results, we compare our model against baselines and ground truth completion.}
\label{tab:userstudy}
\centering
\resizebox{.8\linewidth}{!}{%
\begin{tabular}{@{}lcccc@{}}
\toprule
 & Ground truth & Ours & Continuous & MaskGIT \\ \midrule
1\st choice & 81\% & 17\% & 1\% & 1\% \\
2\nd choice & 18\% & 80\% & 1\% & 1\% \\ \bottomrule
\end{tabular}
}
\vspace{-1em}
\end{table}

\subsection{Ablation Study}
\label{sec:ablation}
In this section, we evaluate the effectiveness of the components introduced in Sec.~\ref{sec:method}, and present the results in Tab.~\ref{tab:ablation}.

\begin{table}[!t]
\caption{Ablation study of our proposed components. \masking, \attbias, and \ourcfg~stand for Interaction-aware Masking, Position-aware Attention Bias, and Multiple Conditions Classifier-free guidance, respectively.}
\centering
\resizebox{1.\linewidth}{!}{%
\begin{tabular}{@{}lcccccccc@{}}
\toprule
\multirow{2}{*}{\masking} &
  \multirow{2}{*}{\attbias} &
  \multirow{2}{*}{Ada-$\lambda$} &
  \multirow{2}{*}{\ourcfg} &
  \multicolumn{5}{c}{FID $\downarrow$}  \\ \cmidrule(l){5-9} 
       &        &        &        & \multicolumn{1}{c}{Block}   & Level  & Random & Square & No ctx  \\ \midrule
\xmark & \xmark & \xmark & \xmark & 8.78          & 11.66         & 18.41          & 7.51          & 233.21 \\
\cmark & \xmark & \xmark & \xmark & 9.23          & 11.65         & 21.92          & 7.17          & 36.96 \\
\cmark & \cmark & \xmark & \xmark & 9.51          & 11.85         & 22.61          & 7.34          & 37.94 \\
\cmark & \cmark & \cmark & \xmark & 8.43          & 10.34         & 19.25          & 6.63          & 30.46 \\
\cmark & \cmark & \cmark & \cmark & \textbf{6.20} & \textbf{7.29} & \textbf{12.69} & \textbf{5.53} & \textbf{30.12} \\ \bottomrule
\end{tabular}
}
\label{tab:ablation}
\vspace{-1em}
\end{table}

\pp{\masking} We start our analysis by training a model without any of our components, with simple random masking similar to BERT (first row). We then introduce \masking~ and, as expected, its role is crucial to perform well at inference time on the synthetic tasks, especially in the \emph{no context} case (second row).

\pp{\attbias} We then incorporate the Position-aware Attention bias (\attbias), using a constant weighting of $\lambda=0.5$ across the diffusion time-steps $t$. We observe an performance degradation in all the tasks, suggesting this fixed bias is not optimal to guide the generative process. 

\pp{Ada-$\boldsymbol{\lambda}$} To counter this effect, we introduce adaptive weighting $\lambda$ which modifies the strength of the bias as a function of $\logsnr$ of the diffusion model. The results in the fourth row prove the effectiveness of this choice, reducing the FID in all the tasks, and particularly in the \textit{no ctx} case. 

\pp{\ourcfg} In the last row, we explore the role of \ourcfg~at inference time, which leads to further improvement in the performances. 

\pp{Robustness} Moreover, we study the robustness of our model to different masking ratios. We test the model on \textit{random} task, varying the percentage of masked strokes (see Tab.~\ref{tab:robustness}). Our model exhibits good performance even in a high masking regime, \ie 80\%, showing that even a few strokes are sufficient to provide context to the network.

\pp{Scaling} Lastly, we investigate the scaling laws governing our model and design three configurations: \ourT-S(mall), \ourT-B(ase), and \ourT-L(arge), as described in Tab.~\ref{tab:models}. The results reported in the table correspond to the ablation with all components included except \ourcfg. We observed a consistent performance improvement by scaling up the model, suggesting that performance could be further improved by increasing the capacity.

\begin{table}[!th]
\caption{Performance of our model \wrt masking percentage in the \textit{random} setting.}
\label{tab:robustness}
\centering
\resizebox{.8\linewidth}{!}{%
\begin{tabular}{@{}lccccc@{}}
\toprule
Masking \% & 20\% & 40\% & 60\% & 80\% & 100\% \\ \midrule
FID $\downarrow$ & 6.57 & 10.42 & 13.67 & 19.05 & 30.12 \\ \bottomrule
\end{tabular}
}
\vspace{-1em}
\end{table}

\begin{table}[!th]
\caption{Details of \ourT~models. Model configurations for the Small (S), Base (B), and Large (L) variants and corresponding performance on the proposed task.}
\label{tab:models}
\centering
\resizebox{1.\linewidth}{!}{%
\begin{tabular}{@{}lccccccccc@{}}
\toprule
\multirow{2}{*}{Model} & \multirow{2}{*}{Layers N} & \multirow{2}{*}{F} & \multirow{2}{*}{Heads} & \multirow{2}{*}{GFLOPS} & \multicolumn{5}{c}{FID $\downarrow$} \\ \cmidrule(l){6-10} 
 &  &  &  &  & Block & Level & Random & Square & Unconditional \\ \midrule
\ourT-S & 6 & 576 & 6 & 9.09 & 8.67 & 11.24 & 22.69 & 7.34 & 55.25 \\
\ourT-B & 8 & 768 & 12 & 21.27 & 6.81 & 8.39 & 14.96 & 6.02 & 36.29 \\
\ourT-L & 12 & 768 & 12 & 31.89 & \textbf{6.20} & \textbf{7.29} & \textbf{12.69} & \textbf{5.53} & \textbf{30.12} \\ \bottomrule
\end{tabular}
}
\vspace{-1em}
\end{table}

\subsection{Real-time analysis}
\label{sec:realtime}

\begin{table}[!th]
\caption{Model performance \wrt number of sampling steps.}
\label{tab:steps}
\centering
\resizebox{1.\linewidth}{!}{%
\begin{tabular}{@{}cccccc@{}}
\toprule
\multirow{2}{*}{Sampling steps} & \multicolumn{5}{c}{FID $\downarrow$} \\ \cmidrule(l){2-6} 
 & Block & Level & Random & Square & Unconditional \\ \midrule
35 & 6.42 & 7.83 & 12.81 & 5.40 & 36.74 \\
50 & 6.10 & 7.59 & 11.97 & 5.34 & 31.64 \\
70 & 5.90 & 7.12 & 11.71 & 5.28 & 28.30 \\
125 & 5.95 & 7.01 & 12.06 & 5.24 & 27.95 \\
250 & 5.98 & 7.19 & 12.12 & 5.42 & 28.05 \\
500 & 6.08 & 7.09 & 12.37 & 5.47 & 29.73 \\
1000 & 6.20 & 7.29 & 12.69 & 5.53 & 30.12 \\ \bottomrule

\end{tabular}
}
\vspace{-1em}
\end{table}

We test our model on a single NVIDIA A100. Performing 1000 denoising steps requires $\sim 8$ seconds, meaning $\sim 8$ ms per step.
In Table~\ref{tab:steps}, we present the effect of using a reduced number of steps for sampling, and we find that the model performs consistently well after $\sim70$ steps. A number of steps under 70 leads to degraded performance. Thus, we opt to use 70 steps for demo purposes, resulting to a sampling time of $\sim 560$ ms, a reasonable time for real-time interactions. 

\section{Conclusions}
\label{sec:conclusion}

We introduced \methodname, a novel task to facilitate collaborative art generation. We exploit a parametrized vector formulation for strokes to achieve editability and composability in painting generation. We proposed a novel Transformer-based architecture to solve the \methodname~task and designed a novel attention mechanism and masking scheme to tackle the challenges brought by the stroke formulation and the arbitrary user interaction, exhibiting state-of-the-art performance both quantitatively and qualitatively. 
Our proposed method is class-agnostic thus being potentially applicable to datasets of higher semantic cardinality.
Future study involves extending the dataset for further investigation and incorporating language models to enable open-set semantics. 
Moreover, we plan to investigate more efficient sampling and diffusion strategies in order to enable a more reactive user experience with the model. Finally, it is also interesting to study how to better blend the generated objects with the background.




\begingroup
\renewcommand{\arraystretch}{0.} 

\begin{figure*}[!t]
\resizebox{.98\linewidth}{!}{%
\setlength\tabcolsep{0.pt}
\begin{tabular}{ccc|ccc}
\toprule

\textbf{Generated image} & \textbf{Segmented animal} & \multicolumn{1}{c|}{\textbf{Stroke representation}} & \textbf{Generated image} & \textbf{Segmented animal} & \textbf{Stroke representation} \\ 
\midrule

\includegraphics[width=0.2\textwidth]{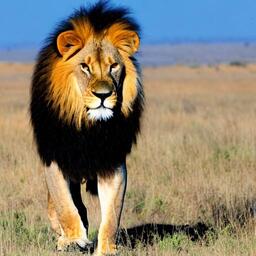} &
\includegraphics[width=0.2\textwidth]{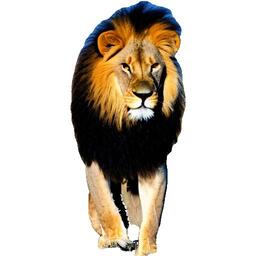} & 
\includegraphics[width=0.2\textwidth]{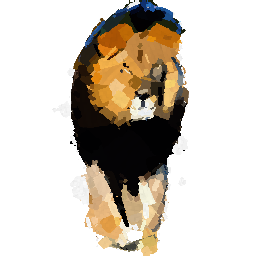} & 
\includegraphics[width=0.2\textwidth]{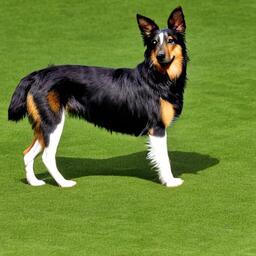} &
\includegraphics[width=0.2\textwidth]{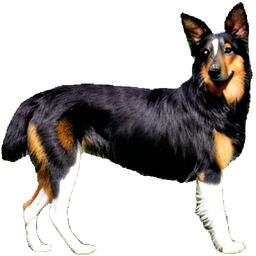} & 
\includegraphics[width=0.2\textwidth]{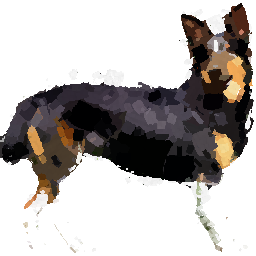} \\

\includegraphics[width=0.2\textwidth]{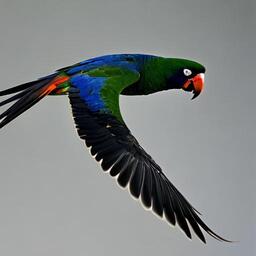} &
\includegraphics[width=0.2\textwidth]{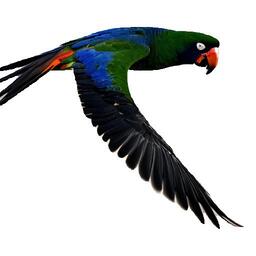} & 
\includegraphics[width=0.2\textwidth]{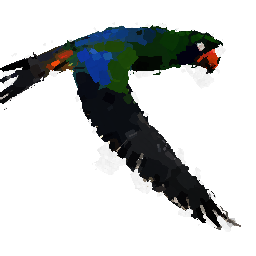} & 
\includegraphics[width=0.2\textwidth]{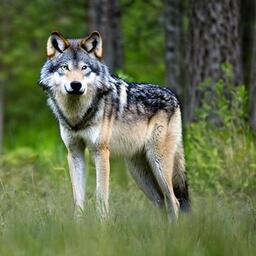} &
\includegraphics[width=0.2\textwidth]{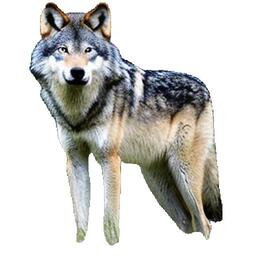} & 
\includegraphics[width=0.2\textwidth]{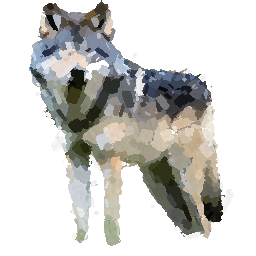} \\

\includegraphics[width=0.2\textwidth]{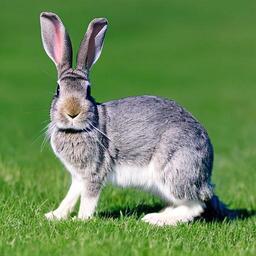} &
\includegraphics[width=0.2\textwidth]{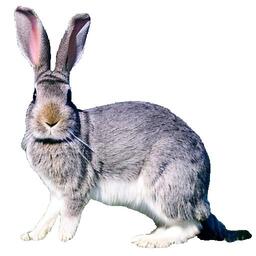} & 
\includegraphics[width=0.2\textwidth]{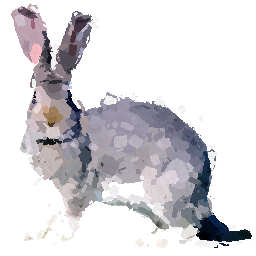} & 
\includegraphics[width=0.2\textwidth]{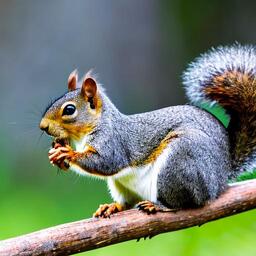} &
\includegraphics[width=0.2\textwidth]{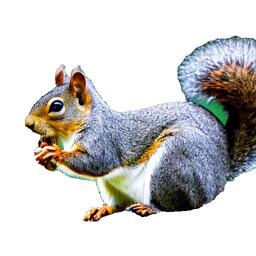} & 
\includegraphics[width=0.2\textwidth]{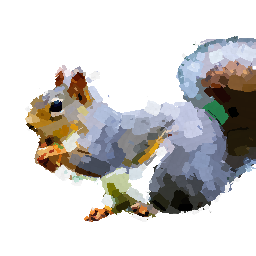} \\

\includegraphics[width=0.2\textwidth]{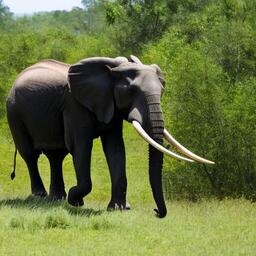} &
\includegraphics[width=0.2\textwidth]{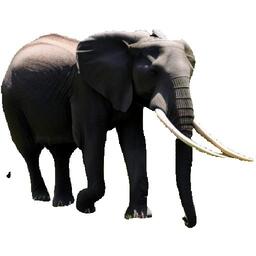} & 
\includegraphics[width=0.2\textwidth]{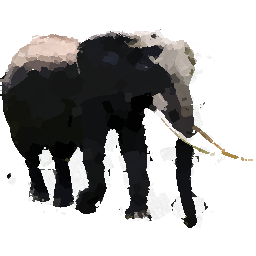} & 
\includegraphics[width=0.2\textwidth]{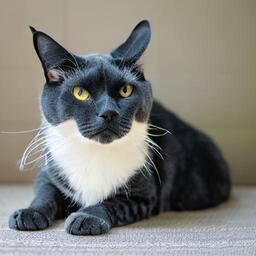} &
\includegraphics[width=0.2\textwidth]{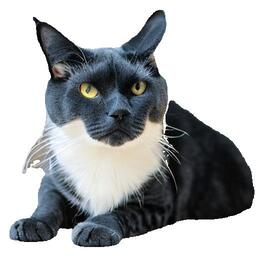} & 
\includegraphics[width=0.2\textwidth]{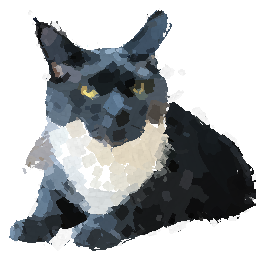} \\

\includegraphics[width=0.2\textwidth]{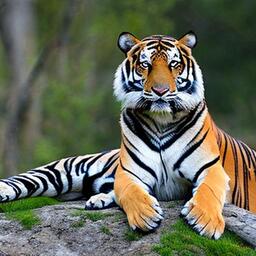} &
\includegraphics[width=0.2\textwidth]{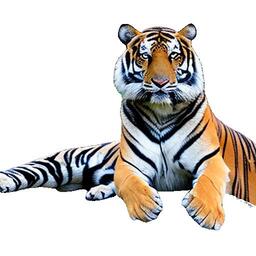} & 
\includegraphics[width=0.2\textwidth]{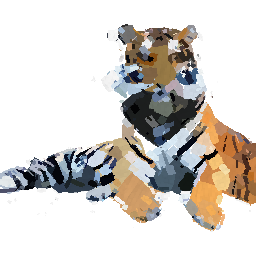} & 
\includegraphics[width=0.2\textwidth]{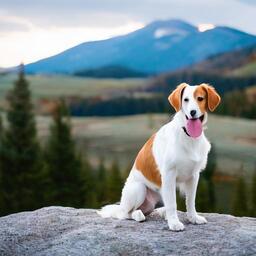} &
\includegraphics[width=0.2\textwidth]{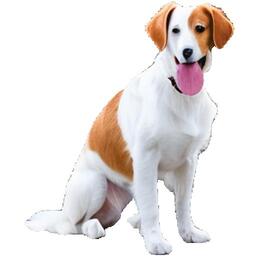} & 
\includegraphics[width=0.2\textwidth]{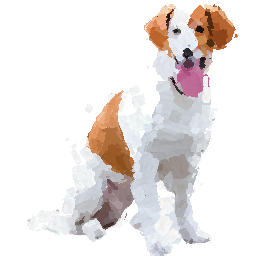} \\

\includegraphics[width=0.2\textwidth]{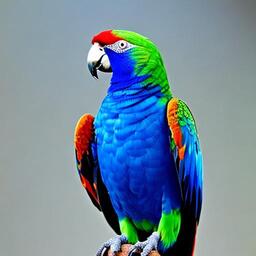} &
\includegraphics[width=0.2\textwidth]{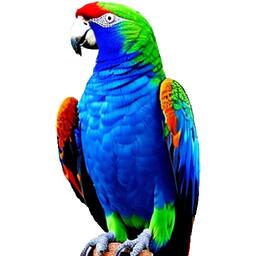} & 
\includegraphics[width=0.2\textwidth]{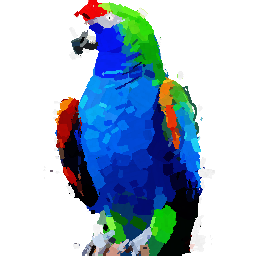} & 
\includegraphics[width=0.2\textwidth]{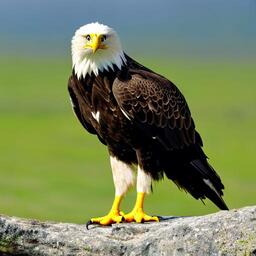} &
\includegraphics[width=0.2\textwidth]{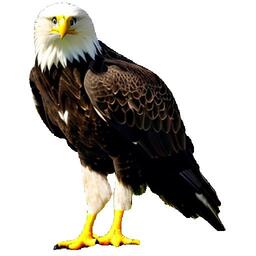} & 
\includegraphics[width=0.2\textwidth]{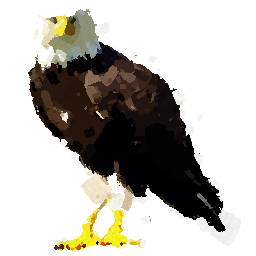} \\

\includegraphics[width=0.2\textwidth]{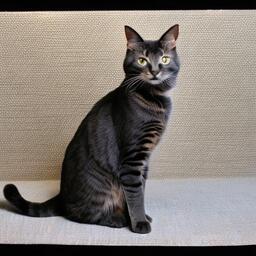} &
\includegraphics[width=0.2\textwidth]{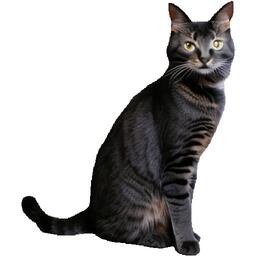} & 
\includegraphics[width=0.2\textwidth]{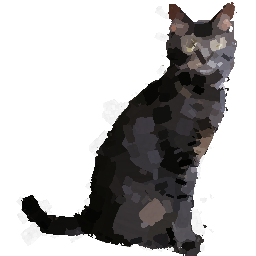} & 
\includegraphics[width=0.2\textwidth]{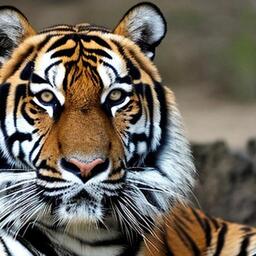} &
\includegraphics[width=0.2\textwidth]{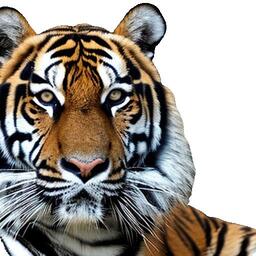} & 
\includegraphics[width=0.2\textwidth]{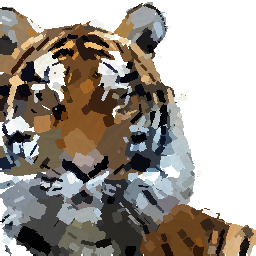} \\

\includegraphics[width=0.2\textwidth]{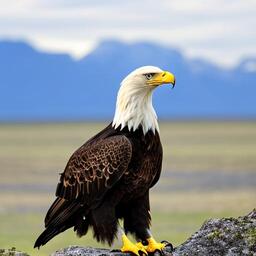} &
\includegraphics[width=0.2\textwidth]{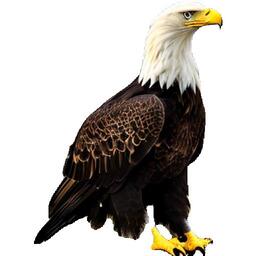} & 
\includegraphics[width=0.2\textwidth]{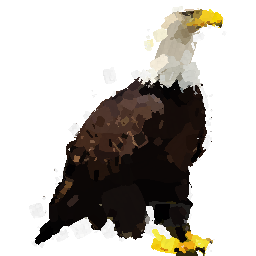} & 
\includegraphics[width=0.2\textwidth]{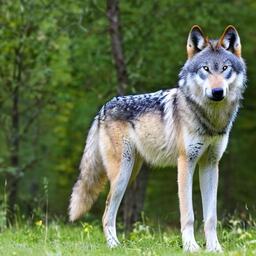} &
\includegraphics[width=0.2\textwidth]{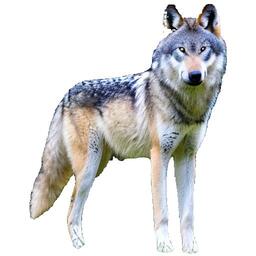} & 
\includegraphics[width=0.2\textwidth]{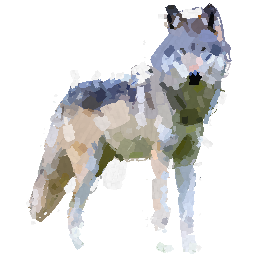} \\

\bottomrule
\end{tabular}
}
\caption{Dataset generation pipeline. We show the images generated with Stable Diffusion~\cite{rombach2022high}, the segmentation obtained with~\cite{luddecke2022clipseg}, and the stroke representation~\cite{zou2021stylized}.}
\label{fig:pipeline}
\end{figure*}

\endgroup




\begingroup
\renewcommand{\arraystretch}{0.} 

\begin{figure*}[!ht]
\resizebox{1.\linewidth}{!}{%
\setlength\tabcolsep{0.pt}
\begin{tabular}{c|ccc|}
\toprule

\multicolumn{1}{c|}{\textbf{Conditioning}} & \textbf{MaskGIT~\cite{chang2022maskgit}} & \textbf{Continuous Transformer} & \multicolumn{1}{c}{\textbf{Ours}} \\ 
\midrule

\includegraphics[width=0.2\textwidth]{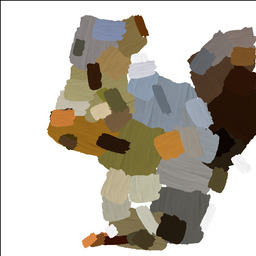} &
\includegraphics[width=0.2\textwidth]{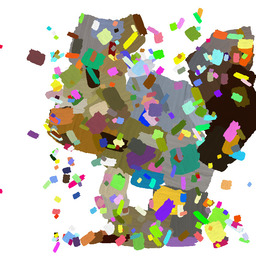} & 
\includegraphics[width=0.2\textwidth]{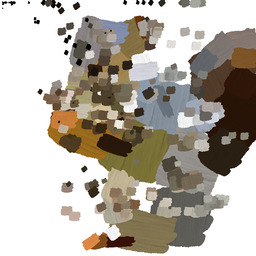} & 
\includegraphics[width=0.2\textwidth]{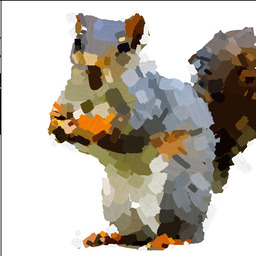} \\

\includegraphics[width=0.2\textwidth]{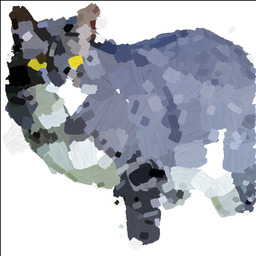} &
\includegraphics[width=0.2\textwidth]{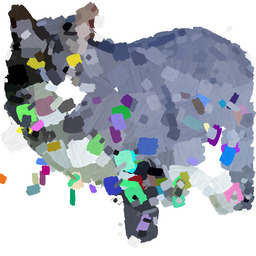} & 
\includegraphics[width=0.2\textwidth]{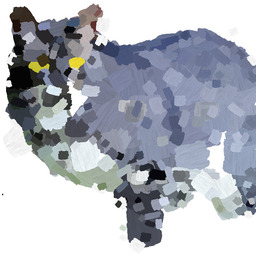} & 
\includegraphics[width=0.2\textwidth]{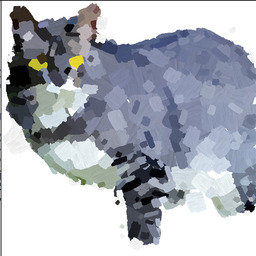} \\

\includegraphics[width=0.2\textwidth]{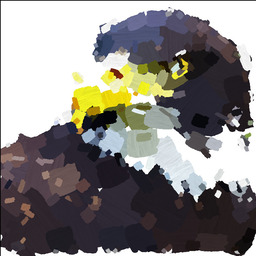} &
\includegraphics[width=0.2\textwidth]{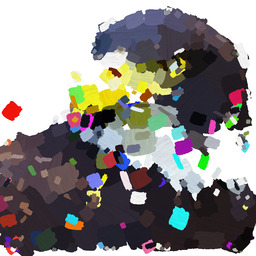} & 
\includegraphics[width=0.2\textwidth]{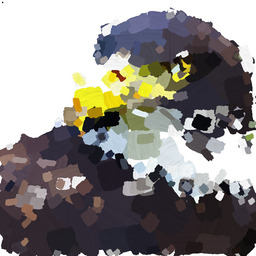} & 
\includegraphics[width=0.2\textwidth]{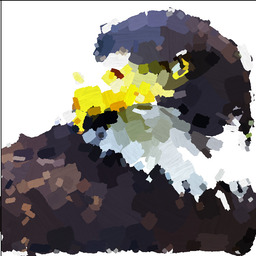} \\

\includegraphics[width=0.2\textwidth]{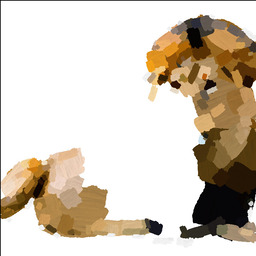} &
\includegraphics[width=0.2\textwidth]{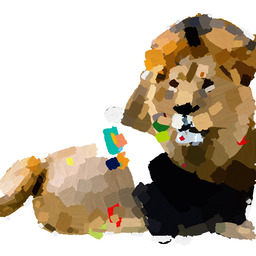} & 
\includegraphics[width=0.2\textwidth]{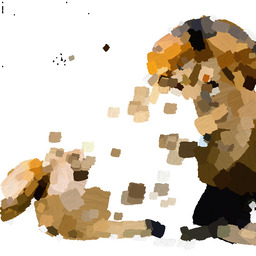} & 
\includegraphics[width=0.2\textwidth]{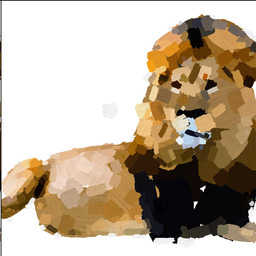} \\

\includegraphics[width=0.2\textwidth]{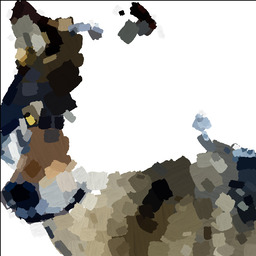} &
\includegraphics[width=0.2\textwidth]{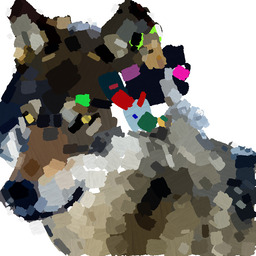} & 
\includegraphics[width=0.2\textwidth]{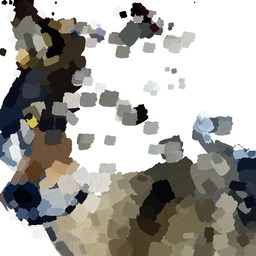} & 
\includegraphics[width=0.2\textwidth]{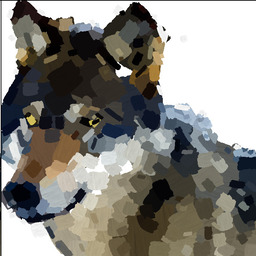} \\

\bottomrule
\end{tabular}
}
\caption{Comparison with baselines. Given the same conditioning sequence, we show the completion obtained with MaskGIT~\cite{chang2022maskgit}, Continuous Transformer and our method.}
\label{fig:baselines}
\end{figure*}

\endgroup

\begingroup
\renewcommand{\arraystretch}{0.} 

\begin{figure*}[!ht]
\resizebox{1.\linewidth}{!}{%
\setlength\tabcolsep{0.pt}
\begin{tabular}{cc|cc|cc|cc|c}
\toprule

\multicolumn{2}{c|}{\textbf{Level}} & \multicolumn{2}{c|}{\textbf{Square}} & \multicolumn{2}{c|}{\textbf{Random}} & \multicolumn{2}{c|}{\textbf{Block}} & \multicolumn{1}{c}{\textbf{No context}} \\ 
\midrule
\multicolumn{1}{c|}{$\vecsequencecond$} & \multicolumn{1}{c|}{$\tilde{\boldsymbol{s}}_0$} & \multicolumn{1}{c|}{$\vecsequencecond$} & \multicolumn{1}{c|}{$\tilde{\boldsymbol{s}}_0$} & \multicolumn{1}{c|}{$\vecsequencecond$} & \multicolumn{1}{c|}{$\tilde{\boldsymbol{s}}_0$} & \multicolumn{1}{c|}{$\vecsequencecond$} & \multicolumn{1}{c|}{$\tilde{\boldsymbol{s}}_0$} & \multicolumn{1}{c}{$\tilde{\boldsymbol{s}}_0$} \\
\midrule

\multicolumn{2}{c|}{\includegraphics[width=0.2\textwidth]{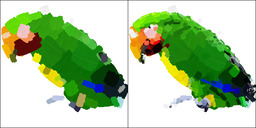}} &
\multicolumn{2}{c|}{\includegraphics[width=0.2\textwidth]{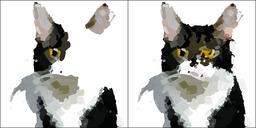}} &
\multicolumn{2}{c|}{\includegraphics[width=0.2\textwidth]{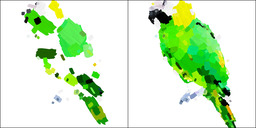}} &
\multicolumn{2}{c|}{\includegraphics[width=0.2\textwidth]{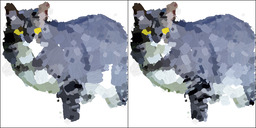}} &

\includegraphics[width=0.1\textwidth]{images/qualitatives/0ce38e73-7849-4957-a643-73fc6f6c95f0_unconditional.jpg} \\

\multicolumn{2}{c|}{\includegraphics[width=0.2\textwidth]{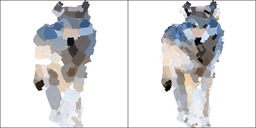}} &
\multicolumn{2}{c|}{\includegraphics[width=0.2\textwidth]{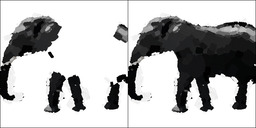}} &
\multicolumn{2}{c|}{\includegraphics[width=0.2\textwidth]{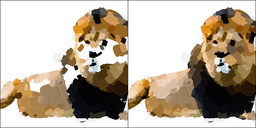}} &
\multicolumn{2}{c|}{\includegraphics[width=0.2\textwidth]{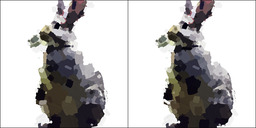}} &

\includegraphics[width=0.1\textwidth]{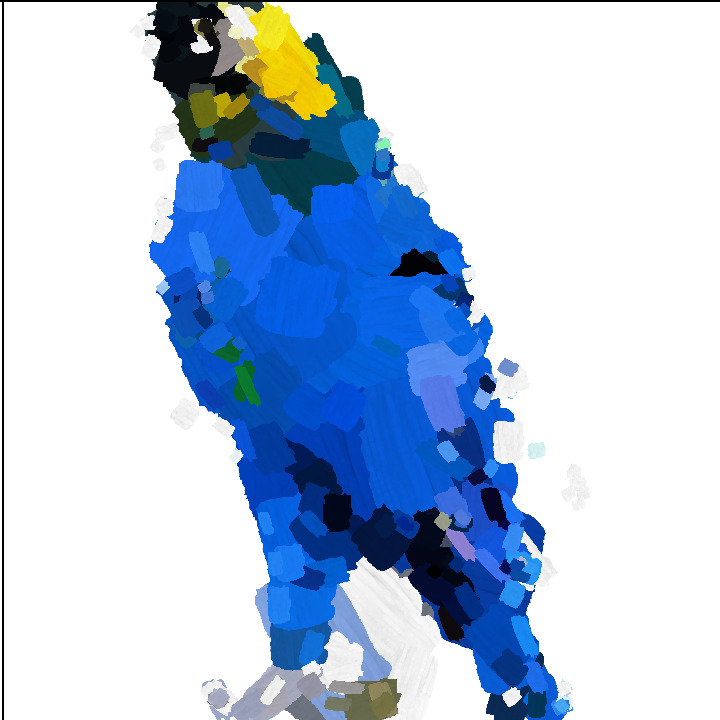} \\

\multicolumn{2}{c|}{\includegraphics[width=0.2\textwidth]{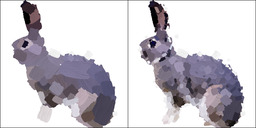}} &
\multicolumn{2}{c|}{\includegraphics[width=0.2\textwidth]{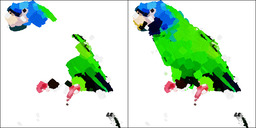}} &
\multicolumn{2}{c|}{\includegraphics[width=0.2\textwidth]{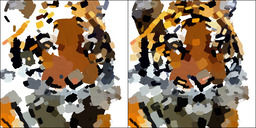}} &
\multicolumn{2}{c|}{\includegraphics[width=0.2\textwidth]{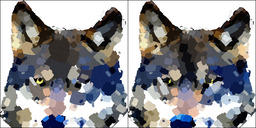}} &

\includegraphics[width=0.1\textwidth]{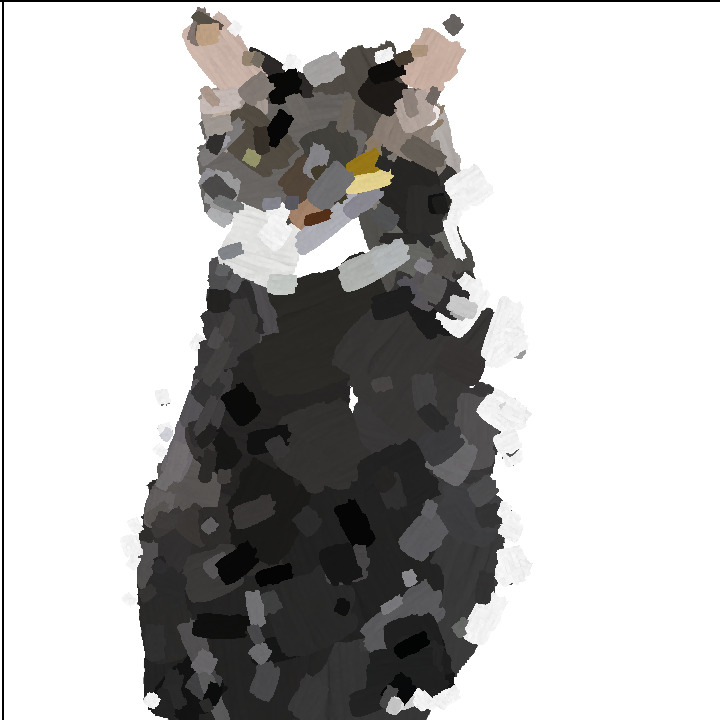} \\

\multicolumn{2}{c|}{\includegraphics[width=0.2\textwidth]{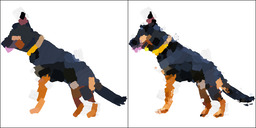}} &
\multicolumn{2}{c|}{\includegraphics[width=0.2\textwidth]{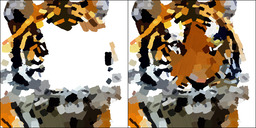}} &
\multicolumn{2}{c|}{\includegraphics[width=0.2\textwidth]{images/supp/more_per_task/6ace5ad5-8dbd-4072-b964-a54dcb4f09ca_random.jpg}} &
\multicolumn{2}{c|}{\includegraphics[width=0.2\textwidth]{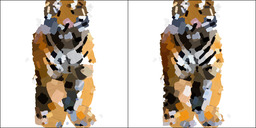}} &

\includegraphics[width=0.1\textwidth]{images/qualitatives/2bcffcc5-2f57-4e5c-a397-15c8434de2f5_unconditional.jpg} \\

\multicolumn{2}{c|}{\includegraphics[width=0.2\textwidth]{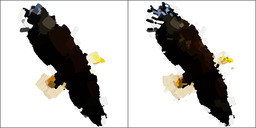}} &
\multicolumn{2}{c|}{\includegraphics[width=0.2\textwidth]{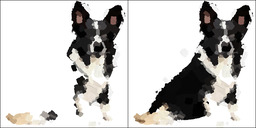}} &
\multicolumn{2}{c|}{\includegraphics[width=0.2\textwidth]{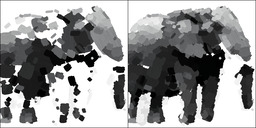}} &
\multicolumn{2}{c|}{\includegraphics[width=0.2\textwidth]{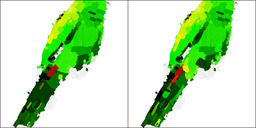}} &

\includegraphics[width=0.1\textwidth]{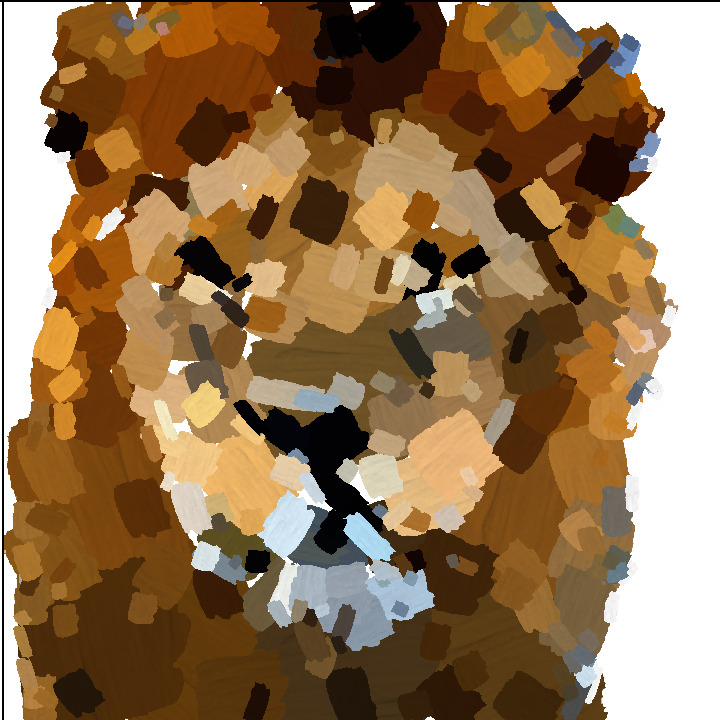} \\

\multicolumn{2}{c|}{\includegraphics[width=0.2\textwidth]{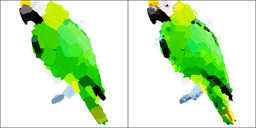}} &
\multicolumn{2}{c|}{\includegraphics[width=0.2\textwidth]{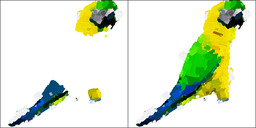}} &
\multicolumn{2}{c|}{\includegraphics[width=0.2\textwidth]{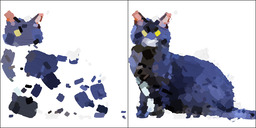}} &
\multicolumn{2}{c|}{\includegraphics[width=0.2\textwidth]{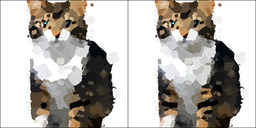}} &

\includegraphics[width=0.1\textwidth]{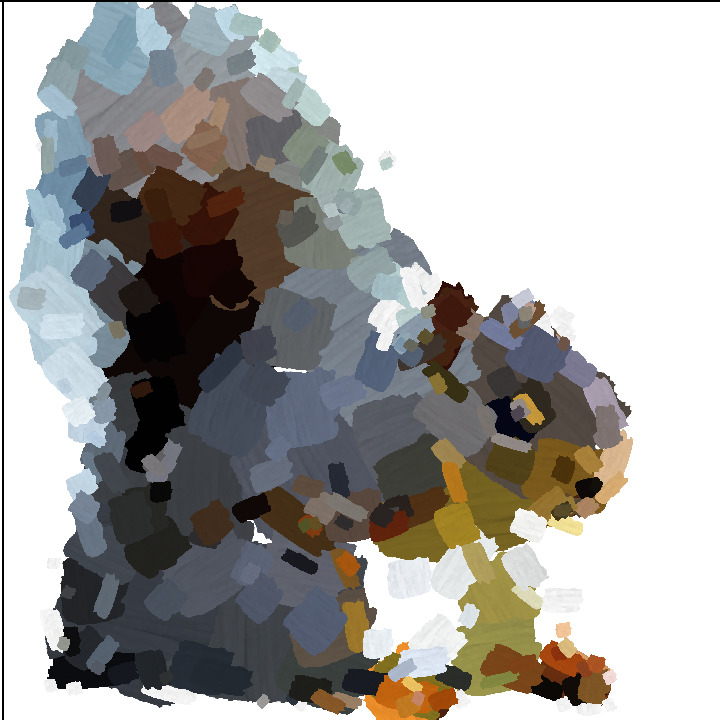} \\

\multicolumn{2}{c|}{\includegraphics[width=0.2\textwidth]{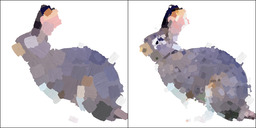}} &
\multicolumn{2}{c|}{\includegraphics[width=0.2\textwidth]{images/supp/more_per_task/8a4bb4dd-3bfc-48cf-bc71-24391fc3d083_square.jpg}} &
\multicolumn{2}{c|}{\includegraphics[width=0.2\textwidth]{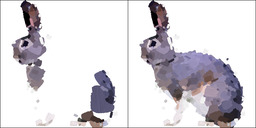}} &
\multicolumn{2}{c|}{\includegraphics[width=0.2\textwidth]{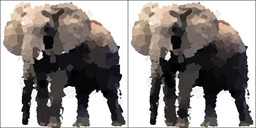}} &

\includegraphics[width=0.1\textwidth]{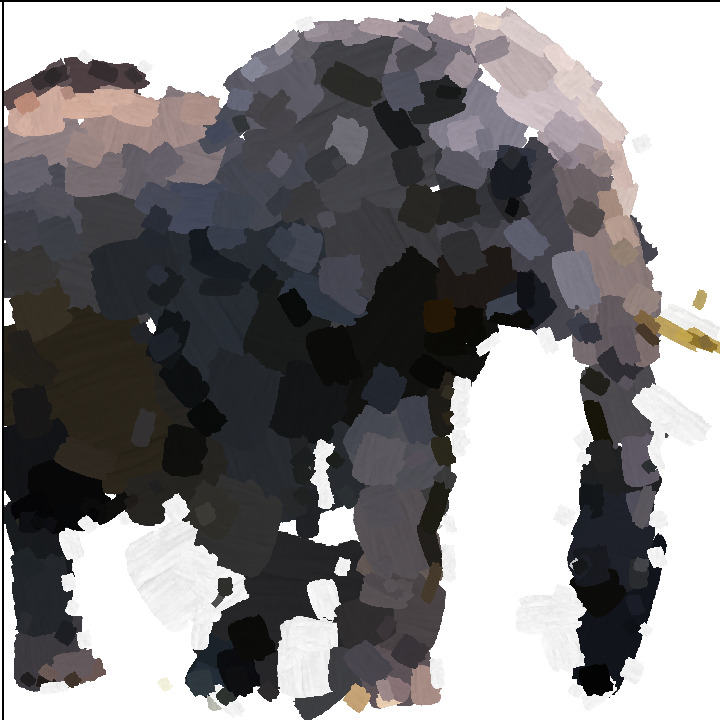} \\

\multicolumn{2}{c|}{\includegraphics[width=0.2\textwidth]{images/supp/more_per_task/c30dccfe-52c3-4563-ac39-c69cb2bee1db_level.jpg}} &
\multicolumn{2}{c|}{\includegraphics[width=0.2\textwidth]{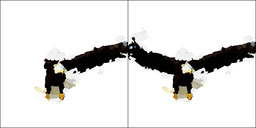}} &
\multicolumn{2}{c|}{\includegraphics[width=0.2\textwidth]{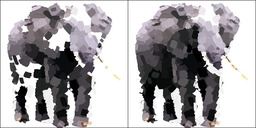}} &
\multicolumn{2}{c|}{\includegraphics[width=0.2\textwidth]{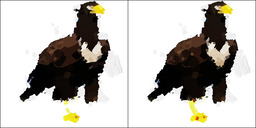}} &

\includegraphics[width=0.1\textwidth]{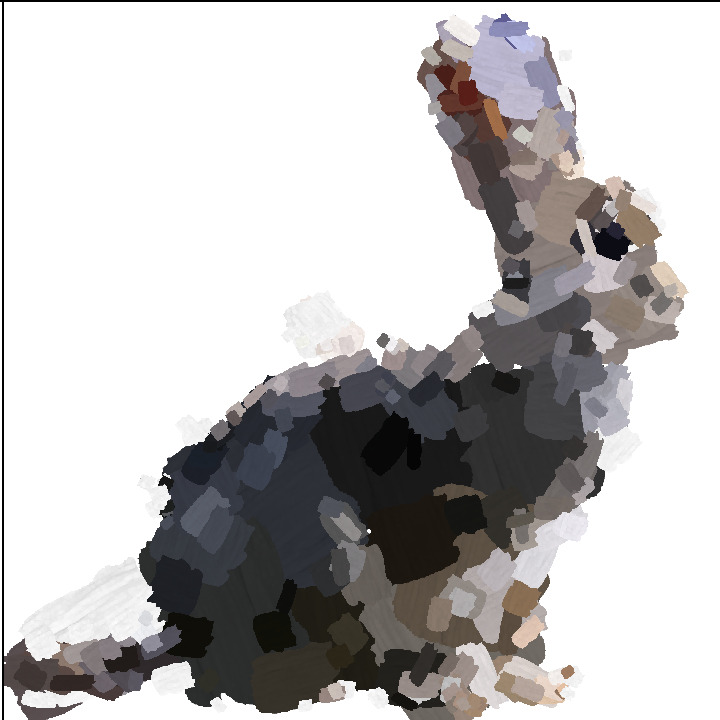} \\

\multicolumn{2}{c|}{\includegraphics[width=0.2\textwidth]{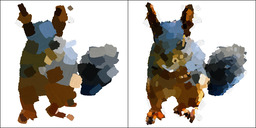}} &
\multicolumn{2}{c|}{\includegraphics[width=0.2\textwidth]{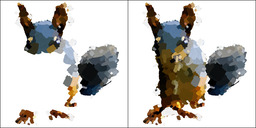}} &
\multicolumn{2}{c|}{\includegraphics[width=0.2\textwidth]{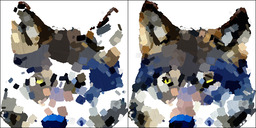}} &
\multicolumn{2}{c|}{\includegraphics[width=0.2\textwidth]{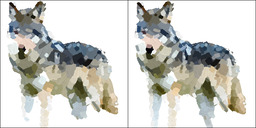}} &

\includegraphics[width=0.1\textwidth]{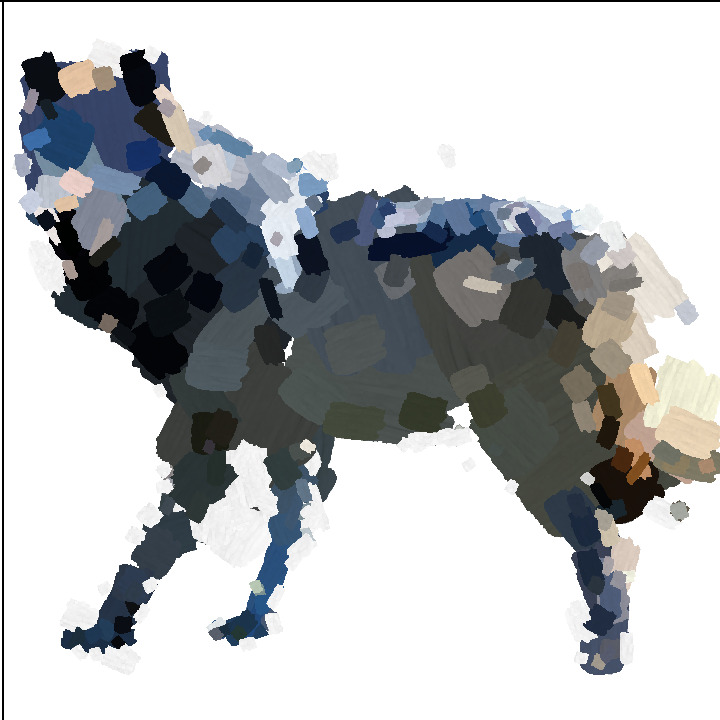} \\

\multicolumn{2}{c|}{\includegraphics[width=0.2\textwidth]{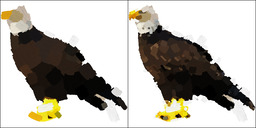}} &
\multicolumn{2}{c|}{\includegraphics[width=0.2\textwidth]{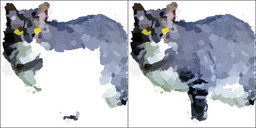}} &
\multicolumn{2}{c|}{\includegraphics[width=0.2\textwidth]{images/supp/more_per_task/6ace5ad5-8dbd-4072-b964-a54dcb4f09ca_random.jpg}} &
\multicolumn{2}{c|}{\includegraphics[width=0.2\textwidth]{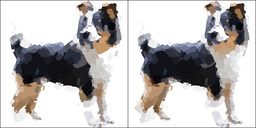}} &

\includegraphics[width=0.1\textwidth]{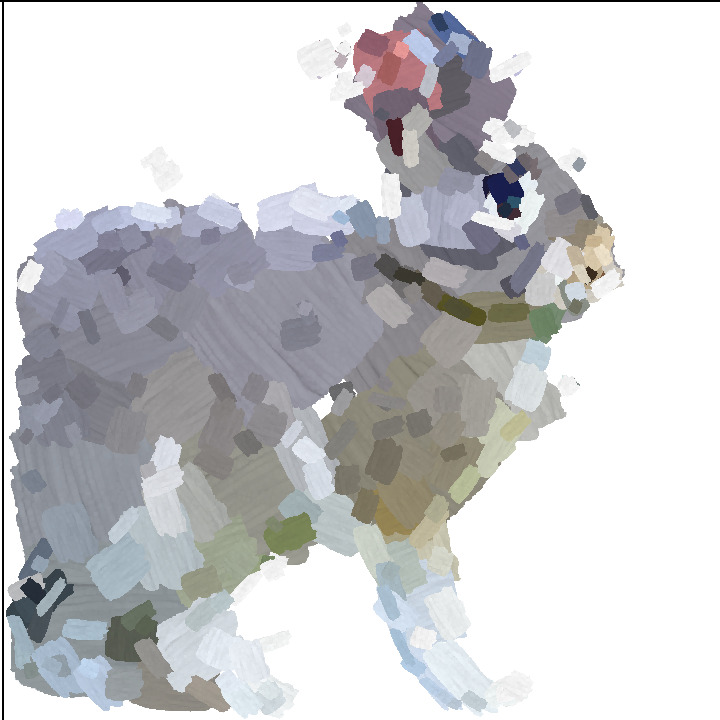} \\
\bottomrule
\end{tabular}
}
\caption{Qualitative completion results of our method. Images are organized in blocks of 2. On the left the conditioning sequence $\vecsequencecond$, on the right the conditioning with the completion added, \ie $\tilde{\boldsymbol{s}}_0$.}
\label{fig:more_qualitatives}
\end{figure*}

\endgroup
\begingroup
\renewcommand{\arraystretch}{0.} 

\begin{figure*}[!ht]
\resizebox{1.\linewidth}{!}{%
\setlength\tabcolsep{0.pt}
\begin{tabular}{cc|cc|cc|cc}
\toprule

\multicolumn{2}{c|}{\textbf{20\%}} & \multicolumn{2}{c|}{\textbf{40\%}} & \multicolumn{2}{c|}{\textbf{60\%}} & \multicolumn{2}{c}{\textbf{80\%}} \\ 
\midrule
\multicolumn{1}{c}{~~~$\vecsequencecond$} & \multicolumn{1}{c|}{$\tilde{\boldsymbol{s}}_0$} & \multicolumn{1}{c}{$\vecsequencecond$} & \multicolumn{1}{c|}{$\tilde{\boldsymbol{s}}_0$} & \multicolumn{1}{c}{$\vecsequencecond$} & \multicolumn{1}{c|}{$\tilde{\boldsymbol{s}}_0$} & \multicolumn{1}{c}{$\vecsequencecond$} & \multicolumn{1}{c|}{$\tilde{\boldsymbol{s}}_0$} \\
\midrule

\multicolumn{2}{c|}{\includegraphics[width=0.2\textwidth]{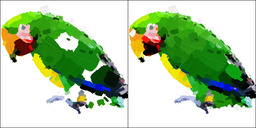}} &
\multicolumn{2}{c|}{\includegraphics[width=0.2\textwidth]{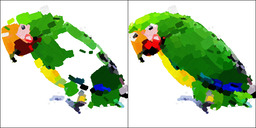}} &
\multicolumn{2}{c|}{\includegraphics[width=0.2\textwidth]{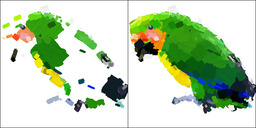}} &
\multicolumn{2}{c}{\includegraphics[width=0.2\textwidth]{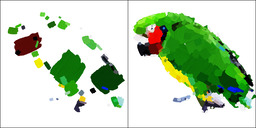}} \\

\multicolumn{2}{c|}{\includegraphics[width=0.2\textwidth]{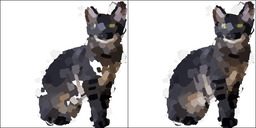}} &
\multicolumn{2}{c|}{\includegraphics[width=0.2\textwidth]{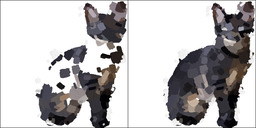}} &
\multicolumn{2}{c|}{\includegraphics[width=0.2\textwidth]{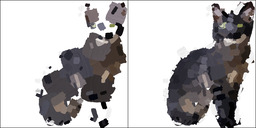}} &
\multicolumn{2}{c}{\includegraphics[width=0.2\textwidth]{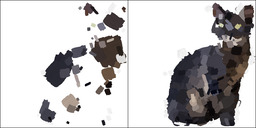}} \\

\multicolumn{2}{c|}{\includegraphics[width=0.2\textwidth]{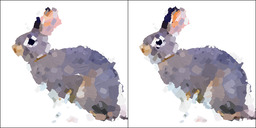}} &
\multicolumn{2}{c|}{\includegraphics[width=0.2\textwidth]{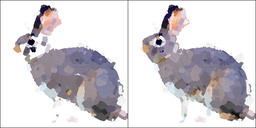}} &
\multicolumn{2}{c|}{\includegraphics[width=0.2\textwidth]{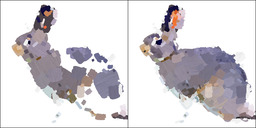}} &
\multicolumn{2}{c}{\includegraphics[width=0.2\textwidth]{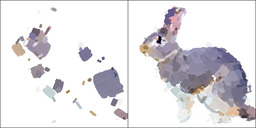}} \\

\multicolumn{2}{c|}{\includegraphics[width=0.2\textwidth]{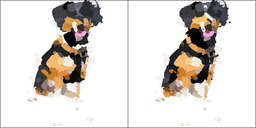}} &
\multicolumn{2}{c|}{\includegraphics[width=0.2\textwidth]{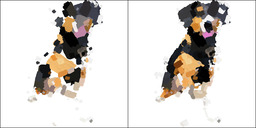}} &
\multicolumn{2}{c|}{\includegraphics[width=0.2\textwidth]{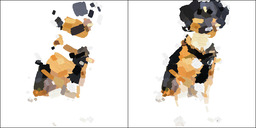}} &
\multicolumn{2}{c}{\includegraphics[width=0.2\textwidth]{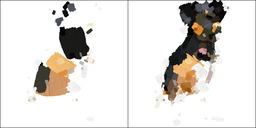}} \\

\multicolumn{2}{c|}{\includegraphics[width=0.2\textwidth]{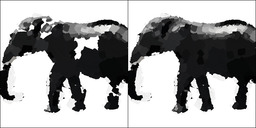}} &
\multicolumn{2}{c|}{\includegraphics[width=0.2\textwidth]{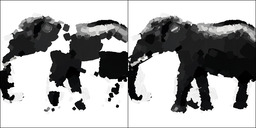}} &
\multicolumn{2}{c|}{\includegraphics[width=0.2\textwidth]{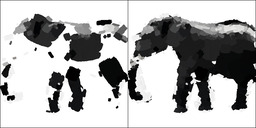}} &
\multicolumn{2}{c}{\includegraphics[width=0.2\textwidth]{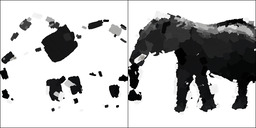}} \\

\multicolumn{2}{c|}{\includegraphics[width=0.2\textwidth]{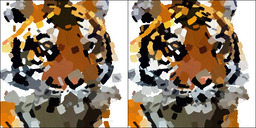}} &
\multicolumn{2}{c|}{\includegraphics[width=0.2\textwidth]{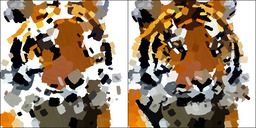}} &
\multicolumn{2}{c|}{\includegraphics[width=0.2\textwidth]{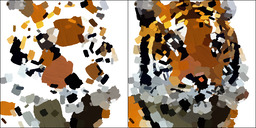}} &
\multicolumn{2}{c}{\includegraphics[width=0.2\textwidth]{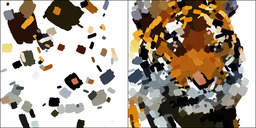}} \\

\multicolumn{2}{c|}{\includegraphics[width=0.2\textwidth]{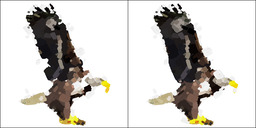}} &
\multicolumn{2}{c|}{\includegraphics[width=0.2\textwidth]{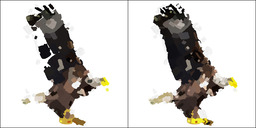}} &
\multicolumn{2}{c|}{\includegraphics[width=0.2\textwidth]{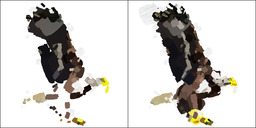}} &
\multicolumn{2}{c}{\includegraphics[width=0.2\textwidth]{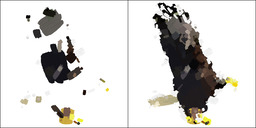}} \\

\multicolumn{2}{c|}{\includegraphics[width=0.2\textwidth]{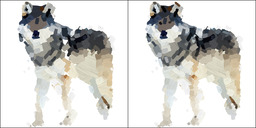}} &
\multicolumn{2}{c|}{\includegraphics[width=0.2\textwidth]{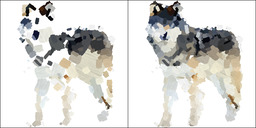}} &
\multicolumn{2}{c|}{\includegraphics[width=0.2\textwidth]{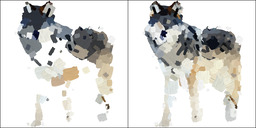}} &
\multicolumn{2}{c}{\includegraphics[width=0.2\textwidth]{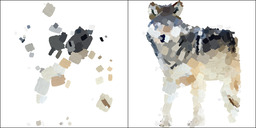}} \\

\multicolumn{2}{c|}{\includegraphics[width=0.2\textwidth]{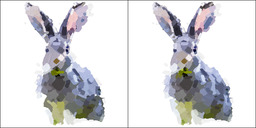}} &
\multicolumn{2}{c|}{\includegraphics[width=0.2\textwidth]{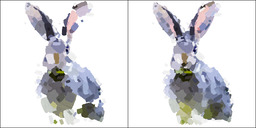}} &
\multicolumn{2}{c|}{\includegraphics[width=0.2\textwidth]{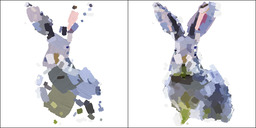}} &
\multicolumn{2}{c}{\includegraphics[width=0.2\textwidth]{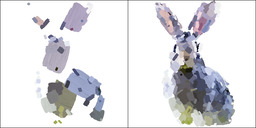}} \\

\multicolumn{2}{c|}{\includegraphics[width=0.2\textwidth]{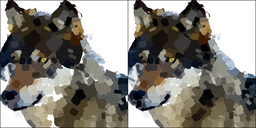}} &
\multicolumn{2}{c|}{\includegraphics[width=0.2\textwidth]{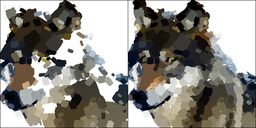}} &
\multicolumn{2}{c|}{\includegraphics[width=0.2\textwidth]{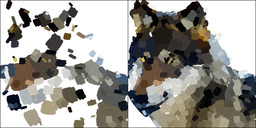}} &
\multicolumn{2}{c}{\includegraphics[width=0.2\textwidth]{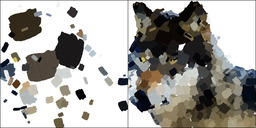}} \\

\bottomrule
\end{tabular}
}
\caption{Robustness to random masking. Images are organized in blocks of 2. On the left the conditioning sequence $\vecsequencecond$, on the right the conditioning with the completion added, \ie $\tilde{\boldsymbol{s}}_0$. The images are presented with an increasing percentage of strokes masked as conditioning from the left to the right.}
\label{fig:random_masking}
\end{figure*}

\endgroup
\begingroup
\renewcommand{\arraystretch}{0.} 

\begin{figure*}[!ht]
\resizebox{.98\linewidth}{!}{%
\setlength\tabcolsep{0.pt}
\begin{tabular}{c|ccccc}
\toprule

\multicolumn{1}{c|}{\textbf{Conditioning}} & \textbf{Completion 1} & \textbf{Completion 2} & \textbf{Completion 3} & \textbf{Completion 4} & \textbf{Completion 5} \\ 
\midrule

\includegraphics[width=0.2\textwidth]{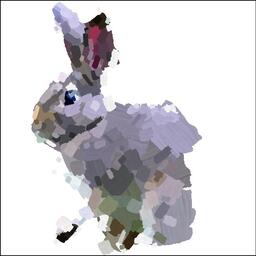} &
\includegraphics[width=0.2\textwidth]{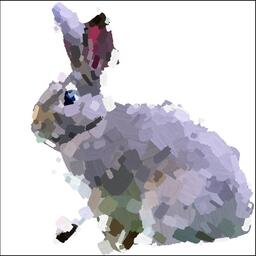} & 
\includegraphics[width=0.2\textwidth]{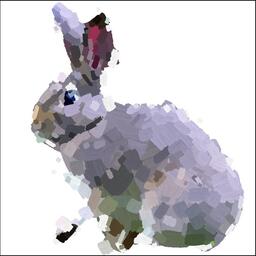} & 
\includegraphics[width=0.2\textwidth]{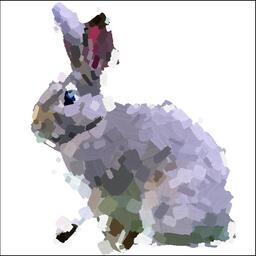} &
\includegraphics[width=0.2\textwidth]{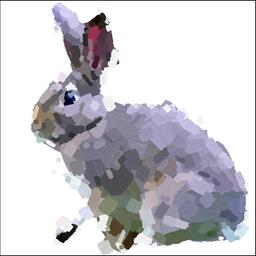} &
\includegraphics[width=0.2\textwidth]{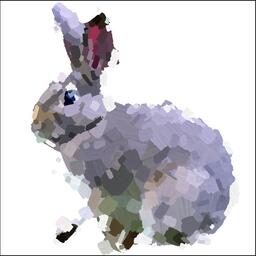} \\

\includegraphics[width=0.2\textwidth]{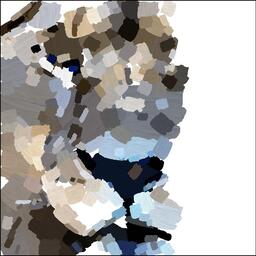} &
\includegraphics[width=0.2\textwidth]{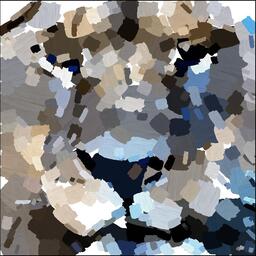} & 
\includegraphics[width=0.2\textwidth]{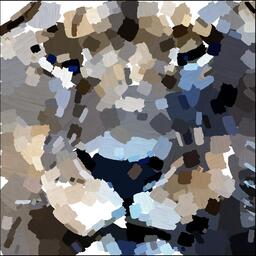} & 
\includegraphics[width=0.2\textwidth]{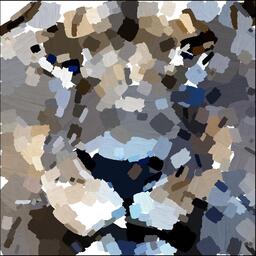} &
\includegraphics[width=0.2\textwidth]{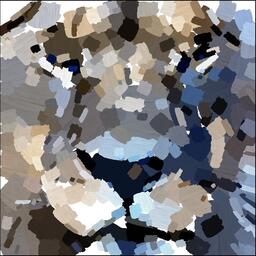} &
\includegraphics[width=0.2\textwidth]{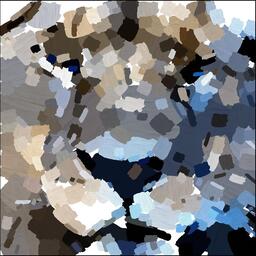} \\

\includegraphics[width=0.2\textwidth]{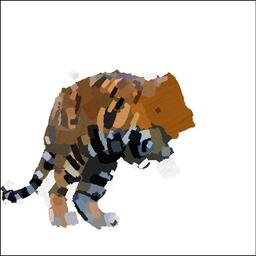} &
\includegraphics[width=0.2\textwidth]{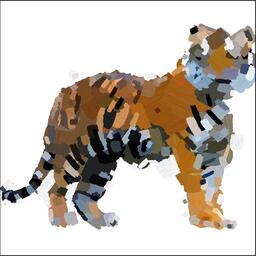} & 
\includegraphics[width=0.2\textwidth]{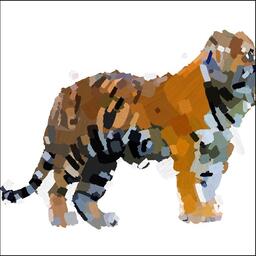} & 
\includegraphics[width=0.2\textwidth]{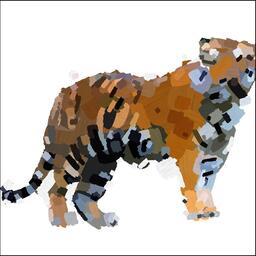} &
\includegraphics[width=0.2\textwidth]{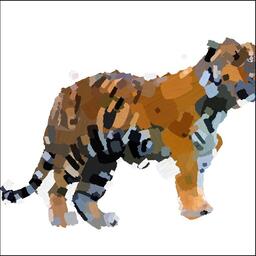} &
\includegraphics[width=0.2\textwidth]{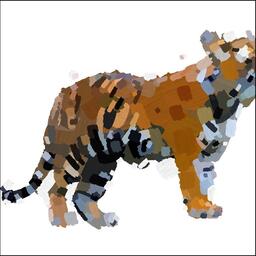} \\

\includegraphics[width=0.2\textwidth]{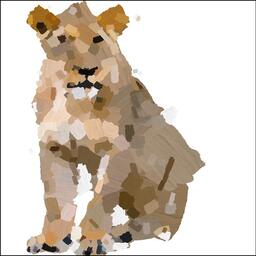} &
\includegraphics[width=0.2\textwidth]{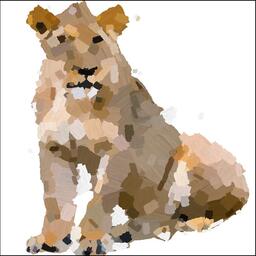} & 
\includegraphics[width=0.2\textwidth]{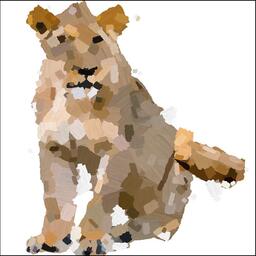} & 
\includegraphics[width=0.2\textwidth]{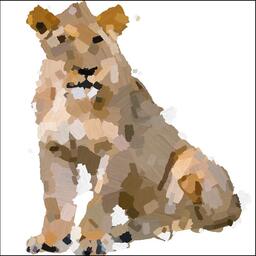} &
\includegraphics[width=0.2\textwidth]{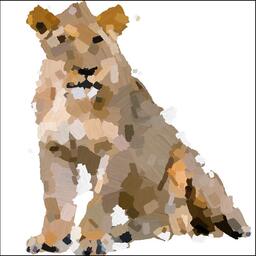} &
\includegraphics[width=0.2\textwidth]{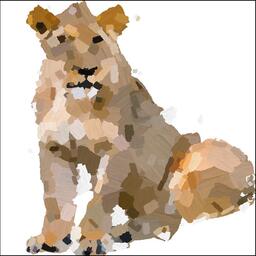} \\

\includegraphics[width=0.2\textwidth]{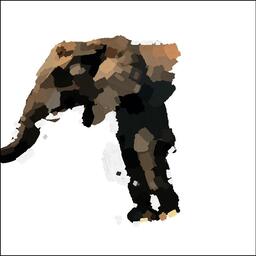} &
\includegraphics[width=0.2\textwidth]{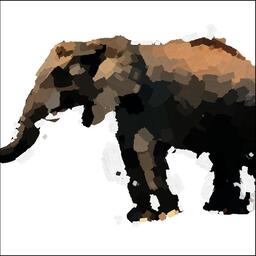} & 
\includegraphics[width=0.2\textwidth]{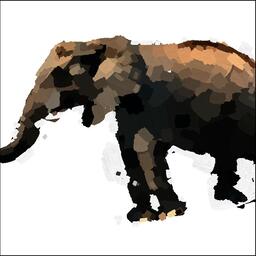} & 
\includegraphics[width=0.2\textwidth]{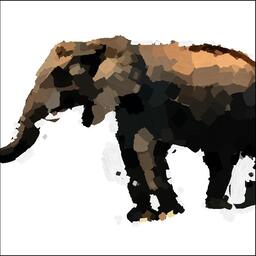} &
\includegraphics[width=0.2\textwidth]{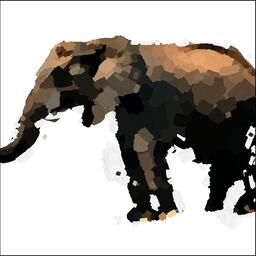} &
\includegraphics[width=0.2\textwidth]{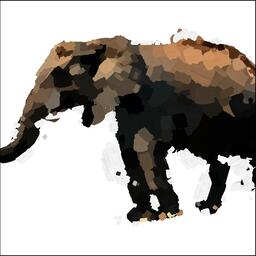} \\

\includegraphics[width=0.2\textwidth]{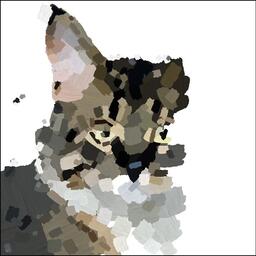} &
\includegraphics[width=0.2\textwidth]{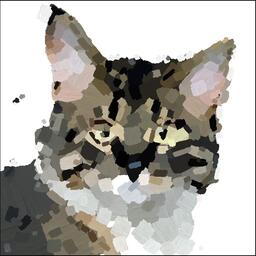} & 
\includegraphics[width=0.2\textwidth]{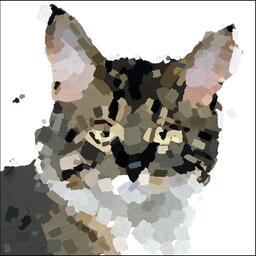} & 
\includegraphics[width=0.2\textwidth]{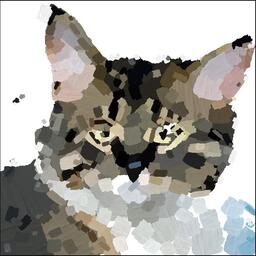} &
\includegraphics[width=0.2\textwidth]{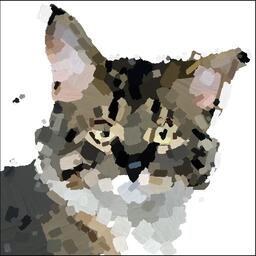} &
\includegraphics[width=0.2\textwidth]{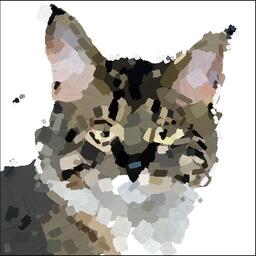} \\

\includegraphics[width=0.2\textwidth]{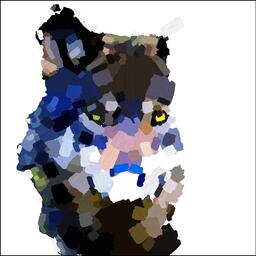} &
\includegraphics[width=0.2\textwidth]{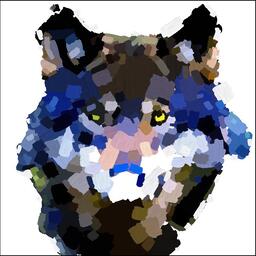} & 
\includegraphics[width=0.2\textwidth]{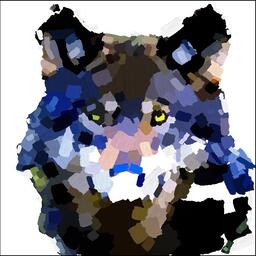} & 
\includegraphics[width=0.2\textwidth]{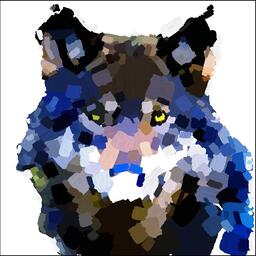} &
\includegraphics[width=0.2\textwidth]{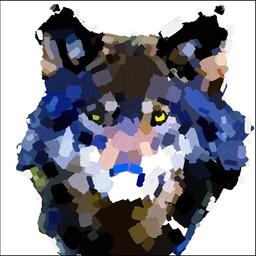} &
\includegraphics[width=0.2\textwidth]{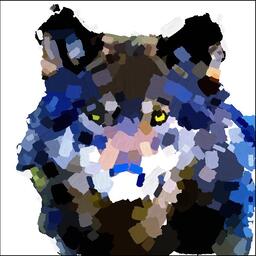} \\

\includegraphics[width=0.2\textwidth]{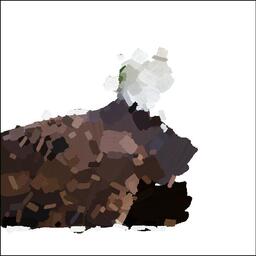} &
\includegraphics[width=0.2\textwidth]{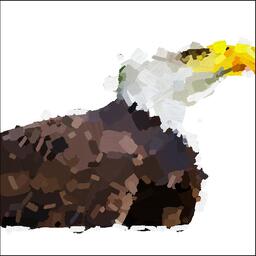} & 
\includegraphics[width=0.2\textwidth]{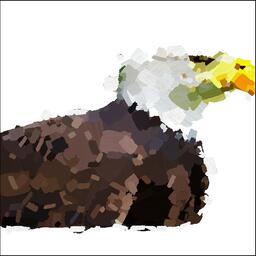} & 
\includegraphics[width=0.2\textwidth]{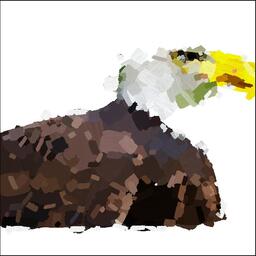} &
\includegraphics[width=0.2\textwidth]{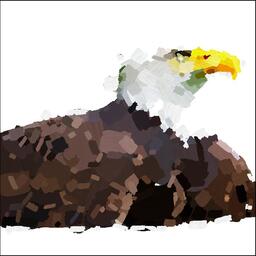} &
\includegraphics[width=0.2\textwidth]{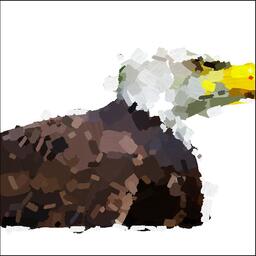} \\

\bottomrule
\end{tabular}
}
\caption{Diverse suggestions from the same context. On the left is the conditioning sequence $\vecsequencecond$, and on the right are the diverse completions.}
\label{fig:multiple}
\end{figure*}

\endgroup

\begingroup
\renewcommand{\arraystretch}{0.} 

\begin{figure*}[!ht]
\resizebox{1.\linewidth}{!}{%
\setlength\tabcolsep{0.pt}
\begin{tabular}{cc|cc|cc|cc|cc}
\toprule

\multicolumn{1}{c}{$\vecsequencecond$} & \multicolumn{1}{c|}{$\tilde{\boldsymbol{s}}_0$} & \multicolumn{1}{c}{$\vecsequencecond$} & \multicolumn{1}{c|}{$\tilde{\boldsymbol{s}}_0$} & \multicolumn{1}{c}{$\vecsequencecond$} & \multicolumn{1}{c|}{$\tilde{\boldsymbol{s}}_0$} & \multicolumn{1}{c}{$\vecsequencecond$} & \multicolumn{1}{c|}{$\tilde{\boldsymbol{s}}_0$} & \multicolumn{1}{c}{$\vecsequencecond$} & \multicolumn{1}{c|}{$\tilde{\boldsymbol{s}}_0$} \\
\midrule

\multicolumn{2}{c|}{\includegraphics[width=0.2\textwidth]{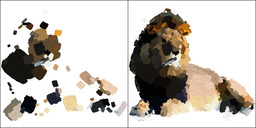}} &
\multicolumn{2}{c|}{\includegraphics[width=0.2\textwidth]{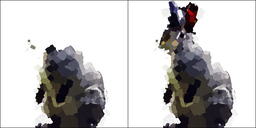}} &
\multicolumn{2}{c|}{\includegraphics[width=0.2\textwidth]{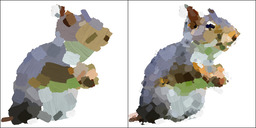}} &
\multicolumn{2}{c|}{\includegraphics[width=0.2\textwidth]{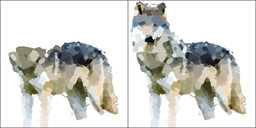}} &
\multicolumn{2}{c|}{\includegraphics[width=0.2\textwidth]{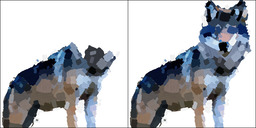}} \\

\multicolumn{2}{c|}{\includegraphics[width=0.2\textwidth]{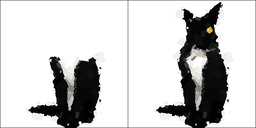}} &
\multicolumn{2}{c|}{\includegraphics[width=0.2\textwidth]{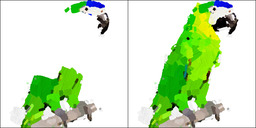}} &
\multicolumn{2}{c|}{\includegraphics[width=0.2\textwidth]{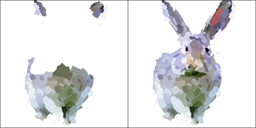}} &
\multicolumn{2}{c|}{\includegraphics[width=0.2\textwidth]{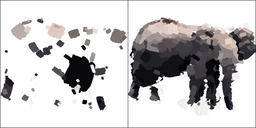}} &
\multicolumn{2}{c|}{\includegraphics[width=0.2\textwidth]{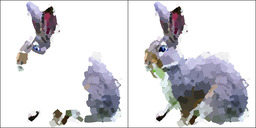}} \\

\multicolumn{2}{c|}{\includegraphics[width=0.2\textwidth]{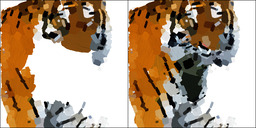}} &
\multicolumn{2}{c|}{\includegraphics[width=0.2\textwidth]{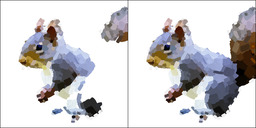}} &
\multicolumn{2}{c|}{\includegraphics[width=0.2\textwidth]{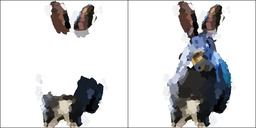}} &
\multicolumn{2}{c|}{\includegraphics[width=0.2\textwidth]{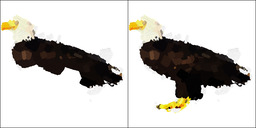}} &
\multicolumn{2}{c|}{\includegraphics[width=0.2\textwidth]{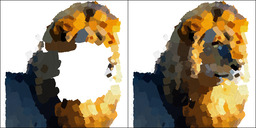}} \\

\multicolumn{2}{c|}{\includegraphics[width=0.2\textwidth]{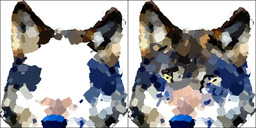}} &
\multicolumn{2}{c|}{\includegraphics[width=0.2\textwidth]{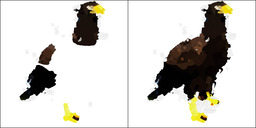}} &
\multicolumn{2}{c|}{\includegraphics[width=0.2\textwidth]{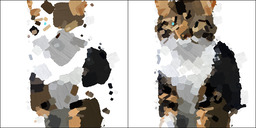}} &
\multicolumn{2}{c|}{\includegraphics[width=0.2\textwidth]{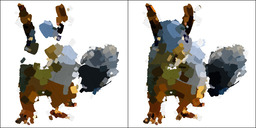}} &
\multicolumn{2}{c|}{\includegraphics[width=0.2\textwidth]{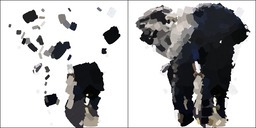}} \\

\multicolumn{2}{c|}{\includegraphics[width=0.2\textwidth]{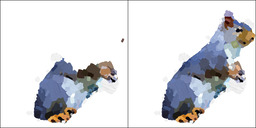}} &
\multicolumn{2}{c|}{\includegraphics[width=0.2\textwidth]{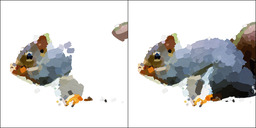}} &
\multicolumn{2}{c|}{\includegraphics[width=0.2\textwidth]{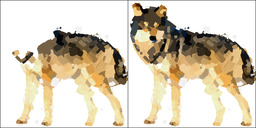}} &
\multicolumn{2}{c|}{\includegraphics[width=0.2\textwidth]{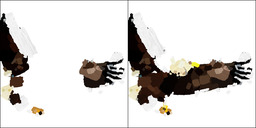}} &
\multicolumn{2}{c|}{\includegraphics[width=0.2\textwidth]{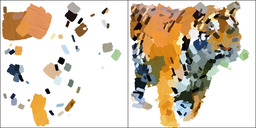}} \\

\multicolumn{2}{c|}{\includegraphics[width=0.2\textwidth]{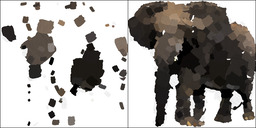}} &
\multicolumn{2}{c|}{\includegraphics[width=0.2\textwidth]{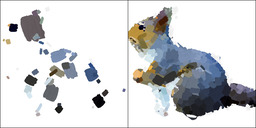}} &
\multicolumn{2}{c|}{\includegraphics[width=0.2\textwidth]{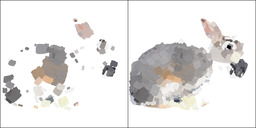}} &
\multicolumn{2}{c|}{\includegraphics[width=0.2\textwidth]{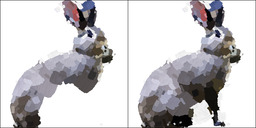}} &
\multicolumn{2}{c|}{\includegraphics[width=0.2\textwidth]{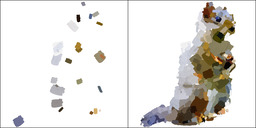}} \\

\multicolumn{2}{c|}{\includegraphics[width=0.2\textwidth]{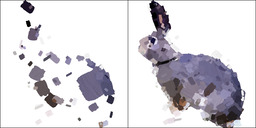}} &
\multicolumn{2}{c|}{\includegraphics[width=0.2\textwidth]{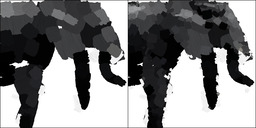}} &
\multicolumn{2}{c|}{\includegraphics[width=0.2\textwidth]{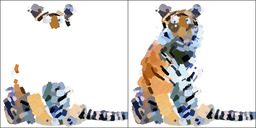}} &
\multicolumn{2}{c|}{\includegraphics[width=0.2\textwidth]{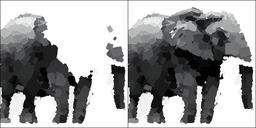}} &
\multicolumn{2}{c|}{\includegraphics[width=0.2\textwidth]{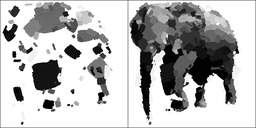}} \\

\multicolumn{2}{c|}{\includegraphics[width=0.2\textwidth]{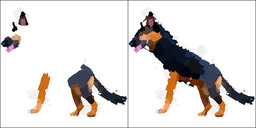}} &
\multicolumn{2}{c|}{\includegraphics[width=0.2\textwidth]{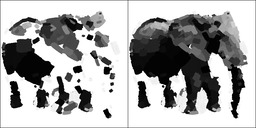}} &
\multicolumn{2}{c|}{\includegraphics[width=0.2\textwidth]{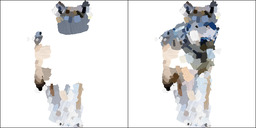}} &
\multicolumn{2}{c|}{\includegraphics[width=0.2\textwidth]{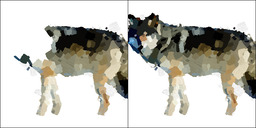}} &
\multicolumn{2}{c|}{\includegraphics[width=0.2\textwidth]{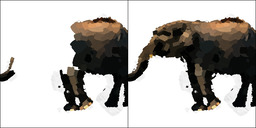}} \\

\multicolumn{2}{c|}{\includegraphics[width=0.2\textwidth]{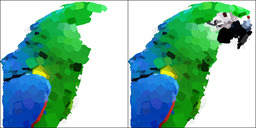}} &
\multicolumn{2}{c|}{\includegraphics[width=0.2\textwidth]{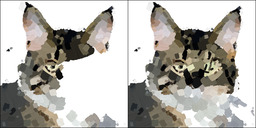}} &
\multicolumn{2}{c|}{\includegraphics[width=0.2\textwidth]{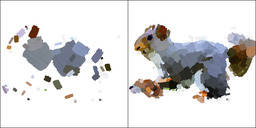}} &
\multicolumn{2}{c|}{\includegraphics[width=0.2\textwidth]{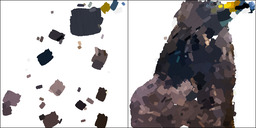}} &
\multicolumn{2}{c|}{\includegraphics[width=0.2\textwidth]{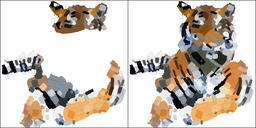}} \\

\multicolumn{2}{c|}{\includegraphics[width=0.2\textwidth]{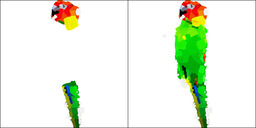}} &
\multicolumn{2}{c|}{\includegraphics[width=0.2\textwidth]{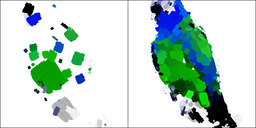}} &
\multicolumn{2}{c|}{\includegraphics[width=0.2\textwidth]{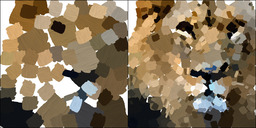}} &
\multicolumn{2}{c|}{\includegraphics[width=0.2\textwidth]{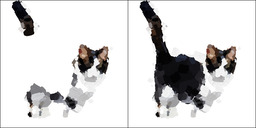}} &
\multicolumn{2}{c|}{\includegraphics[width=0.2\textwidth]{images/supp/no_class/d528f5a4-8c7f-4728-8d6e-1cf0f73072cc_square.jpg}} \\

\bottomrule
\end{tabular}
}
\caption{Automatic inference without providing class. Images are organized in blocks of 2. On the left the conditioning sequence $\vecsequencecond$, on the right the conditioning with the completion added but no class provided.}
\label{fig:no_class}
\end{figure*}

\endgroup

\bibliographystyle{model2-names}
\bibliography{refs}



\end{document}